%% file: main.tex
\documentclass[sigconf,natbib=false]{acmart}

\def\shownotes{1}
\usepackage{macro}
\usepackage[backend=biber,
            style=numeric-comp,
            bibstyle=numeric,
            natbib=true,
            maxbibnames=99,
            minalphanames=3,
            maxcitenames=2,
            sorting=ynt,
            giveninits=true,            
            labeldateparts=true,
            sortcites=true,
            backref=false]{biblatex}
\DefineBibliographyStrings{english}{%
  backrefpage = {\hspace{-0.0500cm}},%
  backrefpages = {\hspace{-0.0500cm}},%
}
\usepackage{multirow, makecell}
\usepackage{graphicx}
\usepackage{booktabs}
\usepackage{bm}
\usepackage{enumitem}
\usepackage{wrapfig}
\usepackage{subcaption}
\usepackage{balance}
\usepackage{algorithm,algorithmicx,algpseudocode}

\copyrightyear{2026}
\acmYear{2026}
\setcopyright{cc}
\setcctype{by}
\acmConference[KDD '26]{Proceedings of the 32nd ACM SIGKDD Conference on Knowledge Discovery and Data Mining V.1}{August 09--13, 2026}{Jeju Island, Republic of Korea}
\acmBooktitle{Proceedings of the 32nd ACM SIGKDD Conference on Knowledge Discovery and Data Mining V.1 (KDD '26), August 09--13, 2026, Jeju Island, Republic of Korea}
\acmPrice{}
\acmDOI{10.1145/3770854.3785677}
\acmISBN{979-8-4007-2258-5/2026/08}

\usepackage{microtype}
\usepackage{subcaption}
\usepackage{amsmath}

\usepackage{graphicx}
\usepackage{caption}
\usepackage{dsfont}
\usepackage{subcaption}
\usepackage{booktabs}
\usepackage{algorithm,algorithmicx,algpseudocode}
\usepackage{enumitem}
\usepackage{multirow}
\usepackage{pifont}
\usepackage{wrapfig}
\usepackage{siunitx}
\usepackage{listings}
\usepackage{xcolor}
\definecolor{wine}{RGB}{128, 0, 32}
\lstdefinelanguage{CSV}{
    morestring=[b]",
    morecomment=[l]{\#},
    alsoletter={0123456789},
    sensitive=false
}

\usepackage{macro}
\newcommand{\cmark}{\textcolor[rgb]{0.13, 0.55, 0.13}{\ding{51}}}%
\newcommand{\xmark}{\textcolor[rgb]{0.5, 0, 0}{\ding{55}}}%

\begin{document}

\title{Learning Multimodal Embeddings for Traffic Accident Prediction and Causal Estimation}

\author{Ziniu Zhang}
\affiliation{%
  \institution{Northeastern University}
  \city{Boston}
  \state{MA}
  \country{USA}
}
\email{zhang.zini@northeastern.edu}

\author{Minxuan Duan}
\affiliation{%
  \institution{Northeastern University}
  \city{Boston}
  \state{MA}
  \country{USA}
}
\email{duan.mi@northeastern.edu}

\author{Haris N. Koutsopoulos}
\affiliation{%
  \institution{Northeastern University}
  \city{Boston}
  \state{MA}
  \country{USA}
}
\email{h.koutsopoulos@northeastern.edu}

\author{Hongyang R. Zhang}
\affiliation{%
  \institution{Northeastern University}
  \city{Boston}
  \state{MA}
  \country{USA}
}
\email{ho.zhang@northeastern.edu}

\renewcommand{\authors}{Ziniu Zhang, Minxuan Duan, Haris N. Koutsopoulos, and Hongyang R. Zhang}
\renewcommand{\shortauthors}{Ziniu Zhang, Minxuan Duan, Haris N. Koutsopoulos, and Hongyang R. Zhang}

\input{abstract}
\begin{CCSXML}
<ccs2012>
<concept>
<concept_id>10010147.10010257.10010293.10010294</concept_id>
<concept_desc>Computing methodologies~Neural networks</concept_desc>
<concept_significance>500</concept_significance>
</concept>
 <concept>
<concept_id>10010147.10010257.10010293.10010319</concept_id>
<concept_desc>Computing methodologies~Learning latent representations</concept_desc>
<concept_significance>500</concept_significance>
 </concept>
 <concept>
<concept_id>10010147.10010178.10010187.10010192</concept_id>
<concept_desc>Computing methodologies~Causal reasoning and diagnostics</concept_desc>
<concept_significance>500</concept_significance>
</ccs2012>
\end{CCSXML}

\ccsdesc[500]{Computing methodologies~Neural networks}
\ccsdesc[500]{Computing methodologies~Learning latent representations}
\ccsdesc[500]{Computing methodologies~Causal reasoning and diagnostics}

\keywords{Multimodal Learning; Satellite Imagery; Road Safety; Causal Inference and Analysis}

\maketitle

\input{content}

\begin{refcontext}[sorting=nyt]
\printbibliography
\end{refcontext}

\clearpage
\appendix
\input{appendix}

\end{document}

%% file: abstract.tex
\begin{abstract}
We consider analyzing traffic accident patterns using both road network data and satellite images aligned to road graph nodes. Previous work for predicting accident occurrences relies primarily on road network structural features while overlooking physical and environmental information from the road surface and its surroundings. In this work, we construct a large multimodal dataset spanning six U.S. states, containing nine million traffic accident records from official sources, and one million high-resolution satellite images for each node of the road network. Additionally, every node is annotated with features such as the region's weather statistics and road type (e.g., residential vs. motorway), and each edge is annotated with traffic volume information (i.e., Average Annual Daily Traffic). Utilizing this dataset, we conduct a comprehensive evaluation of multimodal learning methods that integrate both visual and network embeddings. Our findings show that integrating both data modalities improves prediction accuracy, achieving an average AUROC of $90.1\%$, a $3.7\%$ gain over graph neural network models that use only graph structures. With the improved embeddings, we conduct a causal analysis using a matching estimator to identify the key factors influencing traffic accidents. We find that accident rates rise by $24\%$ under higher precipitation, by $22\%$ on higher-speed roads such as motorways, and by $29\%$ due to seasonal patterns, after adjusting for other confounding factors. Ablation studies confirm that satellite imagery features are essential for achieving accurate prediction. We release the dataset and the experimental code for using this dataset at \href{https://github.com/VirtuosoResearch/MMTraCE}{https://github.com/VirtuosoResearch/MMTraCE}.
\end{abstract}

%% file: content.tex
\section{Introduction}

Road safety remains a persistent and pressing challenge in urban cities and districts. 
According to the World Health Organization, approximately 1.19 million people die in road traffic crashes each year, making it the leading cause of death for children and young adults~\cite{world2019global}. In the United States alone, traffic accidents are projected to cause 39,345 deaths in 2024 and \$871 billion in annual societal costs~\cite{nhtsa2024early, blincoe2015economic}.
Faced with such a severe reality, it is imperative to shift from reactive analysis to proactive prediction, using data-driven methods to identify high-risk road segments and implement interventions to prevent accidents.

Current data-driven models, especially those based on graph neural networks (GNNs), draw on road network topology, traffic flow statistics, and historical accident data~\cite{zhuang2023sauc, yao2019revisiting, li2018diffusion, nippani2023graph}. These data sources have improved the modeling of accident patterns by capturing spatial relationships~\cite{zhuang2022uncertainty}. However, structural features fail to reflect the visual and environmental conditions of the physical road. Factors such as lane width, curvature, surface quality, and nearby land use shape driver behavior and accident likelihood, yet they do not appear in standard network data.

Advances in remote sensing and computer vision provide a path to fill this gap. These tools support a road-level view that links physical scenes to accident patterns. Recent work in transportation and climate modeling has shown clear gains from hybrid models that combine high-dimensional visual inputs with traditional numeric features~\cite{wang2024deep, meng2025physics, wu2025grapheval36k}. Satellite images offer complementary information to network-based representations and make visible the physical road attributes that influence accident risk. Examples are shown in Figure~\ref{fig_satellite_sample} (see also Figure~\ref{fig_omitted_satellite_sample} in the Appendix). These attributes are hard to extract from conventional data sources but are closely linked to accident risk~\cite{he2021riskmap}.

\begin{figure}[t!]
    \centering
    \begin{subfigure}[b]{0.475\textwidth}
    \begin{minipage}[b]{0.32\textwidth}
        \centering
        \includegraphics[width=0.985\textwidth]{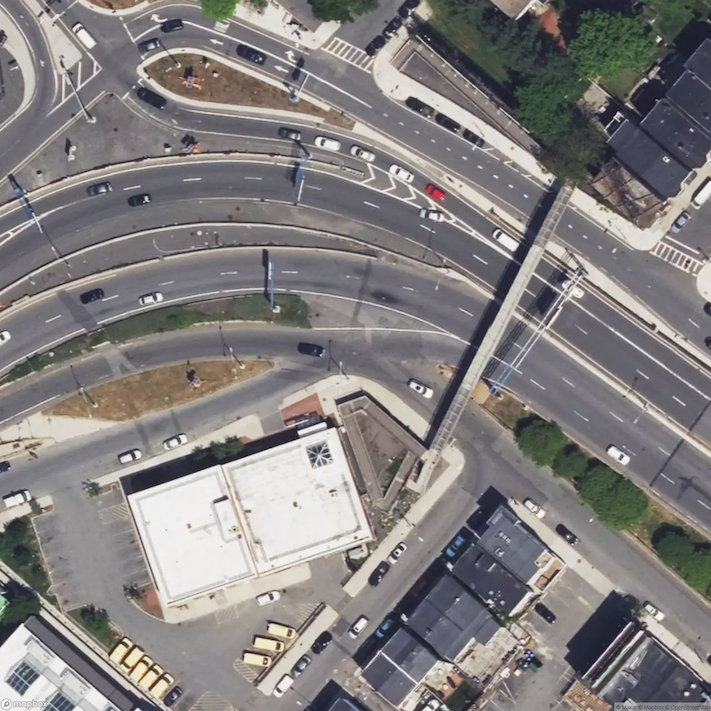}
    \end{minipage}\hfill
    \begin{minipage}[b]{0.32\textwidth}
        \centering
        \includegraphics[width=0.985\textwidth]{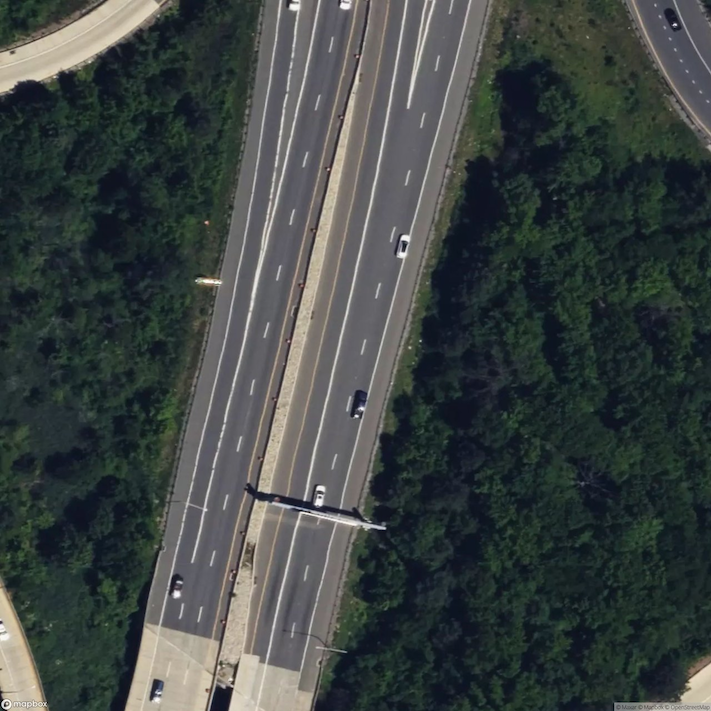}
    \end{minipage}\hfill
    \begin{minipage}[b]{0.32\textwidth}
        \centering
        \includegraphics[width=0.985\textwidth]{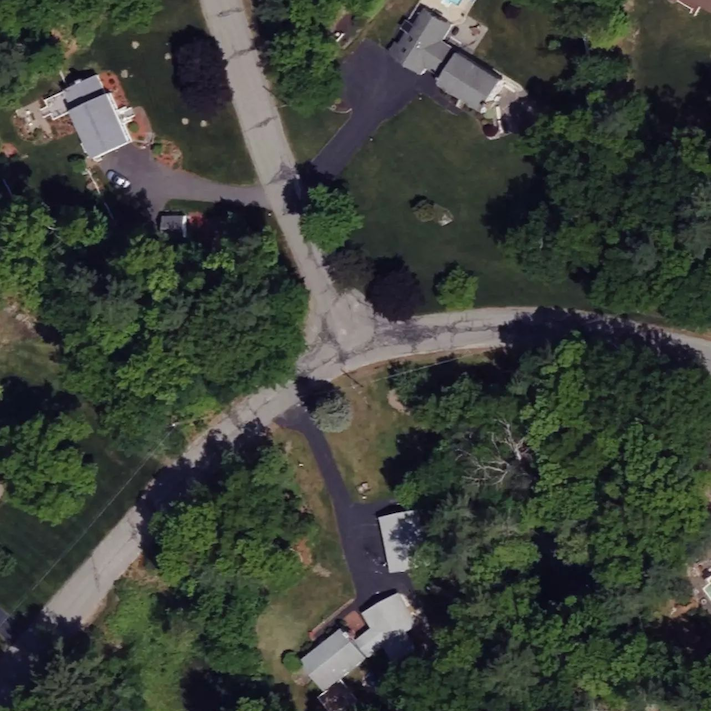}
    \end{minipage}\hfill\\[0.5em]
    \begin{minipage}[b]{0.32\textwidth}
        \centering
        \includegraphics[width=0.985\textwidth]{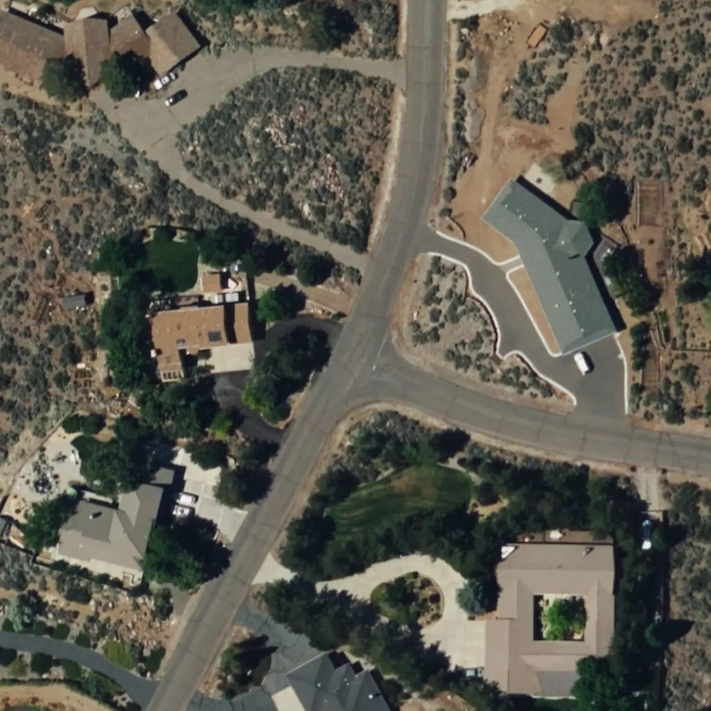}
    \end{minipage}\hfill
    \begin{minipage}[b]{0.32\textwidth}
        \centering
        \includegraphics[width=0.985\textwidth]{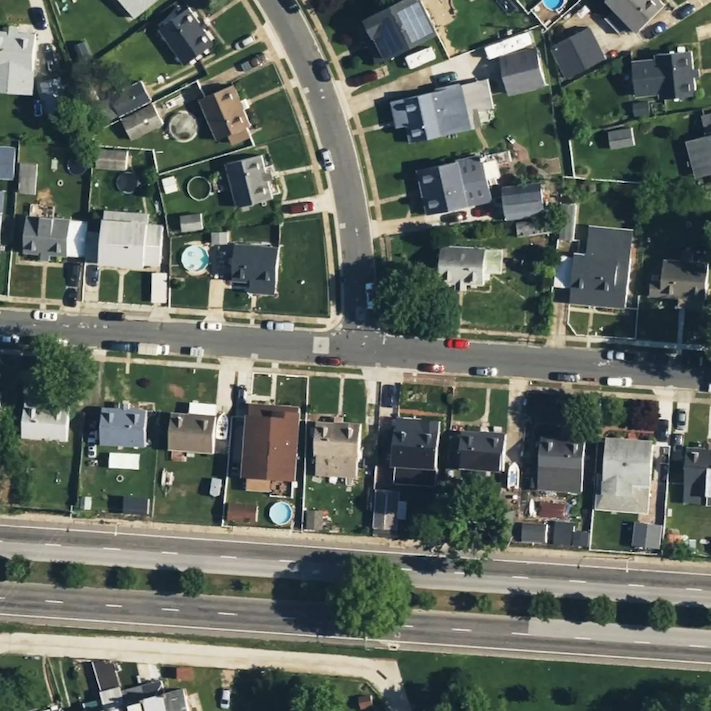}
    \end{minipage}\hfill
    \begin{minipage}[b]{0.32\textwidth}
        \centering
        \includegraphics[width=0.985\textwidth]{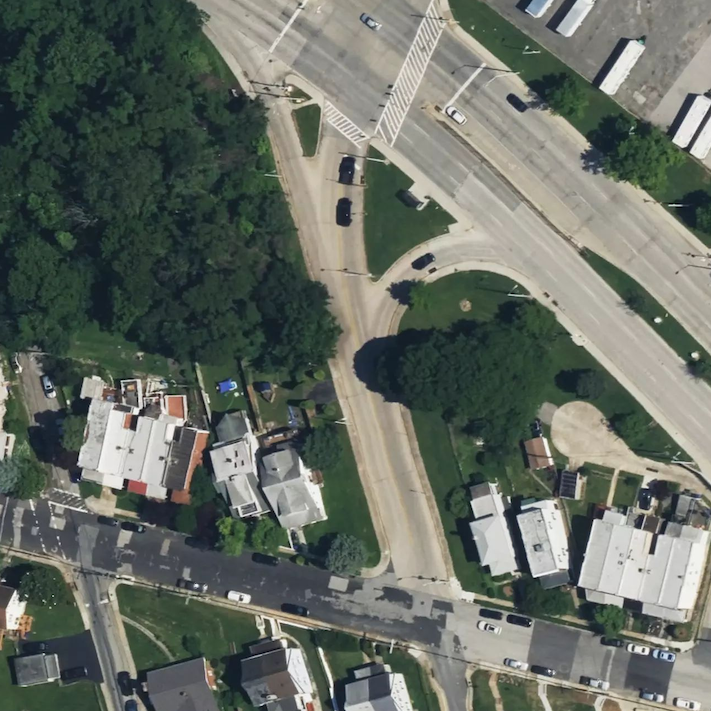}
    \end{minipage}
    \end{subfigure}
    \caption{Example satellite images showing different types of roads. Each image is centered around a road network node and captures both the physical characteristics of the road, such as layout, width, and intersections, and the surrounding context, including vegetation, buildings, and terrain.}\label{fig_satellite_sample}
    \Description{Sample satellite images from different road conditions}
\end{figure}

Despite this potential, the use of satellite imagery in state-level traffic accident analysis remains limited. There are two main challenges. The first challenge is the lack of large public multimodal datasets with enough temporal coverage across different regions~\cite{jenelius2013travel, yang1996microscopic, wang2024comparing}. This gap comes from the heterogeneity of data sources. Each state publishes accident and traffic records in its own format. They also use different variable names and hosting structures. Unifying these datasets requires careful preprocessing and normalization. This includes schema alignment and geospatial matching.
The second challenge is the need to reduce the dimensionality of high-resolution satellite images. These images must also be integrated with graph-structural features and other low-dimensional inputs. Examples include weather, traffic volume, and speed limits. Early experiments show that simple fusion methods do not work well. Direct concatenation fails to capture nonlinear interactions between different modalities. As a result, models struggle to effectively use both visual and non-visual information.

To address these challenges, we begin by constructing a large-scale, multimodal dataset spanning six U.S. states. It comprises over $9$ million accident records and $1$ million satellite images, each with rich feature annotations.
\begin{itemize}%
    \item Each road network node is aligned with high-resolution satellite images and supplemented with road-level features, including weather conditions (e.g., temperature, precipitation, wind speed, and atmospheric pressure). Traffic volume indicators of each edge, such as Average Annual Daily Traffic (AADT), are also associated with each node when available.
    \item Each satellite image has a resolution of $1024 \times 1024$ pixels and covers approximately $200 \, \mathrm{m} \times 200 \, \mathrm{m}$ of physical space.
    \item Every road is also associated with historical accident occurrences.
    Notably, we collect the latest accident data through $2025$, with a maximum temporal span of $24$ years, allowing models to capture long-term trends and evolving risk patterns. This dense spatial alignment enables models to jointly learn from both visual cues and graph-structural features.
\end{itemize}
Compared to existing datasets~\cite{yuan2018hetero,huang2023tap,nippani2023graph}, our dataset covers a much wider geographic range. It also provides detailed annotations of satellite images. These features make it well-suited for large-scale multimodal traffic accident analysis.

Building on this dataset, we develop a multimodal learning framework. The framework combines vision-based encoders with GNNs and supports joint reasoning over visual, structural, and contextual properties of the road environment. We conduct a detailed evaluation of different fusion strategies to measure their impact on traffic accident prediction.
By adding satellite imagery to the model, we obtain an average AUROC of $90.1\%$ across six states. The prediction accuracy improves by $3.7\%$ compared to GNNs that use only graph-structural features. Based on this result, we further perform a causal analysis of key environmental and traffic factors. We estimate the average treatment effect on the treated (ATT) using a matching estimator based on multimodal embeddings. The ATT scores for seasonal variation, road type, and precipitation are $28.6\%$, $21.9\%$, and $24.2\%$. We further estimate the effects with propensity score-based adjustments (PSM) and the double robust estimator (DR).
We run a leave-one-out ablation study to measure the contribution of each feature type. Removing image features leads to a $3.5\%$ drop in AUROC. Excluding weather, traffic volume, and road network features results in decreases of $1.8\%$, $2.4\%$, and $3.7\%$, respectively.

In summary, this paper makes three contributions to the study of traffic accidents on road networks. First, we build a large multimodal dataset with recent traffic accident records from six U.S. states. Each satellite image is aligned with a specific road network node. To our knowledge, this is the most comprehensive dataset for evaluating multimodal learning in the transportation domain.
Second, we provide a detailed evaluation of multimodal learning methods that combine visual embeddings with network embeddings. The results show clear gains in accident prediction accuracy.
Third, we conduct a causal analysis to measure the effects of key factors influencing accident occurrences. We use the learned embeddings to control for confounding variables such as road conditions.
We hope that this dataset can support future research on multimodal learning for transportation and road safety.

\section{Methodology}\label{sec_method}

We now describe our multimodal dataset and the accompanying analytical tools.
First, we provide an overview of the problem setting. We then introduce the dataset collection, including the Satellite images.
Next, we describe the multimodal approaches that integrate network and visual image embeddings.
Finally, we design a causal analysis framework on top of the multimodal embeddings.

\subsection{Preliminaries}\label{sec_preliminaries}

We study the problem of accident prediction and analysis on road networks. The road network is modeled as a directed graph $G = (V, E)$, where each node $v \in V$ represents a road intersection or critical point along a road segment, and each directed edge $e \in E$ represents a road connecting two such points. Our goal is to predict the probability or frequency of accidents occurring at each edge, leveraging both spatial structure and temporal dynamics.

Each node is associated with dynamic features that evolve over time, such as weather conditions, including temperature, wind speed, precipitation, and sea-level pressure. Additionally, each node is linked to a high-resolution satellite image centered at its geographic coordinates. We extract visual features from the image that capture the road structure and the surrounding environment.

Each edge is associated with static and dynamic attributes that characterize the corresponding road segment. These include road length, road type (e.g., residential, motorway), and average annual daily traffic (AADT).

Formally, at each time step $t$, we observe a dynamic graph $G_t = (V, E, X_t)$, where $X_t$ denotes the time-dependent attributes. Given a sequence of previous observations $\{G_{t-T+1}, \ldots, G_t\}$, where $T$ is the observation period, the objective is to predict accident risk at each node for the next time step:
\begin{align*}
    \hat{y}_{v}^{(t+1)} = f(G_{t-T+1}, \ldots, G_t; \mathcal{I}_v),
\end{align*}
where $\mathcal{I}_v$ denotes the visual embedding derived from the satellite image corresponding to node $v$. The function $f(\cdot)$ is parameterized by a multimodal spatiotemporal model that jointly reasons over the graph structure, node dynamics, edge dynamics, and image content.

\subsection{Dataset Collection}

The dataset is assembled in several stages. We begin by constructing road network graphs for six states, then collect high-resolution satellite images spatially aligned with these road graphs.
Finally, we collect historical traffic accident records from official sources and align them with both the road network and satellite images.

\begin{table}[t]
\centering
\caption{Statistics of road network graphs for each state in our dataset. We report the total number of edges ($m$), average edge length in meters, road network density ($m / \binom{n}{2}$, where $n$ is the number of nodes), which is the number of edges per unit area, and availability of traffic volume, which indicates the proportion of road segments that are associated with available traffic volume measurements.}\label{tab_graph_dataset_statistic}
\resizebox{1.00\columnwidth}{!}
{
\begin{tabular}{@{}lcccc@{}}
\toprule
 & {$\#$ Edges} & {Avg Length (m)} & {Density} & {Volume (\%)}\\ 
\midrule
Delaware        & $116,196$ & $213.0$ & $9.7 \times 10^{-5}$ & $3.7$ \\
Massachusetts   & $706,402$ & $188.8$ & $1.7 \times 10^{-5}$ & $0.9$ \\
Maryland        & $580,526$ & $211.7$ & $1.8 \times 10^{-5}$ & $2.1$ \\
Nevada          & $292,674$ & $280.1$ & $4.0 \times 10^{-5}$ & $1.3$ \\
Montana         & $351,516$ & $859.2$ & $3.3 \times 10^{-5}$ & $0.8$ \\
Iowa            & $707,072$ & $532.4$ & $2.2 \times 10^{-5}$ & $-$ \\
\bottomrule
\end{tabular}}
\end{table}

\begin{figure}[t!]
    \centering
    \includegraphics[width=0.9995\columnwidth]{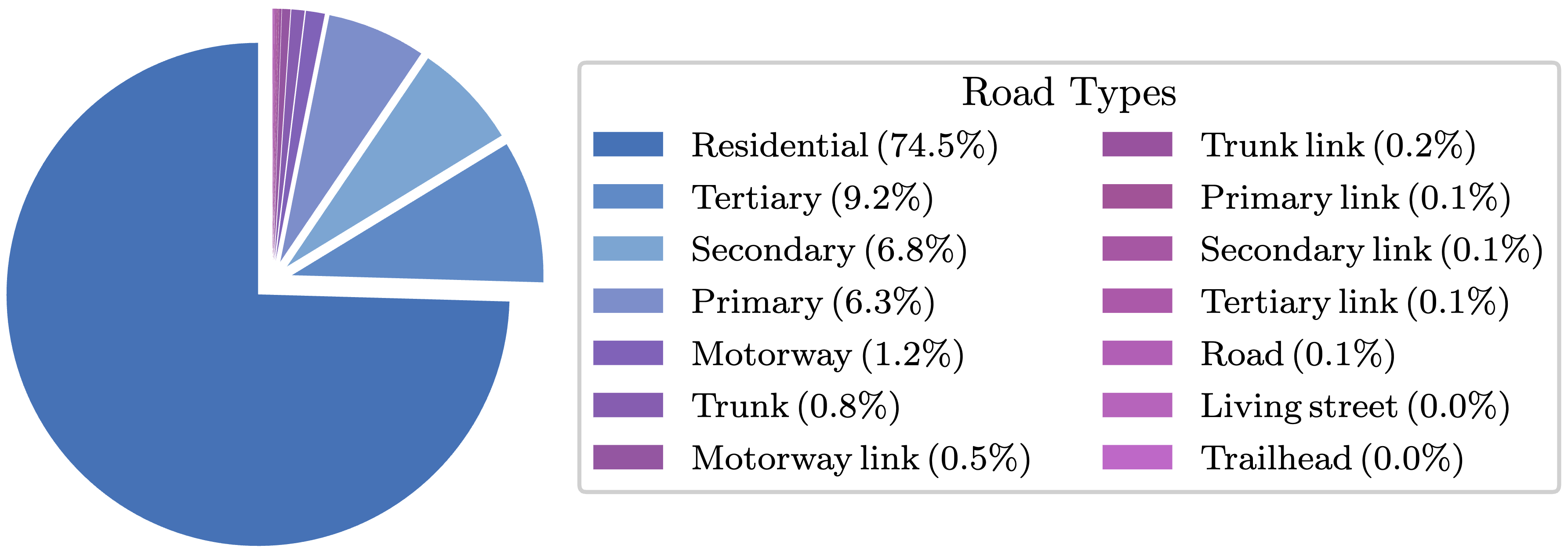}
    \caption{The proportion of different road types among six states' road networks. Residential roads account for the vast majority of all roads, making up approximately $74.5\%$. Other types, such as tertiary, secondary, and primary, contribute much smaller proportions by comparison.}
    \Description{The ratio of different road types.}
    \label{fig_road_count}
\end{figure}

\smallskip\noindent\textbf{Road network.} For each state, we construct a detailed road network graph based on data obtained from OpenStreetMap (OSM).\footnote{Notice that the original data from OSM divides one edge into several parts. Instead, we have preprocessed the raw data to combine the parts into a single edge. The volume statistic is low because the data includes many residential roads. See Figure \ref{fig_road_count} for an illustration of the distribution of different types of roads.} The road network encompasses five major categories of roadways: city streets, county roads, neighborhood streets, tract roads, and urbanized area roads, ensuring comprehensive spatial coverage across diverse geographic and administrative regions.

Each road segment in the network is represented as a directed edge defined by a start node and an end node, and is annotated with metadata including the road name, whether the segment is one-way, the road type, and the segment's physical length in meters. This representation captures both the structural and functional aspects of the road infrastructure.

Each node in the graph corresponds to a specific geospatial point, uniquely identified by a \texttt{node\_id}. It is associated with precise latitude and longitude coordinates. These nodes typically represent road intersections or endpoints, serving as key units for spatial reasoning and image alignment.
See Table \ref{tab_graph_dataset_statistic} for an overview of the network statistics.

There are 14 road types in total: living street, motorway, motorway link, primary, primary link, residential, road, secondary, secondary link, tertiary, tertiary link, trailhead, trunk, and trunk link. We report the probability of different road types in Figure~\ref{fig_road_count}. This processed road graph serves as the structural backbone for integrating visual and contextual data in our framework.

\smallskip\noindent\textbf{Satellite image data.} Building upon the extracted road network, we obtain high-resolution satellite images centered at each node, each of which typically corresponds to a road intersection. The geographic coordinates of these nodes guide the image collection process, ensuring tight spatial alignment between the visual and structural representations. For each state, we collect hundreds of thousands of satellite images, resulting in a large-scale visual dataset. We implement two strategies for image acquisition: (1) directly fetching static images using the Mapbox Static API, and (2) reconstructing large images by stitching together $25$ smaller tiles. Each final image has a resolution of $1024 \times 1024$ pixels and covers approximately $200 \, \mathrm{m} \times 200 \, \mathrm{m}$ of physical space, capturing rich visual cues and fine-grained road features.
In summary, for each node in the road network, we align one satellite image to that node across all six states.
Therefore, the number of satellite images equals the number of nodes.

\smallskip\noindent\textbf{Accident records.} As accident occurrence serves as the primary prediction objective in our framework, we obtain accident records from official sources, specifically the Departments of Transportation (DOT) of each U.S. state, to ensure high reliability and real-world relevance. We collected accident records dating back up to two decades, with states providing data ranging from the early 2000s to 2025. However, these records vary significantly across states in terms of file format, schema, field names, spatial encoding, and level of detail. To enable unified modeling and large-scale analysis, we perform extensive preprocessing and normalization to convert all state-specific datasets into a standardized format. Further, we align their axes with the satellite images.

The overall statistics of our dataset are shown in Table~\ref{tab_img_dataset_statistic}, including the accident period, the total number of accidents, and the total number of satellite images. More details about the dataset collection process can be found in Appendix~\ref{app_data_collection}.

\begin{table}[t!]
\centering
\caption{Overview of the collected dataset, including accident counts, period of accident records, and aligned satellite imagery across 6 U.S. states. The start and end years of the accident records are based on the latest released data from the Department of Transportation.}\label{tab_img_dataset_statistic}
{
\begin{tabular}{@{}lcccc@{}}
\toprule
 & {Start} & {End} & {$\#$ Labels} & {$\#$ Satellite Images}\\ 
\midrule
 Delaware        & $2009$ & $2024$ & 533,112 & 49,023 \\
 Massachusetts   & $2002$ & $2025$ & 5,165,834 & 285,942 \\
 Maryland        & $2015$ & $2023$ & 997,532 & 250,565 \\
 Nevada          & $2016$ & $2025$ & 376,252 & 121,392 \\
 Montana         & $2016$ & $2023$ & 140,011 & 145,525 \\
 Iowa            & $2014$ & $2024$ & 594,492 & 253,623 \\
\bottomrule
\end{tabular}}
\end{table}

\begin{remark}
    Compared with the most recent dataset on traffic accidents \cite{nippani2023graph}, our dataset now includes over one million satellite images. Additionally, we align the axes of the satellite images with the road networks to perform analysis with both data modalities.
\end{remark}

\subsection{Learning Multimodal Embeddings}

We use graph neural networks (GNNs) to learn network embeddings.
Consider a graph $G = (V, E)$, annotated with both node and edge features. GNNs learn node representations by iteratively aggregating information from local neighborhoods.
Let $x_i^{(k)}$ denote the representation of node $i$ at layer $k$, and $v_{i,j}$ the feature of edge $(i,j)$. The update rule of a message-passing layer is given by:
\begin{equation*}
    x_i^{(k+1)} = \phi \left( x_i^{(k)},\ h\left( \left\{ \psi(x_i^{(k)},\ x_j^{(k)},\ v_{i,j}) \mid j \in N(i) \right\} \right) \right),
\end{equation*}
where $N(i)$ is the set of neighbors of $i$, $h$ is a permutation-invariant aggregator (e.g., sum, mean), and $\phi$, $\psi$ are neural networks with nonlinearities.
This iterative procedure enables GNNs to capture neighborhood connections from the network.

Next, we use a vision model, such as Vision Transformer (ViT), to extract semantic information from images and encode each image into a visual embedding.

Given the GNN and image embeddings, we develop a multimodal learning framework to combine network embeddings and visual features. To incorporate visual semantics into the graph-based modeling process, we extract image features corresponding to each node and combine them with node embeddings generated by a graph neural network. We develop three combination methods to handle different types of features in traffic accident prediction.
\begin{itemize}%
    \item Basic fusion: we build a multilayer perceptron (MLP) to combine both embeddings.
    \item Gated fusion: we train a scalar gate value to determine the relative importance of the two modalities and then derive a weighted combination of both of them.
    \item Mixture of experts: we leverage multiple specialized networks, referred to as experts, each of which learns to combine features from different perspectives. We utilize the gated network that computes the probability distribution over experts for each node.
\end{itemize}
See Algorithm \ref{alg_moe} for the complete procedure of the implementation of the mixture of experts.
Further details of the fusion methods can be found in Appendix~\ref{app_implementation_details}. 

\begin{figure*}[t!]
    \centering
    \begin{subfigure}[b]{0.535\textwidth}
    \begin{minipage}[b]{0.49\textwidth}
        \centering
        \includegraphics[width=0.97\textwidth]{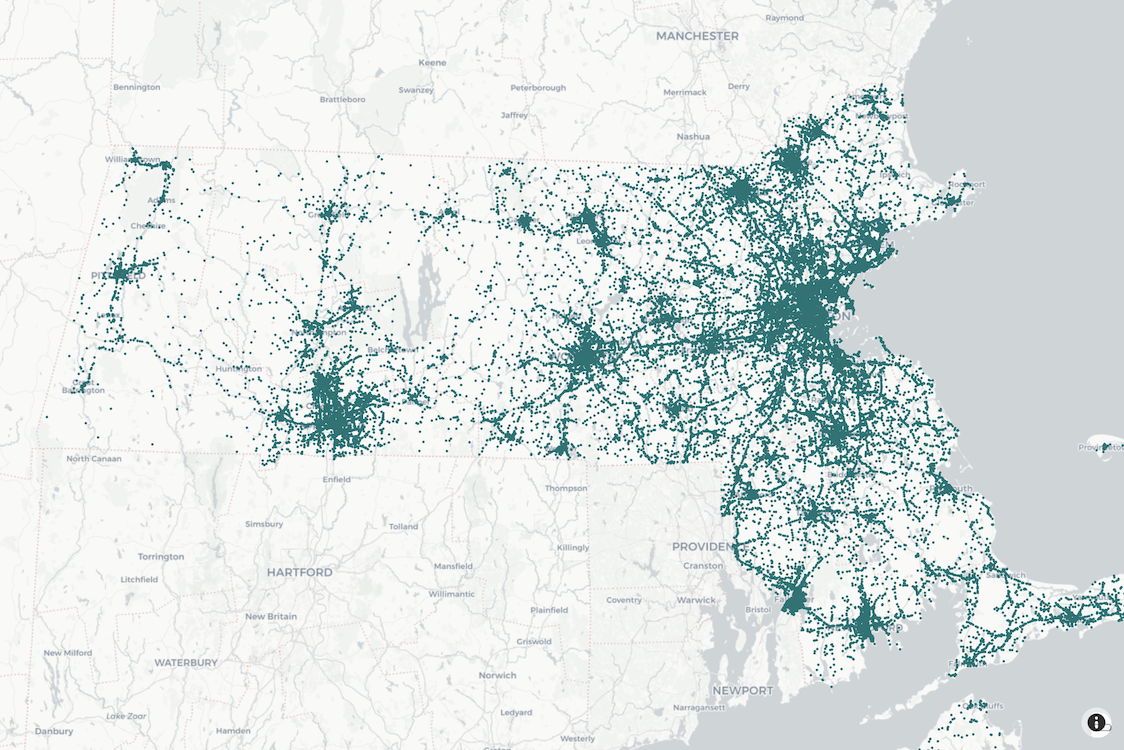}
        \caption{MA accidents in spring}\label{fig_ma_spr}
    \end{minipage}\hfill
    \begin{minipage}[b]{0.49\textwidth}
        \centering
        \includegraphics[width=0.97\textwidth]{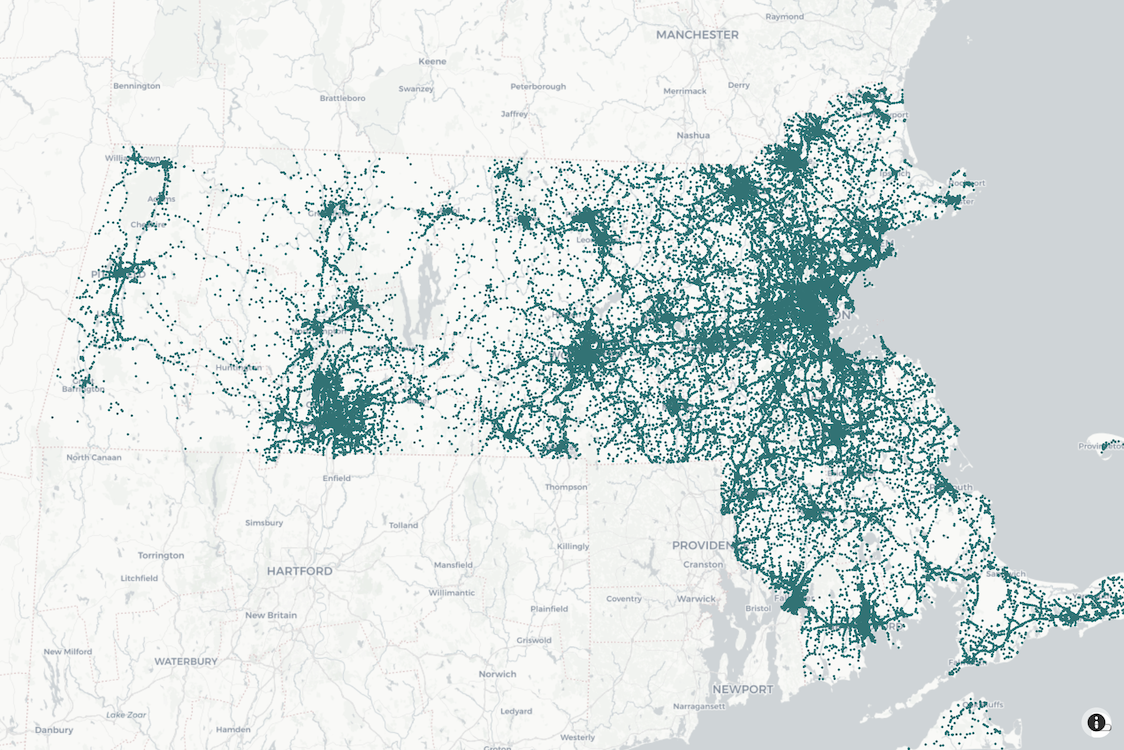}
        \caption{MA accidents in winter}\label{fig_ma_win}
    \end{minipage}
    \end{subfigure}
    \begin{subfigure}[b]{0.430\textwidth}
    \begin{minipage}[b]{0.49\textwidth}
        \centering
        \includegraphics[width=0.97\textwidth]{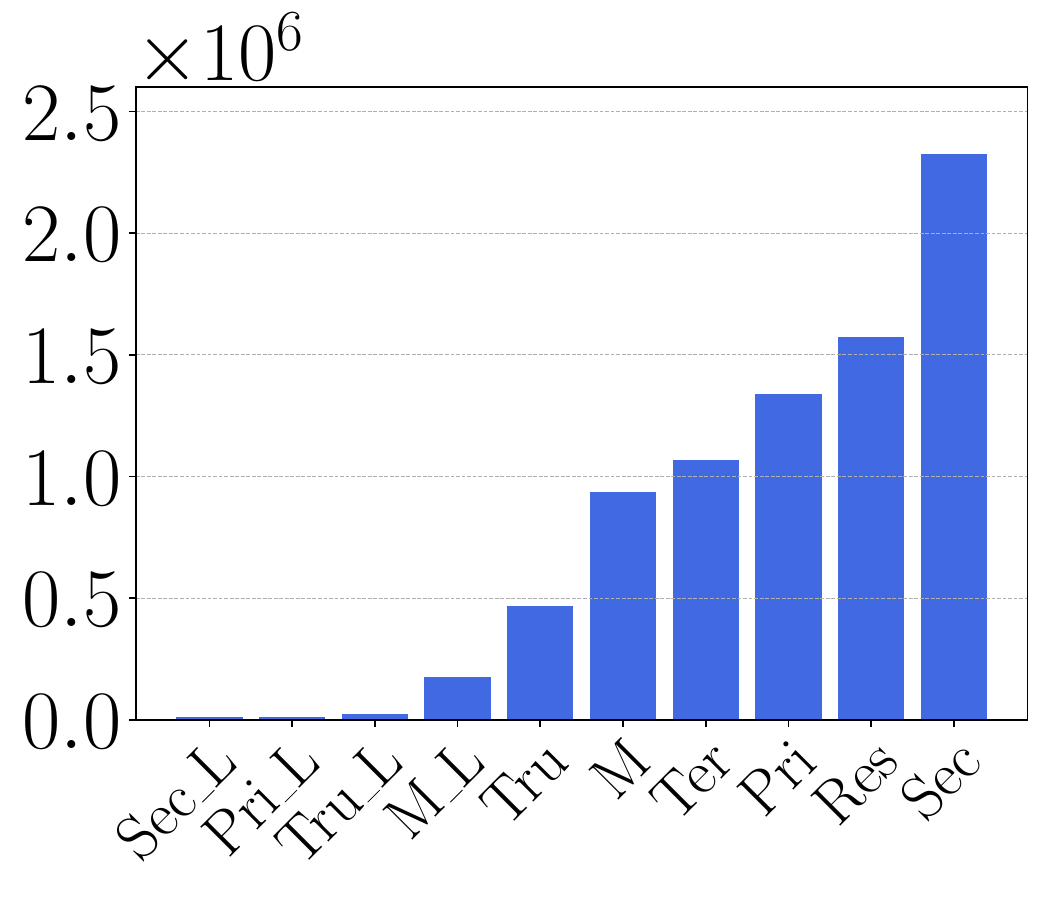}
        \caption{Total accident count}\label{fig_acc_total}
    \end{minipage}\hfill
    \begin{minipage}[b]{0.490\textwidth}
        \centering
        \includegraphics[width=0.97\textwidth]{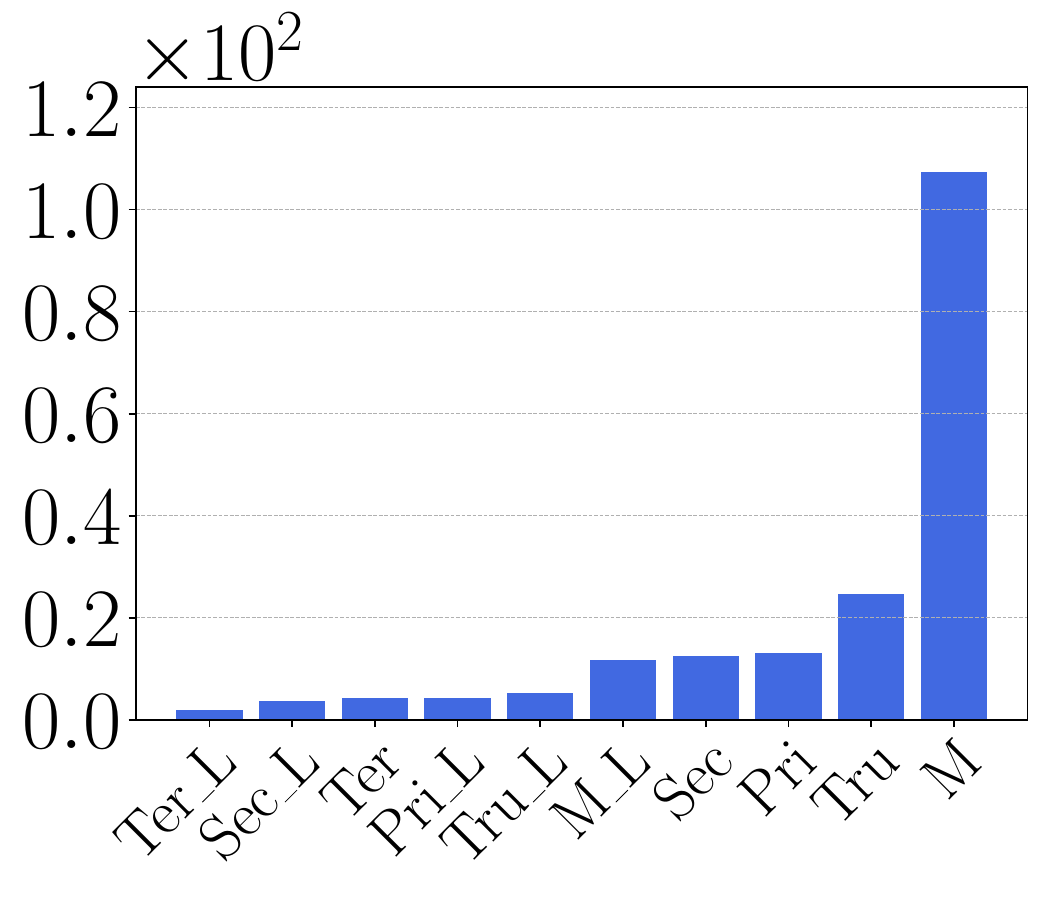}
        \caption{Average accident count}\label{fig_acc_avg}
    \end{minipage}
    \end{subfigure}
    \caption{Seasonal comparison of traffic accidents in Massachusetts.
    \ref{fig_ma_spr}, \ref{fig_ma_win}: Accident records in Massachusetts during spring and winter. It is evident that accident points are more densely distributed in winter, indicating a higher frequency of incidents likely due to adverse weather conditions.
    \ref{fig_acc_total}, \ref{fig_acc_avg}: Accident count of motorway (M), motorway link (M\_L), primary (Pri), primary link (Pri\_L), residential (Res), secondary (Sec), secondary link (Sec\_L), tertiary (Ter), tertiary link (Ter\_L), trunk (Tru), trunk link (Tru\_L), living street, road, and trailhead. 
    Figure \ref{fig_acc_total} gives the top-$10$ total count on different road types, while Figure \ref{fig_acc_avg} provides the top-$10$ average count on different road types.}\label{fig_season}
    \Description{Seasonal comparison of traffic accidents in Massachusetts and the accident count of different road types.}
\end{figure*}

\begin{algorithm}[t!]
\raggedright
\caption{Mixture-of-Experts Fusion}\label{alg_moe}
\textbf{Input:} Visual embedding $z$, graph embedding $x^{(\text{GNN})}$, number of experts $K$, and edge embedding $x_{\mathrm{edge}}$\\
\textbf{Require:} Expert networks $\{f_k\}_{k=1}^K$, gating network $f_{\text{gate}}$, prediction head $f_{\mathrm{pred}}$ \\
\textbf{Output:} Prediction $\hat{y}$
\begin{algorithmic}[1]
\State Concatenate visual and graph features: $\tilde{x} \gets \big[x^{(\text{GNN})} \parallel z\big]$
\For{$k = 1$ to $K$}
    \State Compute expert output: $e^{(k)} \gets f_k\big(\tilde{x}\big)$
\EndFor
\State Compute gating weights: $\lambda \gets \mathrm{softmax}\big(f_{\text{gate}}(\tilde{x})\big)$
\State Fuse expert outputs: $\tilde{e} \gets \sum_{k=1}^{K} \lambda^{(k)} \cdot e^{(k)}$
\State Predict output: $\hat{y} \gets f_{\mathrm{pred}}\big(\tilde{e}, x_{\mathrm{edge}}\big)$
\State \Return $\hat{y}$
\end{algorithmic}
\end{algorithm}

\subsection{Causal Estimation Using Multimodal Embeddings}

To provide a more fine-grained and context-aware estimation of causal effects, we perform matching-based causal analysis in the learned multimodal embedding space. Let $x_i \in \mathbb{R}^d$ denote the multimodal embedding of edge $i$, incorporating both visual and structural information. We define a binary treatment indicator $t_i = \mathds{1}$ for each edge $i$. The corresponding outcome variable $y_i$ represents the accident count or occurrence.

Our goal is to estimate the average treatment effect on the treated group (ATT). Consider the treated set $T = \{i \mid t_i = 1\}$, and a given treated edge $i \in T$, we compute the expected influence by
\[
\tau = \mathbb{E}\left[\hat{y}_{i,1} - \hat{y}_{i,0}\right],
\]
where $\hat{y}_{i,1}$ and $\hat{y}_{i,0}$ denote the potential outcomes under treatment and control. Since only one of these outcomes is observed, we approximate the missing counterfactuals using nearest-neighbor matching in the embedding space. For each treated sample, we identify its nearest neighbor $j$ from the control set,
\[
j = \arg\min_{k} \bignorms{x_i - x_k},
\]
and use its outcome $y_j$ as an estimate of $\hat{y}_{i,0}$. The ATT is then computed by
\[
\hat{\tau} = \frac{1}{|T|} \sum_{i \in T} (y_i - y_j).
\]

This embedding-based approach allows us to control for confounding factors encoded in the multimodal representation. It lets us compare treatment effects under similar road networks and weather conditions. To strengthen this analysis, we also estimate treatment effects using propensity score matching (PSM), which matches treated and control units with similar treatment probabilities. In addition, we use a doubly robust (DR) estimator that combines outcome prediction with propensity weighting to provide a more stable estimate. Full details of both procedures are reported in Appendix~\ref{app_experiment_details}.

\section{Experiments}\label{sec_experiment}

Given the constructed multimodal dataset and the proposed fusion methods, we evaluate the performance of various graph learning and multimodal strategies for predicting traffic accidents. We aim to assess the contribution of satellite imagery and the impact of different contextual factors on the performance.
We begin by detailing the experimental setup, including the baselines and the evaluation metrics. We then report the prediction results across different baselines and fusion strategies. Lastly, we conduct a causal estimation analysis to investigate the influence of key structural and environmental factors on accident occurrences.

\subsection{Experimental Setup}\label{sec_experimental_setup}

\subsubsection{Baselines} Recall that we focus on an edge-level link prediction task. As baselines, we include embedding methods, graph neural networks, and feature fusion approaches. 

First, we evaluate multilayer perceptrons (MLPs) using node features such as node degrees, betweenness centrality, road type, weather data, road length, and traffic volume. This setup assesses node features without incorporating the underlying network structure. We test embedding methods, including DeepWalk~\cite{perozzi2014deepwalk}, CLIP, and Vision Transformer, by appending a layer that concatenates the learned node embeddings with the original node features.

Second, we utilize common GNN architectures, including GraphSAGE, GCN, GIN, and Graph Transformer (Graphormer) with attention \cite{velickovic2017graph}. We use a spatial-temporal framework such as DCRNN~\cite{li2018diffusion}. 
We also consider supervised contrastive learning to improve the classification of positive and negative edges.

We then consider three approaches to integrating graph-based and vision-based features. The first approach basically employs a three-layer multilayer perceptron that jointly processes the concatenated graph and image representations. The second approach uses a gated fusion network that adaptively learns the contribution weights of each modality during training. The last one uses the MoE architecture to extract features from multiple perspectives.

\subsubsection{Implementations} 
For each state, we partition the available accident records into training, validation, and test sets based on temporal splits. Specifically, historical data up to a designated cutoff year is used to train the model, while accidents occurring after that year are reserved for evaluation. We focus on monthly accident prediction. This avoids future information leakage.
We consider both classification and regression tasks. For classification, we report the Area Under the Receiver Operating Characteristic curve (AUROC). For regression, we use the Mean Absolute Error (MAE) to measure the difference between the predicted and observed accident counts on each road segment.
We provide additional implementation details, including concrete hyperparameters, formal definitions of the fusion method, and formal definitions of the metrics, in Appendix~\ref{app_implementation_details}. The experimental details, including the omitted regression and classification task prediction result, the precision score, and the recall score, are shown in Appendix~\ref{app_omi_experiment_results}.

\renewcommand{\arraystretch}{0.85}
\begin{table*}[ht]
\centering
\caption{Comparing the results of GNNs, vision models, and multimodal fusion strategies. The performance is evaluated using the mean absolute error (MAE) and area under the ROC curve (AUROC) on the test split. A leave-one-out analysis is also provided. To account for variability, each experiment is repeated with three different random seeds, and we report the average results along with their standard deviations.}\label{tab_main_result}
{\small
\begin{tabular}{@{}l|l|cccccc@{}}
\toprule
\multirow{2}{*}{Category} 
& MAE ($\downarrow$)      & Delaware & Massachusetts & Maryland & Nevada & Montana & Iowa \\ 
& Average count & $4.59$ & $7.25$ & $1.72$ & $1.29$ & $0.40$ & $0.84$ \\
\midrule
\multirow{8}{*}{GNNs} 
& MLP         & $0.5\pm0.07$ & $0.6\pm0.10$ & $0.2\pm0.08$ & $0.3\pm0.03$ & $0.2\pm0.04$ & $0.3\pm0.09$ \\
& DeepWalk    & $0.2\pm0.02$ & $0.6\pm0.06$ & $0.3\pm0.02$ & $0.1\pm0.02$ & $0.3\pm0.01$ & $0.2\pm0.04$\\
& GraphSAGE   & $0.1\pm0.01$ & $0.2\pm0.01$ & $0.2\pm0.00$ & $0.1\pm0.02$ & $0.2\pm0.01$ & $0.1\pm0.02$ \\
& Graph Transformer & $0.1\pm0.03$ & $0.2\pm0.02$ & $0.2\pm0.01$ & $0.2\pm0.01$ & $0.2\pm0.04$ & $0.1\pm0.03$ \\
& DCRNN & $0.2\pm0.02$ & $0.3\pm0.01$ & $0.1\pm0.03$ & $0.1\pm0.03$ & $0.2\pm0.00$ & $0.1\pm0.03$ \\
& GCN         & $0.1\pm0.02$ & $0.8\pm0.20$ & $0.3\pm0.01$ & $0.2\pm0.01$ & $0.2\pm0.03$ & $0.2\pm0.01$ \\
& SupConGCN   & $0.2\pm0.06$ & $0.3\pm0.03$ & $0.3\pm0.03$ & $0.2\pm0.02$ & $0.2\pm0.03$ & $0.2\pm0.02$ \\
& GIN         & $0.1\pm0.02$ & $0.5\pm0.05$ & $0.3\pm0.01$ & $0.1\pm0.02$ & $0.1\pm0.03$ & $0.2\pm0.01$ \\
\midrule
\multirow{2}{*}{Vision models} 
& CLIP        & $0.3\pm0.11$ & $0.5\pm0.11$ & $0.3\pm0.02$ & $0.3\pm0.03$ & $0.1\pm0.03$ & $0.2\pm0.01$ \\
& Vision Transformer & $0.2\pm0.04$ & $0.6\pm0.03$ & $0.2\pm0.02$ & $0.2\pm0.04$ & $0.3\pm0.01$ & $0.2\pm0.01$ \\
\midrule
\multirow{3}{*}{Multimodal fusion} 
& GIN + Basic Fusion & $0.1\pm0.00$ & $0.3\pm0.01$ & $0.2\pm0.01$ & $0.2\pm0.01$ & $0.2\pm0.03$ & $0.2\pm0.01$ \\
& GIN + Gated Fusion & $0.1\pm0.01$ & $0.3\pm0.03$ & $0.2\pm0.01$ & $0.1\pm0.00$ & $0.3\pm0.04$ & $0.2\pm0.01$ \\
& GIN + MoE   & $\mathbf{0.1\pm0.00}$ & $\mathbf{0.3\pm0.02}$ & $\mathbf{0.2\pm0.01}$ & $\mathbf{0.1\pm0.00}$ & $\mathbf{0.2\pm0.01}$ & $\mathbf{0.1\pm0.02}$ \\
\bottomrule
\toprule
\multirow{2}{*}{Category} 
& AUROC ($\uparrow$)      & Delaware & Massachusetts & Maryland & Nevada & Montana & Iowa \\ 
& Positive rate & $0.32$ & $0.28$ & $0.24$ & $0.14$ & $0.10$ & $0.18$ \\
\midrule
\multirow{8}{*}{GNNs} 
& MLP         & $78.5\pm0.1$ & $64.2\pm0.1$ & $60.7\pm0.9$ & $80.8\pm0.3$ & $60.2\pm1.1$ & $66.7\pm0.4$ \\
& DeepWalk    & $82.1\pm0.4$ & $82.3\pm0.6$ & $86.3\pm0.2$ & $89.5\pm0.2$ & $78.1\pm1.5$ & $77.3\pm1.6$\\
& GraphSAGE   & $79.5\pm0.0$ & $80.3\pm0.6$ & $78.7\pm0.0$ & $81.8\pm0.7$ & $78.1\pm1.6$ & $77.4\pm0.2$ \\
& Graph Transformer & $85.8\pm1.5$ & $81.0\pm0.6$ & $77.9\pm3.0$ & $88.7\pm0.9$ & $79.2\pm2.5$ & $80.1\pm1.0$ \\
& DCRNN & $91.6\pm2.5$ & $84.3\pm0.1$ & $86.1\pm0.1$ & $85.5\pm0.7$ & $81.3\pm0.2$ & $80.4\pm1.6$ \\
& GCN         & $87.3\pm0.8$ & $86.2\pm0.2$ & $85.6\pm0.2$ & $89.4\pm0.2$ & $78.1\pm1.4$ & $77.2\pm1.6$ \\
& SupConGCN   & $86.4\pm1.8$ & $79.8\pm1.3$ & $83.8\pm1.2$ & $90.8\pm0.1$ & $83.5\pm0.6$ & $83.2\pm0.3$ \\
& GIN         & $91.6\pm0.7$ & $85.7\pm0.3$ & $85.9\pm0.3$ & $93.5\pm0.2$ & $79.6\pm0.6$ & $85.3\pm0.9$ \\
\midrule
\multirow{2}{*}{Vision models} 
& CLIP        & $86.3\pm0.5$ & $82.6\pm0.0$ & $85.6\pm0.1$ & $91.9\pm0.0$ & $79.6\pm0.8$ & $83.9\pm0.0$ \\
& Vision Transformer & $89.4\pm0.9$ & $82.1\pm0.5$ & $86.4\pm0.0$ & $93.2\pm0.1$ & $80.9\pm0.1$ & $85.9\pm0.2$ \\
\midrule
\multirow{3}{*}{Multimodal fusion} 
& GIN + Basic Fusion & $92.3\pm1.9$ & $84.8\pm0.3$ & $88.1\pm0.2$ & $93.5\pm0.2$ & $79.0\pm3.5$ & $86.7\pm0.4$ \\
& GIN + Gated Fusion & $92.8\pm2.8$ & $85.8\pm0.2$ & $88.5\pm0.3$ & $94.1\pm0.2$ & $86.3\pm1.1$ & $87.7\pm0.2$ \\
& GIN + MoE   & $\mathbf{96.4\pm1.0}$ & $\mathbf{88.1\pm0.4}$ & $\mathbf{88.7\pm0.1}$ & $\mathbf{94.5\pm0.2}$ & $\mathbf{87.5\pm1.6}$ & $\mathbf{88.0\pm0.4}$ \\
\midrule\midrule
\multirow{6}{*}{LOO analysis}
& GCN + Gated Fusion    & $88.5\pm0.2$ & $87.7\pm0.1$ & $89.3\pm0.0$ & $93.5\pm0.2$ & $83.7\pm0.7$ & $84.4\pm0.5$ \\
& w/o visual features   & $87.3\pm0.8$ & $86.2\pm0.2$ & $87.7\pm0.3$ & $89.4\pm0.2$ & $78.1\pm1.4$ & $77.2\pm1.6$ \\
& w/o weather features  & $82.1\pm1.7$ & $86.1\pm0.2$ & $88.9\pm0.3$ & $92.9\pm0.3$ & $82.3\pm0.2$ & $83.8\pm0.3$ \\
& w/o road network features            & $77.5\pm2.4$ & $82.3\pm1.0$ & $85.9\pm0.3$ & $93.2\pm0.3$ & $81.9\pm1.6$ & $83.9\pm0.5$ \\
& w/o traffic volume features        & $85.4\pm1.5$ & $84.2\pm0.4$ & $86.9\pm0.4$ & $93.1\pm0.1$ & $80.9\pm2.7$ & $82.0\pm1.2$ \\
& w/ speed limit features            & $89.5\pm1.4$ & $88.6\pm0.2$ & $89.7\pm0.3$ & $94.2\pm0.1$ & $84.2\pm0.2$ & $84.7\pm0.2$ \\
\bottomrule
\end{tabular}
}
\end{table*}
\renewcommand{\arraystretch}{0.95}

\subsection{Traffic Accident Prediction Results}

We illustrate the experimental results on multiple baselines shown in Section~\ref{sec_experimental_setup}. Below, we summarize two key insights that we believe hold broader relevance beyond the specific scope of our study.

We evaluate all models on six states and compare variants with and without visual inputs. GCN-based models show clear gains from incorporating satellite imagery. In particular, GCN plus Gated Fusion improves average AUROC by $3.8\%$ over the standard GCN, with a maximum gain of $7.2\%$ across states. GCN plus MoE yields a similar pattern, improving AUROC by an average of $3.9\%$ and up to $7.7\%$ in the best case.

For GIN-based models, visual features also provide consistent benefits. GIN plus Gated Fusion improves over the vanilla GIN by an average of $2.3\%$ AUROC, with a maximum gain of $6.7\%$. GIN plus MoE improves over the base GIN by an average of $3.6\%$, with the largest gain of $7.9\%$. Taken together, these results show that satellite imagery improves performance across all architectures and that the relative gains are greater for models with lower base expressivity.

\smallskip\noindent\textbf{Ablation studies.}
We conduct two ablation studies to understand the behavior of our frameworks. First, we evaluate the contribution of different feature types. Then, we examine the sensitivity to key hyperparameters.

For the feature ablation, we perform a leave-one-out (LOO) analysis on GCN plus Gated Fusion. We consider four feature categories: vision structure, weather observations, road layout, and traffic volumes. In each trial, we remove one category and measure the resulting change in AUROC. The results in Table~\ref{tab_main_result} show that dropping image features leads to a $3.5\%$ decrease. This confirms their strong predictive value. Removing weather features reduces AUROC by $1.8\%$. Excluding traffic volume produces a $2.4\%$ drop. Eliminating road network features causes the largest decrease of $3.7\%$. This highlights the central role of road layout. Adding speed limit as an extra feature yields a small but consistent improvement, suggesting that it provides complementary information.

For the hyperparameter study, we vary the number of GCN layers and the number of training epochs on the Nevada subset. Adjusting the depth from $2$ to $10$ layers changes AUROC by no more than $3\%$. This shows that the model remains stable across a wide range of depths. Varying the number of training epochs from $15$ to $50$ shows that performance plateaus around epoch $30$, suggesting that extended training offers limited benefit.

\begin{figure}[h!]
    \centering
    \includegraphics[width=0.5\columnwidth]{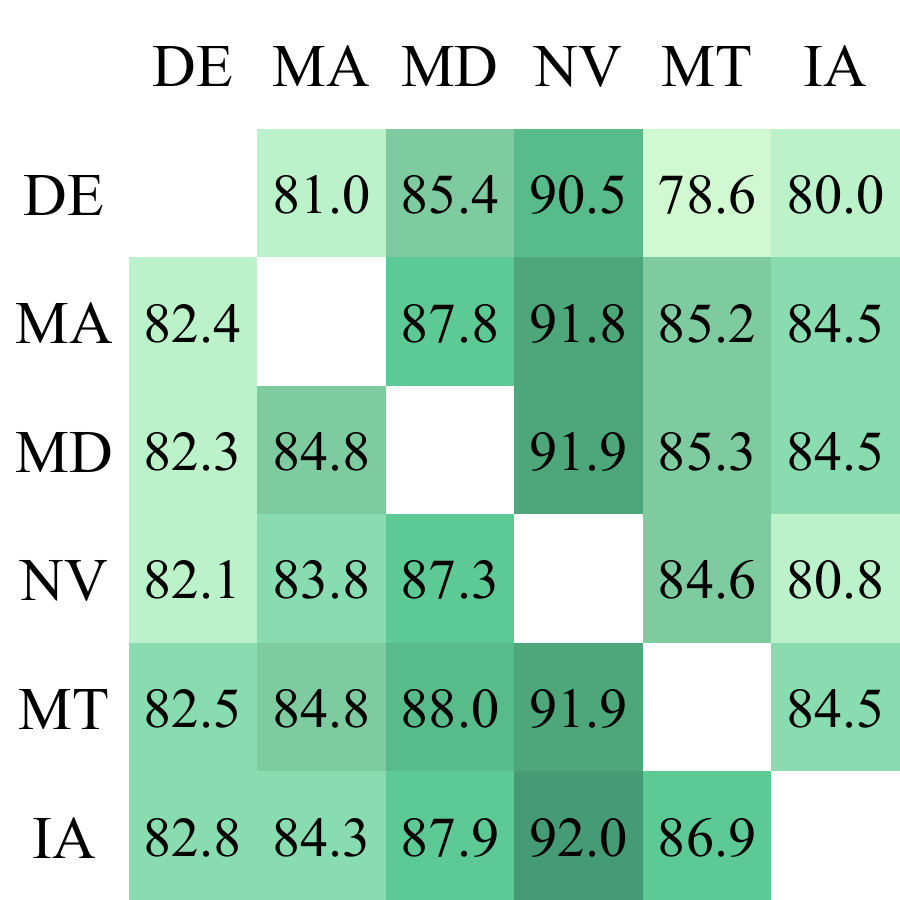}
    \vspace{-0.025in}
    \caption{Cross-state AUROC performance of the GIN + MoE model, computed over six states. Each entry shows the score when training on one state (rows) and testing on another (columns). Darker colors indicate better transferability.}
    \Description{The generalization ability.}
    \label{fig_cross_state}
\end{figure}

\smallskip\noindent\textbf{Cross-state transfer.} To evaluate how well the model transfers to new regions, we conduct a cross-state generalization experiment. In this setting, we train the GIN plus MoE model on one state and test it on another. The AUROC scores for all train–test pairs are shown in Figure~\ref{fig_cross_state}. The results reveal strong cross-state consistency. We find that the accident patterns in Maryland and Nevada are more consistent with the model's learned shared representations. It also indicates that both structural and visual features in these two states align well with those from other regions.
Further understanding these transfer patterns is an interesting question for future work \cite{wuunderstanding,liidentification,yang2025precise}.

\smallskip\noindent\textbf{Runtime.} The fusion structure may introduce extra computation, so we remove one GNN layer in the fusion modules to keep the cost low. We then empirically measure the computational time required to evaluate one month of data in each state. The results are reported in Table~\ref{tab_run_time}. It shows that the added overhead is small, and the runtime remains almost unchanged across all states.

\smallskip\noindent\textbf{Per-class performance.} To understand how the model behaves on different road types, we evaluate different fusion strategies on Delaware. We report the AUROC for major road classes in Table~\ref{tab_per_class}. Link roads are not included in this analysis. The results indicate that the model performs better on Residential, Road, and Living street, while relatively lower on Primary, Trunk, and Motorway.

\subsection{Causal Estimation Results}\label{sec_causal_analysis}

Building on our model's predictive performance, we analyze how contextual factors influence accident risk. We first conduct descriptive analyses by directly aggregating accident counts across conditions such as season, road type, and traffic volume to identify coarse-grained trends. To further validate these observations, we apply causal estimation techniques using learned multimodal embeddings to quantify the impact of specific factors. 

\smallskip\noindent\textbf{Seasonal variation.}
We first analyze seasonal variations in accident occurrences by dividing the year into four seasons: winter (December to February), spring (March to May), summer (June to August), and autumn (September to November). 

Most states show fewer accidents in spring than in winter. This pattern suggests that winter weather conditions increase the risk of accidents. Snow and ice are likely major factors. In warmer states such as Nevada, where winter conditions are mild, accident counts remain stable across seasons. This pattern is consistent with the causal estimates in Table~\ref{tab_causal_att}. We set winter as the treatment group and spring as the control group. The ATT values are higher in colder states such as Montana and Iowa, and much lower in Nevada. As an example, Figure~\ref{fig_season} shows the seasonal accident pattern in Massachusetts, where accident frequency peaks in winter.

\begin{table}[t]
\centering
\caption{Average prediction time for one month of traffic accident data. We compare the base GIN model with three fusion variants. The results show that adding fusion modules leads to only a small increase in computation.}\label{tab_run_time}
{
\small
\begin{tabular}{@{}lcccccc@{}}
\toprule
Runtime (s)   & DE & MA & MD & NV & MT & IA \\ 
\midrule
 GIN & $6.2$ & $53.1$ & $21.5$ & $9.5$ & $10.8$ & $20.5$ \\
 GIN + Basic & $6.4$ & $53.6$ & $22.2$ & $9.8$ & $12.5$ & $20.8$ \\
 GIN + GatedFusion & $6.5$ & $54.4$ & $22.4$ & $10.0$ & $12.6$ & $20.9$ \\
 GIN + MoE & $6.7$ & $54.5$ & $23.0$ & $10.2$ & $12.8$ & $21.7$ \\
\bottomrule
\end{tabular}}
\end{table}

\begin{table}[t]
\centering
\caption{AUROC on Delaware for different road types. We compare three fusion strategies of the GIN model. The results show that the model reaches higher accuracy on Residential, Road, and Living street, but performs worse on Primary, Trunk, and Motorway road types.}\label{tab_per_class}
{
\small
\begin{tabular}{@{}lcccccccccc@{}}
\toprule
AUROC (\%) & Residential & Tertiary & Secondary & Primary \\
\midrule
GIN+Basic  & $89.8$ & $83.3$ & $81.1$ & $78.0$\\
GIN+GatedFusion  & $90.4$ & $82.8$ & $81.3$ & $78.2$\\
GIN+MoE & $91.8$ & $83.3$ & $81.9$ & $79.0$\\
\midrule
  & Motorway & Trunk & Road & Living street \\
\midrule
GIN+Basic  & $78.6$ & $78.4$ & $95.2$ & $95.1$ \\
GIN+GatedFusion  & $77.4$ & $77.1$ & $92.7$ & $92.2$ \\
GIN+MoE & $80.0$ & $78.7$ & $94.4$ & $96.3$ \\
\bottomrule
\end{tabular}}
\end{table}

\smallskip\noindent\textbf{Road type.}
We also examine how different road types relate to accident occurrence. Road classification reflects structural design and traffic regulation levels, both of which can shape accident risk. The categories include living street, motorway, motorway link, primary, primary link, residential, road, secondary, secondary link, tertiary, tertiary link, trailhead, trunk, and trunk link. By comparing accident counts and frequencies across these categories, we aim to understand how differences in road infrastructure contribute to variations in traffic safety.

The results are shown in Figure~\ref{fig_season}. Motorways have the highest average accident frequency, and the gap compared to other road types is large. This is reasonable because vehicles on motorways travel at high speeds, reducing reaction time in case of unexpected events. Trunk roads show the second-highest accident frequency. They also carry fast-moving traffic, which poses similar risks of road accidents.

Secondary, residential, and primary roads have the highest total number of accidents. This is mainly due to their broad coverage in the road network. Their large presence leads to more cumulative accidents, while less common road types contribute fewer cases. Figure~\ref{fig_road_count} shows the distribution of road types and supports this overall pattern.

In our causal analysis, we treat motorways as the treatment group and other road types as the control group. The estimated ATT is $21.9\%$ shown in Table~\ref{tab_causal_att}. This suggests that motorway road segments contribute to a higher likelihood of accident occurrence after controlling for other factors.

\begin{figure}[t!]
    \centering
    \begin{subfigure}[b]{0.490\textwidth}
    \begin{minipage}[b]{0.49\textwidth}
        \centering
        \includegraphics[width=0.93\textwidth]{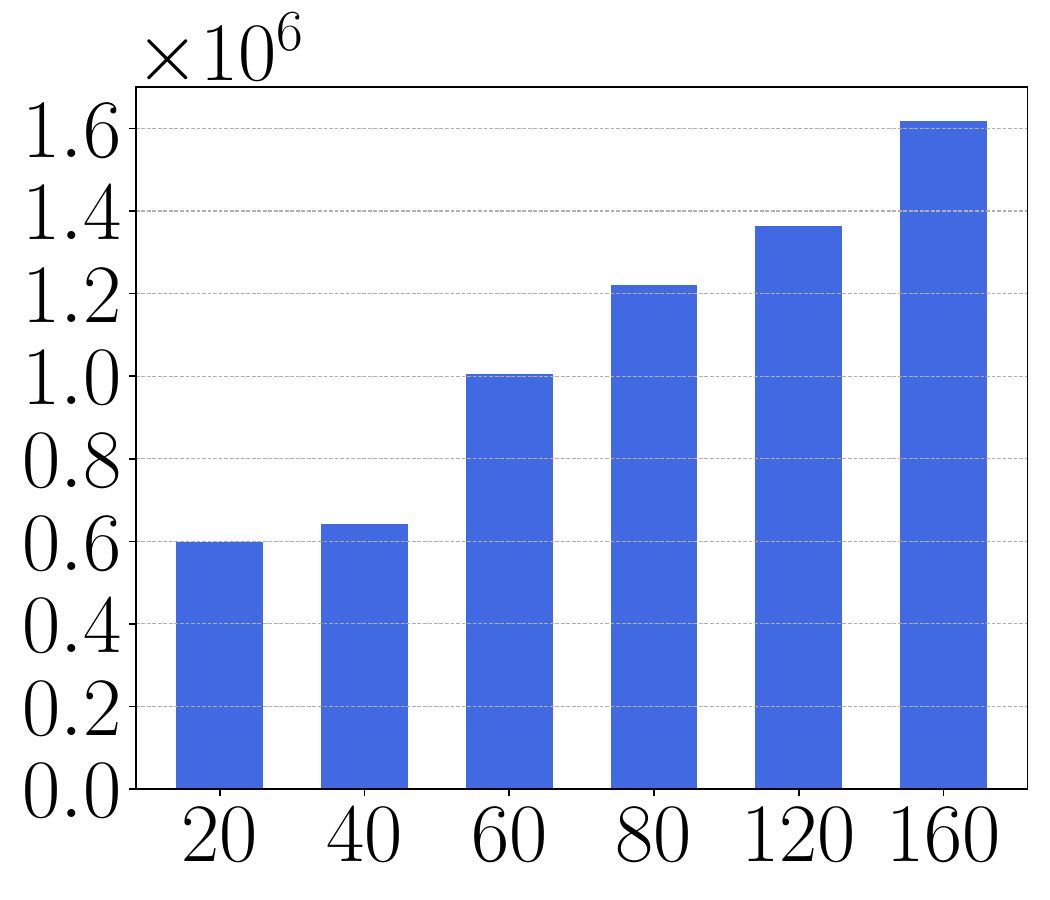}
        \caption{Precipitation}
        \label{fig_precipitation_accident}
    \end{minipage}\hfill
    \begin{minipage}[b]{0.490\textwidth}
        \centering
        \includegraphics[width=0.93\textwidth]{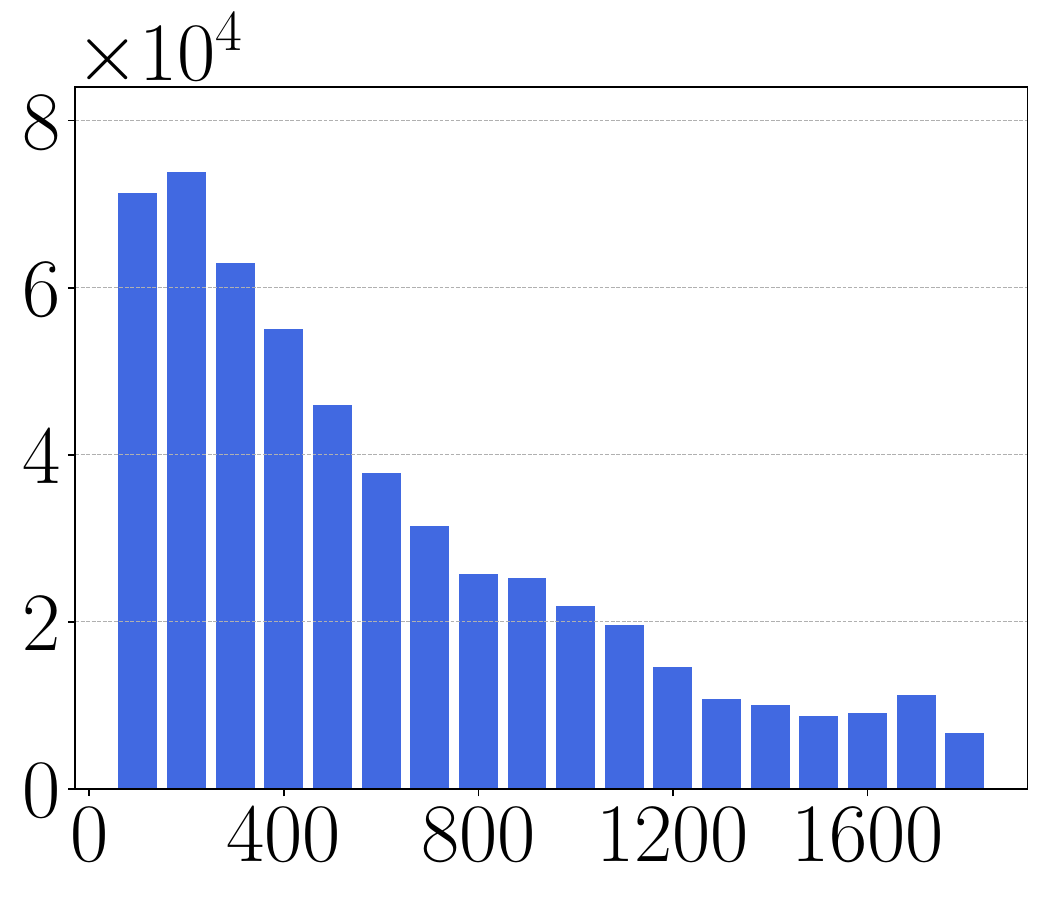}
        \caption{Traffic volume}
        \label{fig_aadt_accident}
    \end{minipage}
    \end{subfigure}
    \caption{Accident records with different ranges of precipitation and traffic volume.}\label{fig_precipitation_AADT}
    \Description{Accident records with different ranges of precipitation and traffic volume.}
\end{figure}

\smallskip\noindent\textbf{Precipitation.}
We align accident counts with precipitation levels and observe a clear rising trend in Figure~\ref{fig_precipitation_accident}. 

When precipitation is below $40$ mm, accident totals remain relatively low. As precipitation increases beyond $60$ mm, accident counts rise rapidly and reach more than $1.6 \times 10^6$ at $160$ mm. This trend suggests that heavy rainfall increases accident risk, likely due to reduced tire traction, lower visibility, and longer stopping distances. To further quantify this effect, we treat precipitation above $120$ mm as the treatment group and precipitation below $40$ mm as the control group. The causal estimation result in Table~\ref{tab_causal_att} shows an average ATT of $24.2\%$, supporting the positive effect of high precipitation on accident occurrence.

\smallskip\noindent\textbf{Traffic volume.}
We analyze the relationship between traffic volume and accident occurrence by examining the daily traffic volume corresponding to each recorded accident.

Our analysis reveals that as traffic volume increases up to approximately $200$ vehicles per day, the number of accidents rises accordingly. However, beyond this threshold, the accident rate declines gradually. This trend is reasonable: higher traffic volumes often lead to congestion, which reduces vehicle speeds and, consequently, lowers the likelihood of severe accidents. To understand the relationship between traffic volume and accident frequency, we visualize the distribution in Figure~\ref{fig_aadt_accident}.

\begin{table}[t]
\centering
\caption{Average treatment effect on the treated (ATT) among all six states. We analyze the effect of seasonal variation, road type, and precipitation. We vary for different years to compute the mean and standard deviations.}\label{tab_causal_att}
\resizebox{1.00\columnwidth}{!}
{
\begin{tabular}{@{}lcccccc@{}}
\toprule
ATT (\%)      & DE & MA & MD & NV & MT & IA \\ 
\midrule
Season      & $23.2_{\pm0.7}$ & $28.5_{\pm1.1}$ & $29.7_{\pm0.2}$ &$15.7_{\pm1.4}$  & $38.3_{\pm1.6}$ & $35.9_{\pm0.4}$ \\
Road type               & $25.8_{\pm1.2}$ & $23.3_{\pm1.1}$ & $19.2_{\pm1.7}$ & $22.4_{\pm0.8}$ & $21.5_{\pm2.7}$ & $19.1_{\pm1.4}$ \\
Precipitation          & $18.1_{\pm2.4}$ & $25.1_{\pm1.2}$ & $24.9_{\pm0.4}$ & $28.3_{\pm0.3}$ & $23.3_{\pm1.7}$ & $25.2_{\pm1.5}$ \\
\bottomrule
\end{tabular}}
\end{table}

\section{Related Work}\label{sec_related_work}
Traffic accident prediction is a dynamic task that has been explored using various foundational models. The critical role of spatial and temporal features in time-series forecasting is well established in prior work~\cite{li2018diffusion}. In addition, image-based features have proven valuable in related applications such as road attribute inference~\cite{he2020roadtagger,he2021riskmap}. We now discuss several areas of research most closely related to our contributions.

\paragraph{Generalization of graph neural networks and language models.} 
GraphGPT~\cite{tang2024graphgpt} shows that LLMs can be aligned with graph structure through instruction tuning. UrbanGPT~\cite{li2024urbangpt} further shows that instruction tuning can align spatio-temporal signals with LLMs, allowing LLMs to handle numeric time-series data and achieve strong zero-shot forecasting across urban tasks. Also, HiGPT~\cite{tang2024higpt} demonstrates that instruction-tuned LLMs can generalize across heterogeneous graphs by encoding node and edge semantics in natural language. 
\citet{ju2023generalization} analyze the generalization in graph neural networks using PAC-Bayesian bounds.
\citet{li2023boosting} use higher-order task affinities to boost multitask learning on graphs.
\citet{li2024scalable} use gradient-based estimation to accelerate the computation of task affinity. GradEx~\cite{li2024finetuning} introduces a first-order approach for scalable model fine-tuning. EnsembleLoRA~\cite{li2025efficient} uses GradEx in multitask learning. GradSel~\cite{zhang2025linear} extends to the in-context learning setting and evaluates the generalization ability of language models.

\begin{table*}[t]
    \centering
    \caption{A summary of our dataset and several existing datasets. Our dataset combines large-scale traffic accident records with aligned satellite imagery, whereas prior datasets either lack imagery or have limited volume.}\label{tab_baseline_comparison}
    {\small
    \begin{tabular}{lcccccc|c}
    \toprule
    Dataset       & Year   & Volume & Spatial & Time Series & Tabular & Satellite Images & Category \\ 
    \midrule
    METR-LA~\cite{li2018diffusion} & $2018$ & $6M+$ & \cmark & \cmark & \cmark & \xmark & Traffic Forecast\\
    RoadTracer~\cite{bastani2018roadtracer} & $2018$ & $300$ & \xmark & \xmark & \xmark & \cmark & Semantic Segmentation\\
    SEVIR~\cite{veillette2020sevir} & $2020$ & $10K+$ & \cmark & \cmark & \xmark & \cmark & Weather Forcast\\
    ML4Roadsafety~\cite{nippani2023graph} & $2023$ & $9M+$ & \cmark & \cmark & \cmark & \xmark & Traffic Accident Analysis\\
    CrashFormer~\cite{karimi2023crashformer} & $2023$ & $6M+$ & \cmark & \cmark & \cmark & \xmark & Traffic Accident Analysis\\
    OAM-TCD~\cite{veitch2024oam} & $2024$ & $5K+$ & \cmark & \cmark & \xmark & \cmark & Semantic Segmentation\\
    SolarCube~\cite{li2024solarcube} & $2024$ & $600K+$ & \cmark & \cmark & \xmark & \cmark & Weather Forcast\\
    FT-AED~\cite{coursey2024ft} & $2024$ & $3M+$ & \cmark & \cmark & \cmark & \xmark & Traffic Detection\\
    Tumtraffic-QA~\cite{zhou2025tumtraffic} & $2025$ & $90K+$ & \cmark & \cmark & \xmark & \cmark & Traffic Detection\\
    \midrule
    \acronym{} (This work) & $2025$ & $10M+$ & \cmark & \cmark & \cmark & \cmark & Traffic Accident Analysis\\
    \bottomrule
    \end{tabular}}
    \end{table*}

\paragraph{Spatiotemporal mining for traffic prediction.}
The importance of jointly modeling spatial and temporal dependencies over graph-structured data for time series prediction is well recognized. Recent advances have focused on developing powerful deep learning architectures to capture these complex patterns in structured data such as traffic flow and road graphs. For instance, DCRNN~\cite{li2018diffusion} treats traffic flow as a diffusion process on directed graphs and incorporates this into a sequence-to-sequence recurrent architecture with scheduled sampling. To improve training efficiency, STGCN~\cite{yu2018spatio} eliminates recurrent units by stacking graph and temporal convolutions into a fully convolutional framework. TEMPO~\cite{cao2024tempo} introduces a novel prompt-based generative pre-training framework that integrates time series decomposition with transformer architectures. MG-TAR~\cite{trirat2023mg} leverages dangerous driving statistics as near-miss indicators within a multi-view GNN for citywide risk prediction. %

The advent of high-resolution remote sensing has opened up new possibilities for analyzing the physical state of road networks at scale~\cite{zhou2024allclear, veitch2024oam, li2024solarcube, bastani2018roadtracer}. Satellite imagery provides visual cues that are strongly correlated with accident risk, but are difficult to capture in traditional datasets. Researchers have used this data for a range of safety-related tasks. Some approaches use imagery to assess safety proxies, such as monitoring road surface conditions or automatically detecting infrastructure like pedestrian crossings. 

\paragraph{Multimodal road network analysis.}
Our approach builds on prior research that combines visual models with graph-based learning for road network analysis. The RoadTagger system~\cite{he2020roadtagger} introduced an end-to-end architecture that combines a CNN with a GNN to infer road attributes, such as lane counts, from satellite imagery. By propagating visual features along the road graph, the GNN overcomes the limited receptive field of a CNN, enabling robust inference in the presence of occlusions. Subsequently, the work on Inferring High-Resolution Traffic Accident Risk Maps~\cite{he2021inferring} developed a deep learning framework that fuses multiple data sources, including satellite imagery, GPS trajectories, road maps, and historical accident data, to generate fine-grained (5m resolution) risk maps. This work demonstrated how to overcome data sparsity challenges by leveraging a rich, multimodal context.
The challenge of integrating heterogeneous data modalities on a graph is also studied in multimodal graph learning~\cite{ektefaie2023multimodal}. Our work contributes to this area in the context of road safety modeling.

Inspired by progress in computer vision, several large-scale satellite imagery datasets have recently been introduced~\cite{zhou2024allclear, veitch2024oam, li2024solarcube, bastani2018roadtracer}. OAM-TCD~\cite{veitch2024oam} focuses on high-resolution tree crown delineation with global diversity, while SolarCube~\cite{li2024solarcube} integrates satellite and ground data for solar radiation forecasting across continents. RoadTracer~\cite{bastani2018roadtracer} addresses road network extraction via iterative graph construction from aerial images.

\section{Discussion}

The satellite images in our dataset capture road characteristics that change very slowly, such as lane width, curvature, and intersection density. These features remain stable over long periods, so small differences between the imagery date and the accident records do not introduce clear sources of bias. The dynamic elements of accident risk come from other modalities. Weather observations and time-varying traffic volume (AADT) provide signals describing short-term changes in road conditions.

This combination allows us to model relative risk across the network in a stable and interpretable way. Our goal is not to forecast the exact time or location of an accident. The predicted values instead represent the likelihood or hazard level of each road segment. They offer a practical measure of traffic risk that can support driver awareness, road maintenance decisions, and safety planning.

Seen from this angle, our dataset opens up several directions for future work. It provides a strong benchmark for multimodal learning methods that operate on both graph and image data. It also demonstrates that visual cues, structural features, and traffic signals can work together to reveal fine-grained patterns in accident risk. This creates space to develop stronger fusion models, explore more detailed temporal patterns, and build tools that help agencies use data-driven insights for safety interventions.

Our causal analysis relies on the common assumption that treatment and control units are independent. In road networks, this assumption may not hold exactly, since neighboring nodes and edges are spatially connected and can influence each other. Therefore, the estimated effects should be interpreted as approximate associations under this modeling assumption.

\smallskip\noindent\textbf{The \acronym{} package.} To support future research, we release our dataset on Huggingface along with an easy-to-use Python package for streamlined access. The full dataset is available via the \textsc{Datasets} library and includes road network graphs, accident records, weather data, traffic volume statistics, and satellite image embeddings. The embeddings are generated using Vision Transformer and CLIP models. We also include a set of satellite images for all states in the package.

With a single line of code, users can load the complete dataset. To retrieve data for a specific state, one simply specifies the state name, and the package will automatically download, cache, and return the corresponding dataset object. We also provide functionality to extract accident records and features for any given month. Additionally, the package includes a trainer module for training and evaluating baseline models in our framework, encompassing basic GNN models, spatiotemporal models, and fusion models. We add some examples of the usage in Appendix~\ref{app_code_example}.

\section{Conclusion}

We present a large, up-to-date multimodal traffic dataset comprising road networks, weather data, traffic volumes, accident records, and satellite imagery from six U.S. states. The dataset is designed to support research on traffic accident prediction and to provide a comprehensive view of the factors that shape road safety. It brings together long-term records from multiple public sources and aligns them at the road-segment level, making it suitable for both predictive modeling and causal analysis.

On top of this dataset, we build a GNN-based framework that integrates visual, structural, and temporal features. The model combines information from all modalities and achieves strong predictive performance. The best baseline reaches an average AUROC of $90.1\%$ across states, showing that multimodal representations are effective for accident prediction. Our analysis also includes causal estimation and feature ablation. These studies help clarify how different contextual factors contribute to accident risk and how each feature type affects model performance.

\section*{Acknowledgments}

Thanks to Yan Liu, Abhinav Nippani, Dongyue Li, and Ruoxuan Xiong for discussions and feedback on this project.
We appreciate the anonymous referees for their comments and suggestions.

This project is supported by Northeastern University's Transforming Interdisciplinary Experiential Research (TIER) $1$ program.
The work of Z. Zhang, M. Duan, and H. R. Zhang is additionally supported in part by NSF award IIS-2412008.
The views and opinions expressed in this paper are solely those of the authors and do not necessarily reflect those of the National Science Foundation.

%% file: appendix.tex
\section{Data Collection Procedure}\label{app_data_collection}

In this section, we provide more details on our data collection procedure, including how to construct the road network, align the locations of accident records, normalize traffic volumes across different formats, and obtain weather features. Then, we summarize all the features we collected.

\subsection{Road Networks}\label{app_road_net}
We construct the road network using the OSMnx Street Network Dataverse dataset. For each state, we load street networks at multiple administrative levels, including cities, counties, neighborhoods, census tracts, and urbanized areas. These individual networks are then concatenated to form a comprehensive statewide road graph.

From the raw data, we extract a list of nodes and edges. For each node, we retain the node ID along with its geographic coordinates (latitude and longitude). For each edge, we record the start node ID, end node ID, whether it is one-way, the corresponding road type, and the road length.

\subsection{Traffic Accident Records}
We collect traffic accident data from each state's DOT, which is provided in different formats.

In our analysis, we focus on extracting latitude and longitude from these records. When geographic coordinates are not directly available in the original dataset, we use the textual address descriptions, such as street names or intersections, and query the corresponding latitude and longitude using the Google Maps API. This step ensures that all accident locations are consistently represented in a geospatial format, which is critical for downstream spatial analysis and visualization.

Next, we align each accident record to a specific road segment in the network. Let the geocoded accident location be denoted as $c \in \mathbb{R}^2$. For each edge $e \in E$, associated with endpoints $e_a$ and $e_b$, we define $D(\cdot, \cdot)$ as the Euclidean distance function. To assign the accident to the most plausible road segment, we adopt the following heuristic:
\[ e_{\text{acc}} = \arg\max_{e \in E} \left( D(e_a, e_b) - \left( D(e_a, c) + D(e_b, c) \right) \right). \]
This formulation prioritizes edges for which the accident point lies closest to the segment span between $e_a$ and $e_b$.

We illustrate the traffic accident records in Figure \ref{fig_accident_count_monthly} and Figure \ref{fig_season_omitted} below.
\begin{figure}[h!]
    \centering
    \begin{subfigure}[b]{0.490\textwidth}
    \begin{minipage}[b]{0.49\textwidth}
        \centering
        \includegraphics[width=0.85\textwidth]{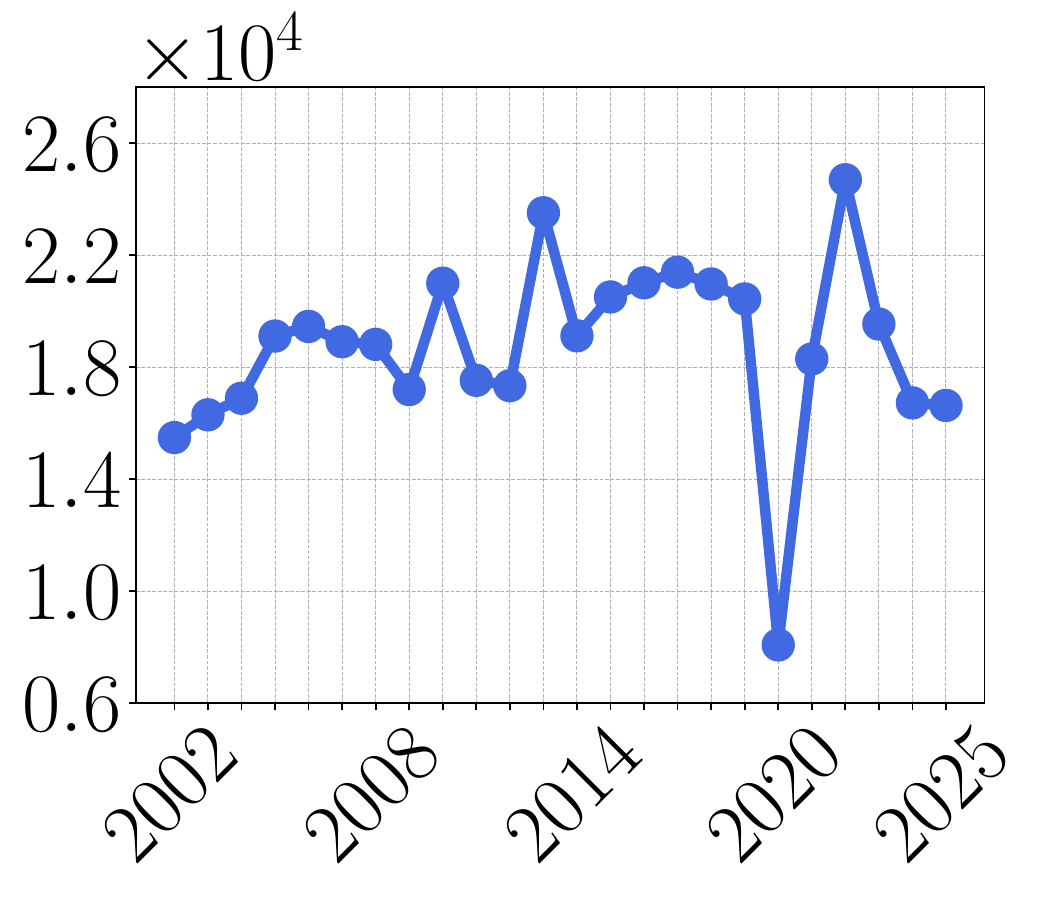}
        \caption{Massachusetts}
    \end{minipage}\hfill
    \begin{minipage}[b]{0.490\textwidth}
        \centering
        \includegraphics[width=0.85\textwidth]{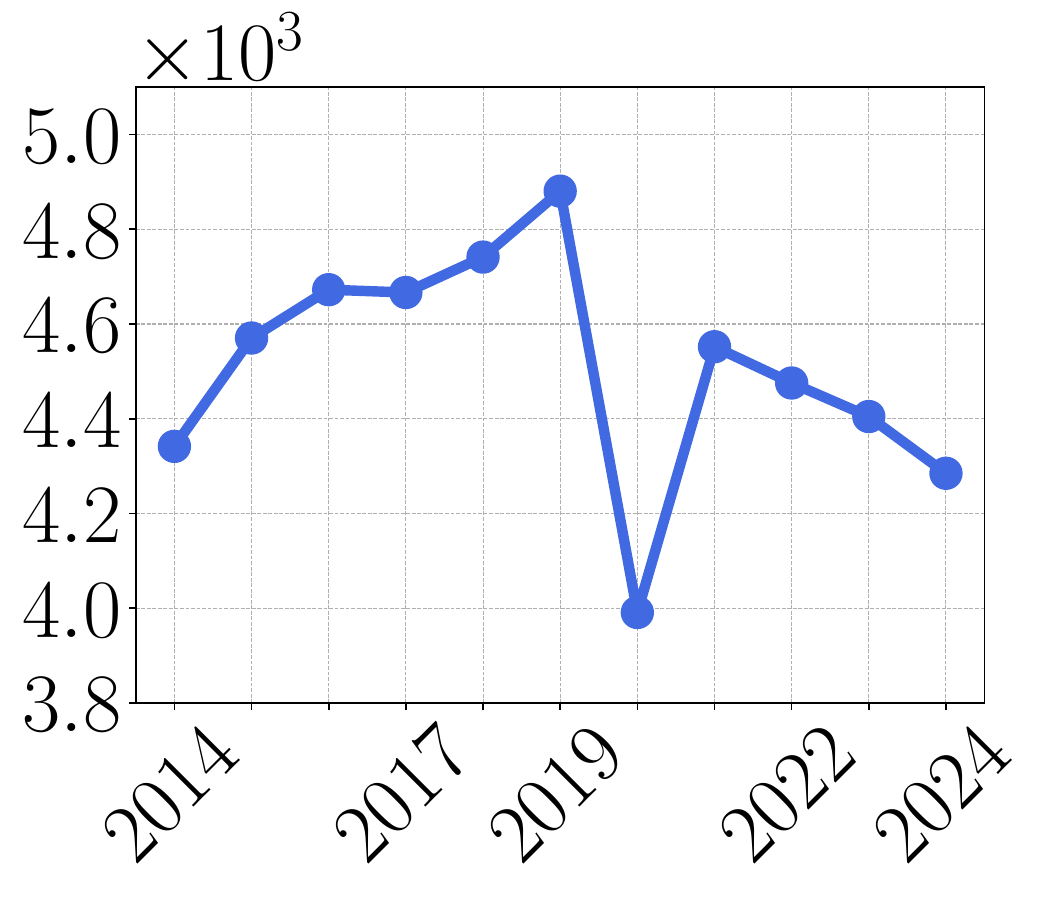}
        \caption{Iowa}
    \end{minipage}
        \begin{minipage}[b]{0.49\textwidth}
        \centering
        \includegraphics[width=0.85\textwidth]{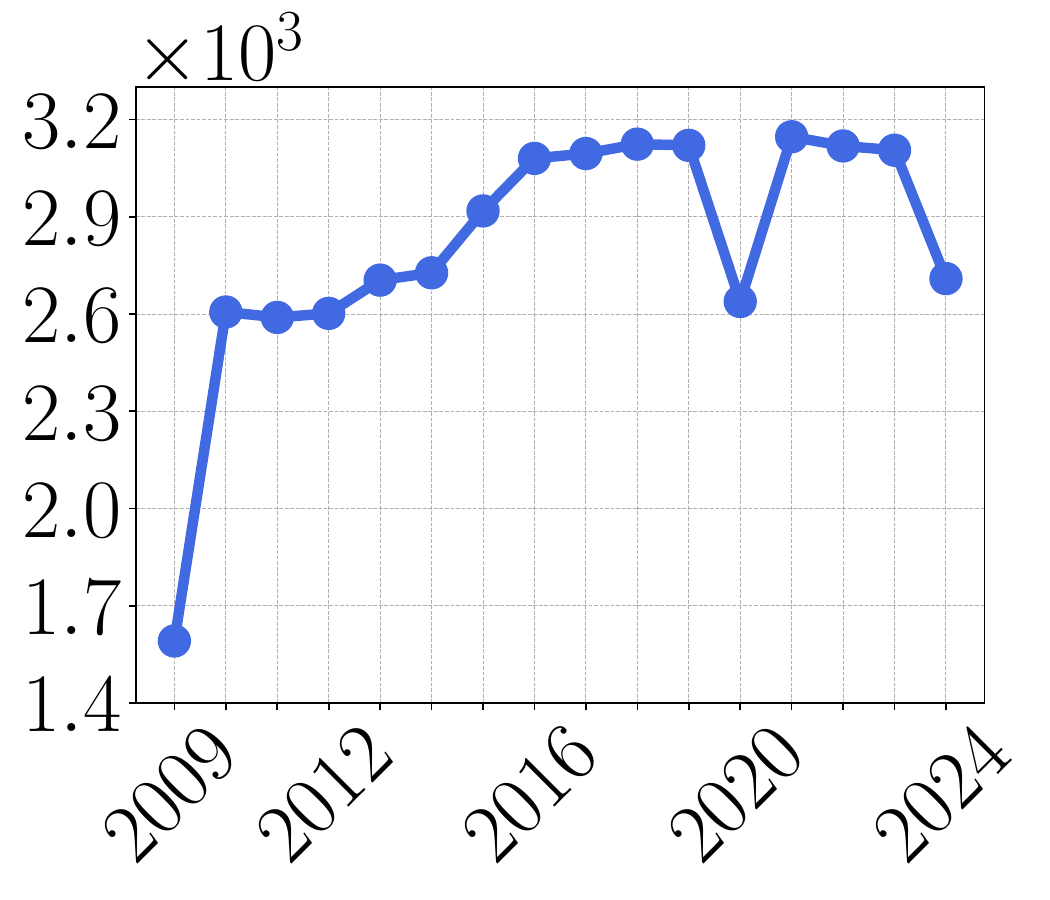}
        \caption{Delaware}
    \end{minipage}\hfill
    \begin{minipage}[b]{0.490\textwidth}
        \centering
        \includegraphics[width=0.85\textwidth]{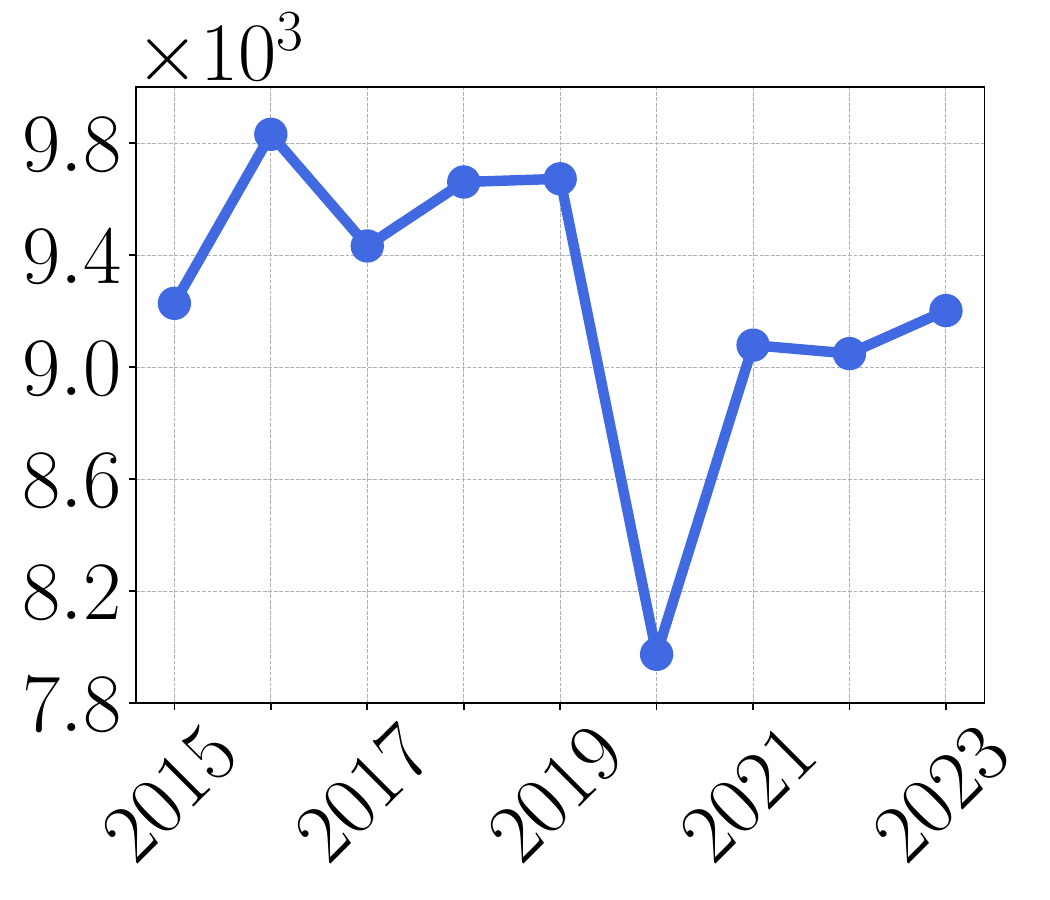}
        \caption{Maryland}
    \end{minipage}
    \end{subfigure}
    \caption{The average number of accidents per month for each year in Massachusetts, Iowa, Delaware, and Maryland. The sharp drop in 2020 is due to the impact of COVID-19.}
    \Description{The average number of accidents per month for each year in Massachusetts, Iowa, Delaware, and Maryland. The sharp drop in 2020 is due to the impact of COVID-19.}
    \label{fig_accident_count_monthly}
\end{figure}
\begin{figure}[h]
    \centering
    \begin{subfigure}[b]{0.490\textwidth}
    \begin{minipage}[b]{0.49\textwidth}
        \centering
        \includegraphics[width=0.97\textwidth]{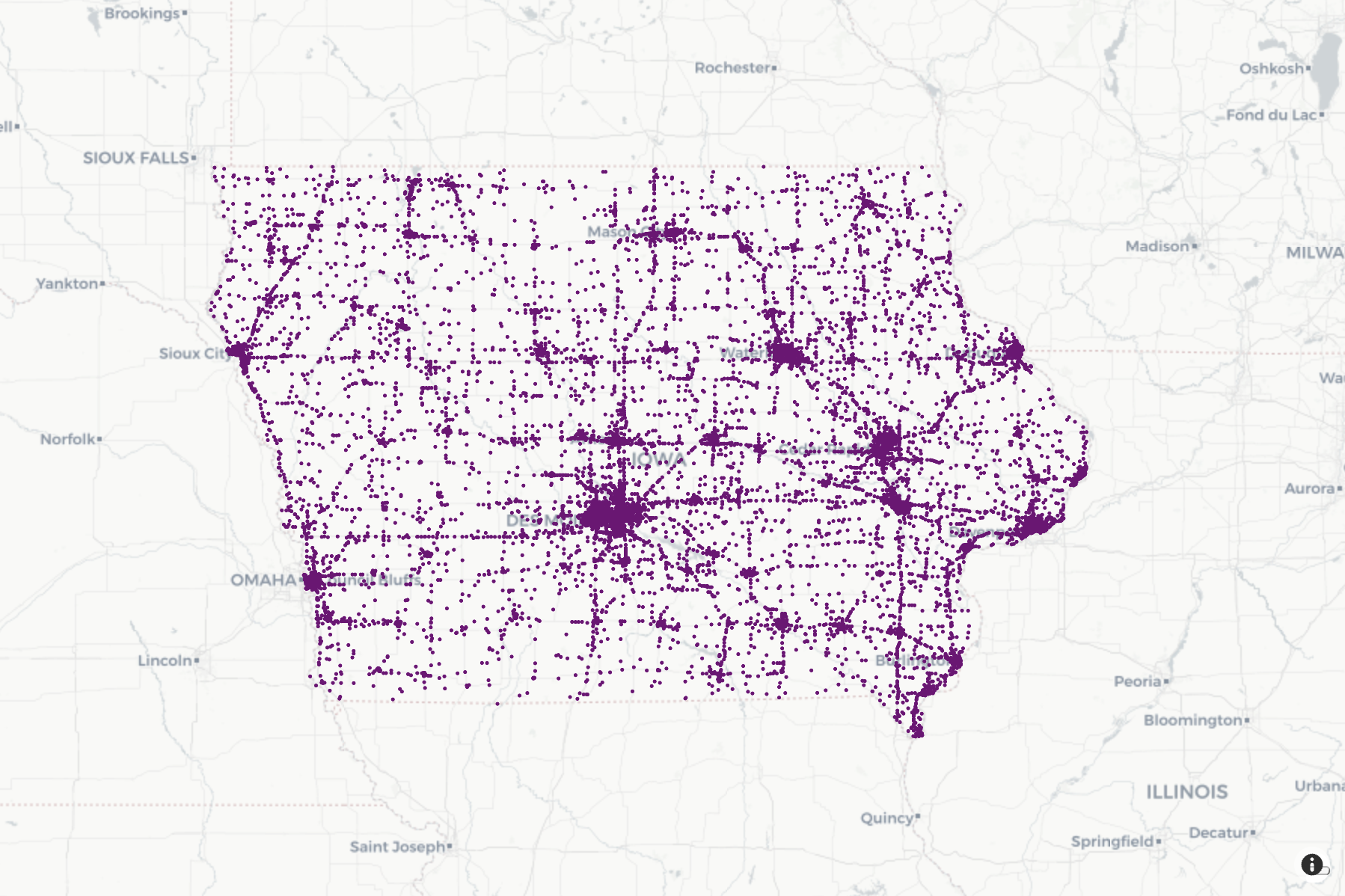}
        \caption{IA accidents in spring}
    \end{minipage}\hfill
    \begin{minipage}[b]{0.490\textwidth}
        \centering
        \includegraphics[width=0.97\textwidth]{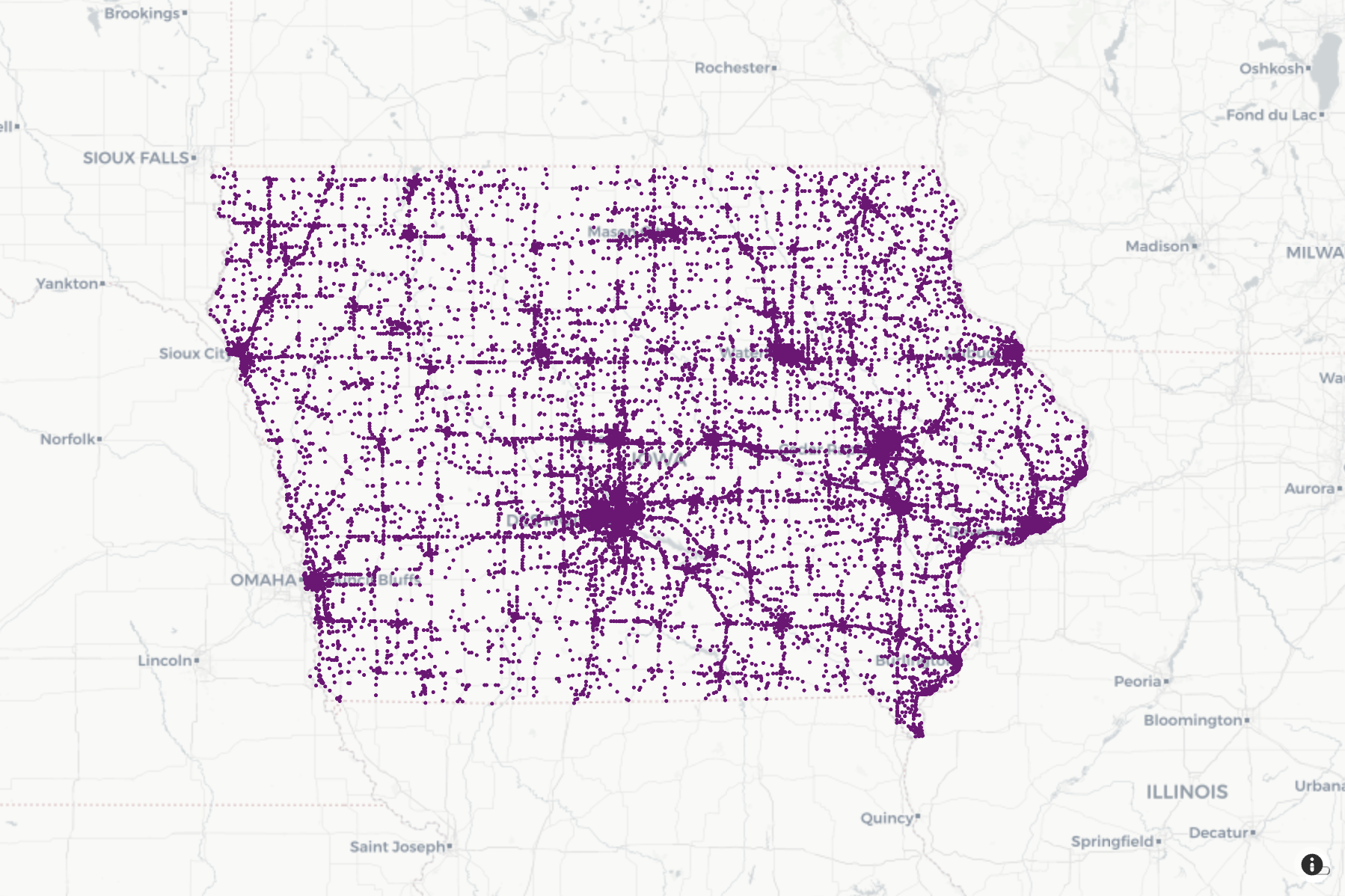}
        \caption{IA accidents in winter}
    \end{minipage}
    \end{subfigure}
    \caption{Seasonal comparison of traffic accidents in Iowa during spring and winter, respectively, showing that accidents in winter occur more than in spring.}
    \Description{Seasonal comparison of traffic accidents in Iowa.}
    \label{fig_season_omitted}
\end{figure}

\subsection{Traffic Volume}

We obtain traffic volume data from official state-level transportation departments and use it as a dynamic edge-level feature in our multimodal framework. 

Some states provide clean CSV exports containing longitude and latitude fields, while others (such as Massachusetts and Montana) publish data in more complex formats, including KML and GeoJSON, with inconsistent column names and nested spatial structures. In these cases, we developed custom parsers for each state and manually mapped the relevant fields to extract usable geospatial information. Additionally, the temporal resolution varies by state, and in some cases, the data spans dozens of measurement points per segment, requiring iterative merging and normalization. Often, the same road segment is represented by multiple sensor points along its path, with no explicit intersection or highway labeling, so we heuristically aggregated these measurements to assign a single traffic volume value to each edge.

\begin{figure}[h]
    \centering
    \begin{subfigure}[b]{0.490\textwidth}
    \begin{minipage}[b]{0.49\textwidth}
        \centering
        \includegraphics[width=0.85\textwidth]{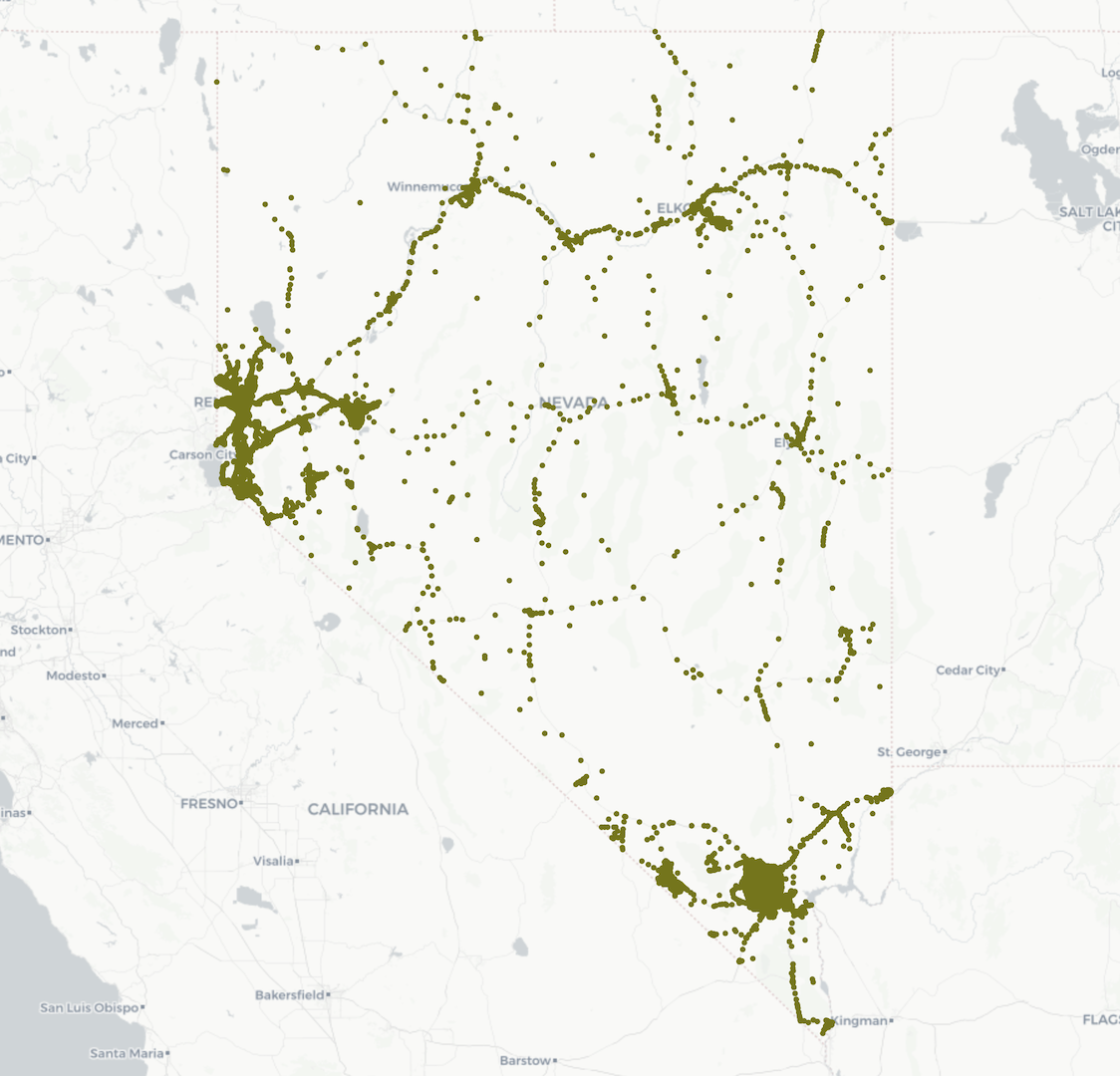}
        \caption{Accidents}
    \end{minipage}\hfill
    \begin{minipage}[b]{0.490\textwidth}
        \centering
        \includegraphics[width=0.85\textwidth]{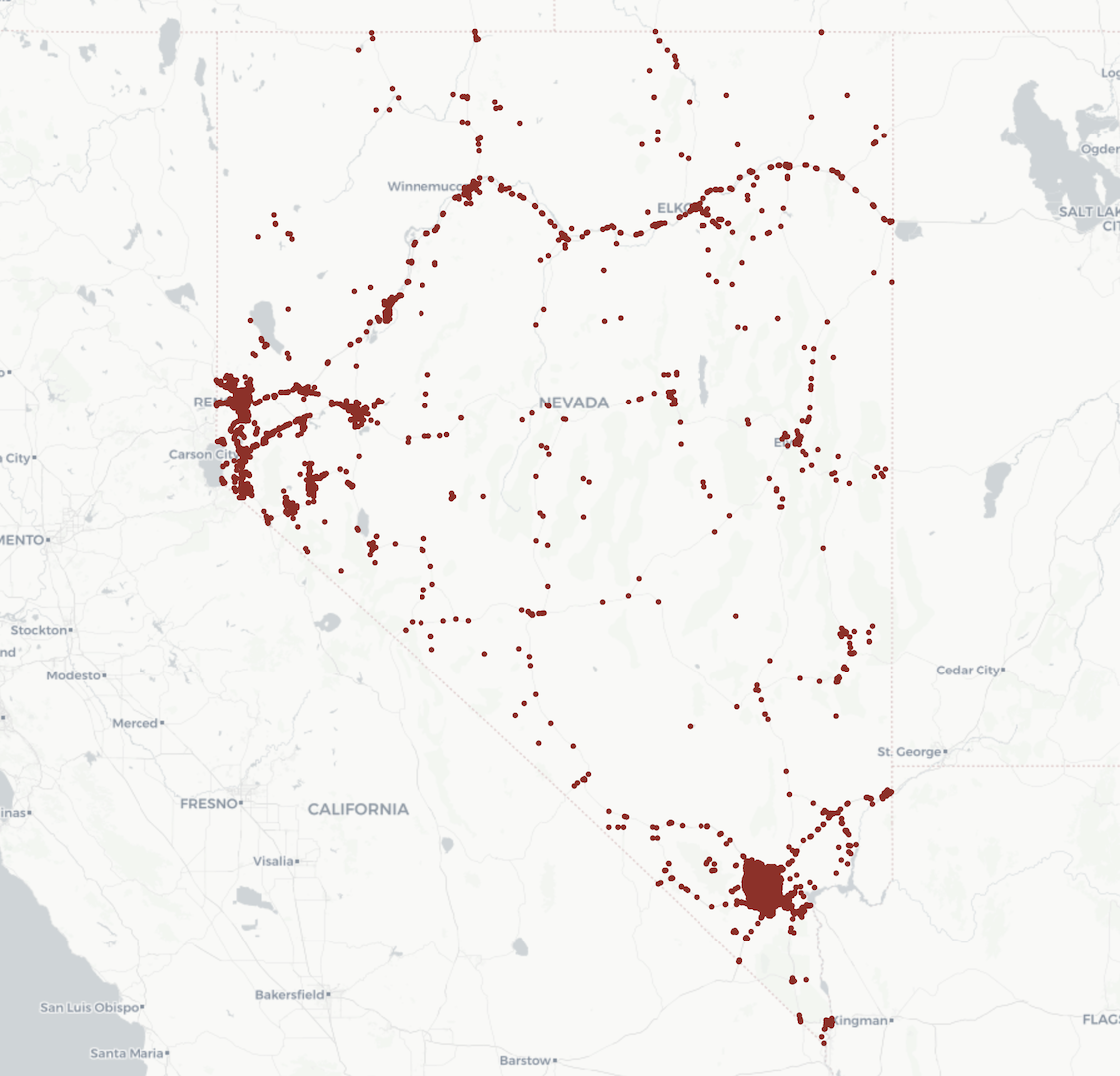}
        \caption{Traffic volume records}
    \end{minipage}
    \end{subfigure}
    \caption{The distribution of accidents and traffic volume monitor coverage in Nevada. (a) marks the historical accident records across the state, and (b) shows the road segments with recorded traffic volumes. The traffic volume records cover nearly all road segments where accidents have occurred.} 
    \Description{Distribution of accidents and traffic volume monitor coverage in Nevada.}
    \label{fig_traffic_volume_tradeoff}
\end{figure}

We present a distribution map of accident counts and AADT records in Figure~\ref{fig_traffic_volume_tradeoff}. It indicates that the traffic volume monitor has covered nearly all potential road segments that have had accidents.

\subsection{Weather Statistics}

The weather data is extracted using the Meteostat API, focusing on monthly weather statistics. For each node in the road network, the nearest weather station is identified based on geographic coordinates (latitude and longitude). We then retrieve monthly weather data for each location over a specified time range. The time range aligns with that of the traffic accident record. After fetching the data, it merges it with the corresponding node IDs and coordinates and stores the results in CSV files. Finally, all files are concatenated to create a comprehensive dataset matching historical weather conditions to each road network node.

We collect six types of features: the average daily air temperature in \SI{}{\degreeCelsius}, the average daily minimum air temperature in \SI{}{\degreeCelsius}, the average daily maximum air temperature in \SI{}{\degreeCelsius}, the monthly precipitation total in mm, the average wind speed in km/h, and the average sea-level air pressure in hPa.

In summary, we report all features collected from our road networks in this section.
For node features, we collect $9$ types: static features (latitude, longitude, and satellite imagery) and dynamic features (average surface temperature, maximum surface temperature, minimum surface temperature, total precipitation, average wind speed, and sea-level air pressure).
For edge features, we collect $3$ types of static features (location, length, and road type) and $1$ dynamic feature (annual average daily traffic).

\section{Experimental Details}\label{app_experiment_details}

\subsection{Implementation Details}\label{app_implementation_details}

We report the hyperparameters used in our experiments and provide the formal definitions of the fusion methods and evaluation metrics in the main result.

\paragraph{Basic fusion method.} The basic fusion method uses multilayer perceptrons to combine different features. Specifically, we first obtain the structure-aware representation $x_i^{(\text{GNN})}$ for each node $i$ by passing its attributes and graph connectivity through a multi-layer GNN. Independently, we extract a visual embedding $z_i = f_{\text{vision}}(I_i)$ for the image $I_i$ associated with node $i$ using a pretrained vision backbone such as CLIP or Vision Transformer (ViT), where $f_{\text{vision}}$ denotes the vision encoder.

To produce the final representation for downstream prediction, we concatenate the GNN embedding and the visual embedding:
\begin{equation}
\tilde{x}_i = [x_i^{(\text{GNN})} \parallel z_i],
\end{equation}
and feed the result into a multilayer perceptron (MLP) $f_{\text{mlp}}$, trained jointly with the GNN:
\begin{equation}
\hat{y}_i = f_{\text{mlp}}(\tilde{x}_i).
\end{equation}

The fusion MLP consists of four fully connected layers with non-linear activation functions (e.g., ReLU) and is optimized end-to-end alongside the GNN during training. This multimodal setup allows the model to leverage both spatial-structural and high-level visual cues, potentially improving predictive performance on downstream tasks such as accident risk prediction or traffic pattern classification.

\paragraph{Gated fusion network.} While MLP-based concatenation provides a straightforward way to combine graph and visual features, it treats both modalities equally for all nodes. To enable more flexible and context-aware fusion, we adopt a gated mechanism that adaptively weighs the contribution of each modality.

Since the vision and GNN embeddings may differ in dimensionality, the visual feature $z$ is first projected to match the GNN embedding space via a learnable linear transformation: $z' = W_{\text{proj}} z$, where $W_{\text{proj}}$ is a trainable matrix.

A scalar gate value $\lambda \in [0, 1]$ is then computed to determine the relative importance of the two modalities. The gate is derived from their concatenation:
\[
\lambda = \sigma \left( f_{\text{gate}} \left( [x^{(\text{GNN})} \parallel z] \right) \right),
\]
where $f_{\text{gate}}$ is a small MLP and $\sigma$ denotes the sigmoid function.

The final fused representation is a convex combination of the two:
\[
\tilde{x} = \lambda \cdot x^{(\text{GNN})} + (1 - \lambda) \cdot z'.
\]

This fused embedding $\tilde{x}$ is then fed into an MLP for prediction.

\paragraph{Mixture of Experts.} Finally, we implement the mixture-of-experts (MoE) architecture, which leverages multiple specialized networks, referred to as experts, that each learn to combine features from different perspectives. Given the visual feature $z$ and the graph embedding $x^{(\text{GNN})}$, we first construct a shared representation by concatenating the two modalities: $\tilde{x} = [x^{(\text{GNN})} \parallel z']$. This shared feature vector is then fed into a set of $K$ expert networks, each parameterized by its own learnable weights. Each expert $f_k$ produces an output representation:
\[ e^{(k)} = f_k(\tilde{x}), \quad k = 1, \ldots, K. \]

To dynamically select and combine the outputs of these experts, we use a gated network that computes a probability distribution over the experts for each node. This gating mechanism is realized as a small multilayer perceptron followed by a softmax activation:
\[ \lambda = \mathrm{softmax}(f_{\text{gate}}(\tilde{x})), \]
where $\lambda \in \mathbb{R}^K$ and $\sum_{k=1}^{K} \lambda^{(k)} = 1$. The final fused representation is then computed as a weighted sum of the expert outputs:
\[\tilde{e} = \sum_{k=1}^{K} \lambda^{(k)} \cdot e^{(k)}. \]

The fused expert output $\tilde{e}$ is subsequently passed through a prediction head along with the edge feature $x_{\mathrm{edge}}$ to generate the model’s output:
\[ \hat{y} = f_{\mathrm{pred}}(\tilde{e}, x_{\mathrm{edge}}). \]

This MoE architecture enables the model to dynamically select among multiple fusion pathways, capturing complex relationships between different modalities. With a learned gating mechanism, the MoE framework can adaptively select the most relevant experts for each node, yielding richer, more flexible multimodal representations.

\paragraph{Experiment hyperparameters.} For all baselines, we construct edge representations by concatenating the embeddings of the two connected nodes with the associated edge features. Node embeddings are fixed to a dimension of $128$. We use two-layer MLP and GNN architectures, both with a hidden size of $256$. The GNN, MLP, and fusion modules are fully trainable. All models are optimized with the Adam optimizer at a learning rate of $0.001$ and trained for $30$ epochs.

We split all the accident records into training, validation, and test sets for each state, based on different time periods. The summarization is shown in Table~\ref{tab_data_split}. Since our current analysis is at the monthly level, we note that our framework can also be used to conduct analysis at the yearly level.

\paragraph{Evaluation metrics.} Formally, for the regression task, let $y_i \in \mathbb{R}$ denote the ground-truth number of accidents for road segment $i$, and let $\hat{y}_i$ be the corresponding model prediction. The Mean Absolute Error (MAE) is computed as: $\mathcal{L}_{\text{MAE}} = \frac{1}{N} \sum_{i=1}^{N} |\hat{y}_i - y_i|$, where $N$ is the total number of road segments in the evaluation set.

For the classification task, let $y_i \in \{0,1\}$ be the binary indicator of whether an accident occurred on road segment $i$, and let $p_i \in [0,1]$ denote the predicted probability score. The AUROC measures the probability that a randomly chosen positive example is assigned a higher score than a randomly chosen negative one, and is formally defined as:
\[
\mathcal{L}_{\text{AUROC}} = \frac{1}{|S^+||S^-|} \sum_{i \in S^+} \sum_{j \in S^-} f(p_i > p_j) ,
\]
where $S^+$ and $S^-$ denote the sets of indices corresponding to positive and negative samples, respectively, and $f(\cdot)$ is the indicator function.

\paragraph{Omitted causal analysis}

Beyond direct matching, we also estimate causal effects using propensity score matching (PSM). We model the treatment assignment probability with
\[
e(x_i) = \Pr(t_i = 1 \mid x_i),
\]
using logistic regression on the multimodal embeddings. Matching in this score space aligns treated and control samples with similar treatment likelihoods and yields the PSM estimate
\[
\hat{\tau}_{\text{PSM}} = \frac{1}{|T|} \sum_{i \in T} (y_i - y_{j(i)}),
\]
where $j(i)$ is the matched control with the closest propensity score.

To increase robustness, we further apply a doubly robust estimator (DR) that combines outcome regression and propensity weighting. Let $\hat{m}_1(x_i)$ and $\hat{m}_0(x_i)$ be predicted outcomes under treatment and control. The DR estimator is written in a compact form as
\[
\hat{\tau}_{\text{DR}} = \frac{1}{|T|} \sum_{i \in T} \Big( \hat{m}_1(x_i) - \hat{m}_0(x_i) \Big) + \frac{1}{|T|} \sum_{i \in T} \omega_i,
\]
where the correction term is
\[
\omega_i = \frac{t_i\, (y_i - \hat{m}_1(x_i))}{\hat{e}(x_i)} - \frac{(1 - t_i)\, (y_i - \hat{m}_0(x_i))}{1 - \hat{e}(x_i)}.
\]

This embedding-based approach allows the causal analysis to control for confounding factors encoded in the multimodal representations. The PSM and DR adjustments provide more stable estimates when treatment imbalance exists, and support comparisons of treatment effects under similar structural, visual, and weather conditions.

\begin{table}[t]
\centering
\caption{The train, validation, and test data splitting of Delaware, Massachusetts, Maryland, Nevada, Montana, and Iowa in our framework. We report the period and associated accident records.}\label{tab_data_split}
\resizebox{1.00\columnwidth}{!}
{
\begin{tabular}{@{}lcccccc@{}}
\toprule
& \multicolumn{2}{c}{Train} & \multicolumn{2}{c}{Valid} & \multicolumn{2}{c}{Test} \\
 & period & records & period & records & period & records\\
\midrule
DE & 2009 -- 2013 & 145127 & 2014 -- 2018  & 179335 & 2019 -- 2024  & 208650 \\
MA & 2002 -- 2014 & 2887528 & 2015 -- 2020 & 1348597 & 2021 -- 2025 & 970504 \\
MD & 2015 -- 2017 & 341902 & 2018 -- 2019  & 232002 & 2020 -- 2024  & 423628 \\
NV & 2016 -- 2018 & 154834 & 2019 -- 2020  & 92359 & 2021 -- 2025   & 129059 \\
MT & 2016 -- 2018 & 60717 & 2019 -- 2020   & 15400 & 2021 -- 2023   & 63894 \\
IA & 2014 -- 2017 & 219013 & 2018 -- 2020  & 161186 & 2021 -- 2023  & 161186 \\
\bottomrule
\end{tabular}}
\end{table}

\subsection{Omitted Experiment Results}\label{app_omi_experiment_results}

In addition to AUROC, we evaluate the performance of different baselines using precision and recall as additional metrics.

\paragraph{Results using more metrics.} Table~\ref{tab_app_main_result} reports the test precision and recall scores across six U.S. states using a range of baselines, including node embedding methods, vision-based models, graph neural networks, and their multimodal extensions via fusion techniques. Across all models, GIN + MoE in Delaware (67.04\%), while the highest recall is observed in the same model for Montana (99.76\%), suggesting that mixture-of-experts can significantly boost detection capability under certain regional conditions.

In general, GIN-based fusion models show clear improvements over their unimodal counterparts. On average across six states, GIN + MoE improves precision by $2.41\%$  and recall by $0.03\%$  compared to the basic GIN model. These gains highlight the effectiveness of mixture-of-experts multimodal fusion in enhancing detection precision while maintaining strong recall performance across diverse traffic environments.

Nonetheless, we observe substantial variance in precision across regions, with some models showing high recall but relatively low precision, especially in Montana and Iowa. This indicates that while these models successfully capture most positive cases, they also generate a significant number of false positives. These findings underscore the effectiveness of multimodal fusion strategies while also highlighting challenges in balancing precision and recall, especially in regions with sparse accident data or heterogeneous road conditions.

\paragraph{Causal analysis results of PSM and DR} To enhance the causal analysis, we add Propensity Score Matching (PSM) and Doubly Robust (DR) estimators for more reliable treatment effect estimation, using one nearest neighbor and evaluating effects by road type. The results show a clear increase in accident risk for motorway segments, with effects of around $20\%$. To demonstrate consistent treatment effects rather than sensitivity to a specific matching setup, we also vary the number of nearest neighbors in the matching step. The estimates stay consistent across settings, showing that the DR estimator is stable.

\begin{table}[t]
\centering
\caption{Average treatment effect on the treated (ATT) across six U.S. states. 
The top block reports results from PSM and DR under one-nearest-neighbor matching ($k=1$). 
The bottom block varies the number of neighbors for DR to assess robustness.
We evaluate treatment effects related to road type. We calculate the mean and standard deviation across multiple years.
}\label{tab_psm_dr}
\resizebox{1.00\columnwidth}{!}
{
\begin{tabular}{@{}lcccccc@{}}
\toprule
      & DE & MA & MD & NV & MT & IA \\ 
\midrule
 PSM & $18.1\pm0.2$ & $19.0\pm0.4$ & $16.5\pm0.1$ & $17.2\pm0.2$ & $14.8\pm0.2$ & $12.2\pm0.1$ \\
 DR ($k=1$) & $37.7\pm4.2$ & $21.0\pm3.1$ & $21.7\pm3.8$ & $21.8\pm3.4$ & $19.7\pm3.1$ & $16.3\pm2.5$ \\
\midrule
 DR ($k=3$) & $37.7\pm3.4$ & $20.9\pm3.2$ & $21.8\pm3.8$ & $21.8\pm3.4$ & $19.7\pm3.2$ & $16.2\pm2.4$ \\
 DR ($k=5$) & $37.7\pm3.4$ & $20.9\pm3.2$ & $21.8\pm3.8$ & $21.8\pm3.4$ & $19.7\pm3.2$ & $16.2\pm2.4$ \\
\bottomrule
\end{tabular}}
\end{table}

\subsection{Examples of Using \acronym{} Package}\label{app_code_example}

Here we report some examples of how to implement \acronym{} package, including data loader and trainer.

\begin{lstlisting}[language=Python, caption=MMTraCE data loader and trainer, label=code_trainer]
>>> from mmtrace import Trainer, Evaluator, MMDataset
# Create the dataset
>>> dataset = MMDataset(state_name = "MA")
>>> data = dataset.load_monthly_data(year = 2024, month = 12)
>>> acci, acci_cnt = data["accidents"], data["accident_cnt"]
# Load different types of features
>>> node_attr, edge_attr = data["x"], data["edge_attr"]
>>> img_attr = data["img_embeddings"]
# Get an evaluator for accident prediction
>>> evaluator = Evaluator(type = "classification")
# Initialize a trainer with the model, dataset, and evaluator
>>> trainer = Trainer(model, dataset, evaluator, ...)
# Conduct training and evaluation inside the trainer
>>> log = trainer.train()
\end{lstlisting}

From the code in Listing~\ref{code_trainer}, we can access the data from a specific state, year, and month. We can also fetch the static and dynamic features. Node features, edge features, and edge embeddings are all available. Then, we can initialize a trainer for implementation. The model can be applied to any baseline model.

\section{Privacy Details and Leakage Audit}

All data sources used in this study are publicly available and released under open data licenses by the corresponding agencies. State Departments of Transportation provide traffic accident and traffic volume records through official open data portals that support academic and non-commercial use. The OSMnx Street Network Dataverse is licensed under the Open Database License (ODbL), which allows reuse under standard attribution and share-alike terms. Mapbox imagery is sourced through the official Static Tiles API under the standard developer terms of service. Weather data is collected from the official Meteostat API, which offers open access to historical and near-real-time meteorological observations for research. Table~\ref{tab_datasource_link} lists all sources, license statements, and links.

All datasets used in this work contain only anonymized records released by public agencies. No record includes personal identifiers, and no field can be linked back to individuals. The spatial resolution of the satellite and grid-level imagery does not allow identification of people, vehicles, or private property details. Our processing pipeline uses only node- and segment-level information derived from public road networks, ensuring that the final dataset remains fully anonymous and consistent with public data release standards.

In addition to these safeguards, we conduct a full leakage audit to verify that no hidden pathways for re-identification remain. We check that accident counts are aggregated at the segment or monthly level, that imagery does not contain fine-grained visual cues tied to individuals, and that no timestamps, device traces, or location histories are included in the dataset during pre-processing. We also confirm that structural information from OSMnx cannot be cross-matched with any external source to recover personal movement patterns. These checks show that the final dataset cannot be used to infer identities or sensitive attributes. In our experiments, we divide the entire time series into three non-overlapping segments for training, validation, and testing, as shown in Table~\ref{tab_data_split}. The splits are strictly chronological, ensuring that no future information leaks into the training or validation sets.

\clearpage
\begin{table*}[!ht]
\centering
\caption{Summary of the data sources used in our dataset construction, including links to official repositories for traffic accident records, traffic volume statistics, and APIs to road network data, weather observations, and satellite imagery. They are official records and form the foundation of our multimodal traffic safety dataset.}\label{tab_datasource_link}
% \resizebox{\textwidth}{!}
{
\begin{tabular}{@{}ll@{}}
\toprule
\multicolumn{2}{c}{\textbf{Traffic Accident Records}} \\
\midrule
Delaware DOT & \href{https://data.delaware.gov/Transportation/Public-Crash-Data/827n-m6xc/about_data}{https://data.delaware.gov/Transportation/Public-Crash-Data/827n-m6xc/about\_data}\\
Massachusetts DOT & \href{https://massdot-impact-crashes-vhb.opendata.arcgis.com/search}{https://massdot-impact-crashes-vhb.opendata.arcgis.com/search}\\
Maryland DOT & \href{https://mdsp.maryland.gov/Pages/Dashboards/CrashDataDownload.aspx}{https://mdsp.maryland.gov/Pages/Dashboards/CrashDataDownload.aspx}\\
Nevada DOT & \href{https://geohub-ndot.hub.arcgis.com/datasets/NDOT::crashdata-opendata/explore?location=38.464396%2C-116.977900%2C7.12&showTable=true}{https://geohub-ndot.hub.arcgis.com/datasets/NDOT::crashdata-opendata/}\\
Montana DOT & \href{https://www.mdt.mt.gov/publications/datastats/crashdata.aspx}{https://www.mdt.mt.gov/publications/datastats/crashdata.aspx}  \\ & \href{https://www.mdt.mt.gov/publications/docs/datastats/crashdata/PublicCrashData2019-2023.xlsx}{\parbox[t]{0.60\textwidth}{https://www.mdt.mt.gov/publications/docs/datastats/crashdata/PublicCrashData2019-2023.xlsx}}\\
Iowa DOT & \href{https://icat.iowadot.gov/}{https://icat.iowadot.gov/}\\
OSMnx Street Network Dataverse & \href{https://dataverse.harvard.edu/dataset.xhtml?persistentId=doi:10.7910/DVN/CUWWYJ}{\parbox[t]{0.60\textwidth}{https://dataverse.harvard.edu/dataset.xhtml?persistentId=doi:10.7910/DVN/CUWWYJ}}\\

\midrule
\multicolumn{2}{c}{\textbf{Traffic Volumes}} \\
\midrule
Delaware & \href{https://de-firstmap-delaware.hub.arcgis.com/datasets/delaware::delaware-traffic-counts-2-0/explore?location=-0.000000%2C0.000000%2C1.79&showTable=true}{\parbox[t]{0.60\textwidth}{https://de-firstmap-delaware.hub.arcgis.com/datasets/delaware::delaware-traffic-counts-2-0/}}\\
Massachusetts & \href{https://mhd.public.ms2soft.com/tcds/tsearch.asp?loc=Mhd&mod=}{https://mhd.public.ms2soft.com/tcds/tsearch.asp?loc=Mhd\&mod=}\\
Maryland & \href{https://data-maryland.opendata.arcgis.com/datasets/maryland::mdot-sha-annual-average-daily-traffic-aadt-locations/explore?location=38.824341%2C-77.269759%2C8.44&showTable=true}{\parbox[t]{0.60\textwidth}{https://data-maryland.opendata.arcgis.com/datasets/maryland::mdot-sha-annual-average-daily-traffic-aadt-locations/}}\\
Nevada & \href{https://geohub-ndot.hub.arcgis.com/datasets/015483ecc6ca422fb62136acc502c793_0/explore?location=37.145264%2C-114.025413%2C5.21&showTable=true}{\parbox[t]{0.60\textwidth}{https://geohub-ndot.hub.arcgis.com/datasets}}\\
Montana & \href{https://mdt.public.ms2soft.com/tcds/tsearch.asp?loc=Mdt&mod=}{https://mdt.public.ms2soft.com/tcds/tsearch.asp?loc=Mdt\&mod=}\\
Iowa & \href{https://experience.arcgis.com/experience/291d4e2c64cf490a95a6660b1349a088/page/Historical-AADT}{https://experience.arcgis.com/experience}\\
\midrule
\multicolumn{2}{c}{\textbf{Weather Observations}} \\
\midrule
Meteostat API & \href{https://meteostat.net/en/}{https://meteostat.net/en/}\\
\midrule
\multicolumn{2}{c}{\textbf{Supplementary Websites}} \\
\midrule
Google Map API & \href{https://maps.google.com/}{https://maps.google.com/} \\
\bottomrule
\end{tabular}}
\end{table*}

\clearpage
\renewcommand{\arraystretch}{0.90}
\begin{table*}[ht]
\centering
\caption{We report the mean absolute error (MAE), precision, and recall score on the test split using node embedding, graph neural network embeddings, supervised contrastive learning, and feature fusion methods. To account for variability, each experiment is repeated with three different random seeds, and we report the averaged results along with standard deviations.}\label{tab_app_main_result}
{\small
\begin{tabular}{@{}l|cccccc@{}}
\toprule
        & Delaware & Massachusetts & Maryland & Nevada & Montana & Iowa \\ 
MAE / Average count & $4.59$ & $7.25$ & $1.72$ & $1.29$ & $0.40$ & $0.84$ \\
\midrule
GCN + Basic Fusion & $0.1\pm0.00$ & $0.4\pm0.02$ & $0.2\pm0.01$ & $0.1\pm0.01$ & $0.1\pm0.02$ & $0.3\pm0.14$ \\
GCN + Gated Fusion & $0.1\pm0.01$ & $0.5\pm0.02$ & $0.2\pm0.00$ & $0.1\pm0.01$ & $0.1\pm0.01$ & $0.1\pm0.03$ \\
GCN + MoE & $0.1\pm0.01$ & $0.3\pm0.01$ & $0.1\pm0.02$ & $0.1\pm0.01$ & $0.1\pm0.01$ & $0.1\pm0.02$ \\
\midrule
AUROC / Positive rate & $0.32$ & $0.28$ & $0.24$ & $0.14$ & $0.10$ & $0.18$ \\ 
\midrule
GCN + Basic Fusion & $88.5\pm0.3$ & $87.0\pm0.1$ & $89.2\pm0.1$ & $92.9\pm0.1$ & $80.0\pm0.9$ & $82.3\pm0.6$ \\
GCN + Gated Fusion & $88.5\pm0.2$ & $87.7\pm0.1$ & $89.3\pm0.0$ & $93.5\pm0.1$ & $83.7\pm0.7$ & $84.4\pm0.5$ \\
GCN + MoE & $88.7\pm0.7$ & $87.9\pm0.3$ & $88.3\pm0.0$ & $93.7\pm0.1$ & $83.0\pm0.8$ & $84.9\pm0.4$ \\
\midrule
Precision \\ 
\midrule
MLP         & $6.33\pm1.1$ & $1.99\pm0.8$ & $2.49\pm0.5$ & $5.33\pm0.6$ & $0.75\pm0.1$ & $1.19\pm0.1$ \\
DeepWalk    & $16.07\pm3.9$ & $4.19\pm0.4$ & $5.69\pm0.8$ & $6.17\pm0.5$ & $1.73\pm0.1$ & $3.04\pm0.3$\\
GraphSAGE   & $24.11\pm3.4$ & $10.06\pm2.5$ & $11.44\pm0.2$ & $5.24\pm0.6$ & $3.23\pm0.7$ & $5.94\pm1.5$ \\
SupConGCN      & $17.21\pm3.5$ & $5.23\pm0.3$ & $4.84\pm0.8$ & $4.89\pm0.7$ & $2.28\pm0.6$ & $2.38\pm0.4$ \\
Graph Transformer & $26.13\pm5.0$ & $8.96\pm0.9$ & $11.25\pm0.8$ & $7.18\pm1.1$ & $0.64\pm0.2$ & $6.21\pm1.1$ \\
DCRNN       & $38.08\pm4.2$ & $4.06\pm0.1$ & $5.08\pm0.2$ & $2.11\pm0.5$ & $2.63\pm0.4$ & $6.34\pm1.2$ \\
ViT         & $16.84\pm2.9$ & $4.24\pm0.8$ & $5.21\pm0.2$ & $6.00\pm0.5$ & $1.71\pm0.0$ & $2.69\pm0.3$ \\
CLIP        & $11.49\pm2.5$ & $3.56\pm0.2$ & $4.37\pm0.1$ & $5.02\pm0.3$ & $1.27\pm0.1$ & $2.55\pm0.5$ \\
\midrule
GCN         & $11.53\pm0.6$ & $6.27\pm0.3$ & $5.62\pm0.2$ & $6.09\pm0.4$ & $1.90\pm0.3$ & $8.66\pm2.3$ \\
GCN + Basic Fusion & $36.69\pm4.5$ & $4.25\pm0.1$ & $4.67\pm0.3$ & $5.67\pm0.1$ & $3.64\pm0.3$ & $2.32\pm0.3$ \\
GCN + Gated Fusion & $46.85\pm4.3$ & $4.17\pm0.0$ & $4.50\pm0.4$ & $5.41\pm0.1$ & $2.95\pm0.3$ & $3.83\pm2.1$ \\
GCN + MoE & $52.43\pm3.9$ & $5.96\pm0.1$ & $4.82\pm0.5$ & $6.71\pm0.8$ & $0.61\pm0.2$ & $3.07\pm0.4$ \\
\midrule
GIN         & $45.94\pm5.2$ & $5.05\pm0.9$ & $5.13\pm0.2$ & $8.19\pm0.9$ & $1.39\pm0.0$ & $2.62\pm0.3$ \\
GIN + Basic Fusion & $49.94\pm3.3$ & $6.97\pm2.1$ & $6.18\pm0.1$ & $5.98\pm0.5$ & $2.32\pm0.4$ & $2.64\pm0.2$ \\
GIN + Gated Fusion & $55.38\pm3.5$ & $5.68\pm0.3$ & $5.88\pm0.2$ & $5.32\pm0.1$ & $1.97\pm0.5$ & $3.15\pm0.3$ \\
GIN + MoE & $67.04\pm4.1$ & $5.16\pm0.2$ & $6.39\pm1.0$ & $6.41\pm0.9$ & $1.41\pm0.2$ & $2.67\pm0.2$ \\
\midrule
Recall      \\
\midrule
MLP         & $70.35\pm6.0$ & $53.46\pm7.7$ & $87.66\pm9.7$ & $90.76\pm4.5$ & $79.59\pm4.3$ & $69.57\pm2.2$ \\
DeepWalk    & $88.76\pm1.9$ & $81.24\pm3.4$ & $87.55\pm5.4$ & $90.82\pm2.7$ & $67.36\pm2.4$ & $80.99\pm1.5$\\
GraphSAGE   & $44.27\pm6.1$ & $67.49\pm1.2$ & $69.89\pm7.5$ & $80.17\pm1.5$ & $79.47\pm4.2$ & $67.34\pm4.7$ \\
SupConGCN      & $76.23\pm7.0$ & $60.85\pm3.1$ & $80.68\pm3.5$ & $87.23\pm0.8$ & $77.98\pm1.3$ & $76.60\pm1.9$ \\
Graph Transformer  & $60.39\pm2.9$ & $63.08\pm3.8$ & $69.76\pm3.7$ & $77.35\pm5.0$ & $79.68\pm5.1$ & $77.75\pm6.5$ \\
DCRNN       & $79.80\pm4.3$ & $79.02\pm2.7$ & $81.26\pm0.9$ & $87.96\pm2.3$ & $78.34\pm3.2$ & $76.55\pm2.8$ \\
ViT         & $92.23\pm5.4$ & $78.99\pm1.1$ & $82.06\pm1.7$ & $90.94\pm3.1$ & $66.41\pm1.7$ & $80.28\pm2.5$ \\
CLIP        & $74.40\pm3.7$ & $75.84\pm3.3$ & $82.81\pm0.7$ & $88.62\pm0.9$ & $86.69\pm1.9$ & $76.56\pm1.5$ \\
\midrule
GCN         & $79.47\pm5.4$ & $75.41\pm1.5$ & $78.15\pm1.2$ & $74.80\pm2.0$ & $64.31\pm2.1$ & $58.55\pm0.5$ \\
GCN + Basic Fusion & $85.87\pm1.3$ & $85.24\pm0.8$ & $88.50\pm0.3$ & $87.38\pm1.0$ & $55.01\pm4.9$ & $93.15\pm4.0$ \\
GCN + Gated Fusion & $74.28\pm3.9$ & $84.79\pm1.4$ & $89.58\pm1.0$ & $90.04\pm0.8$ & $98.26\pm1.0$ & $73.12\pm3.6$ \\
GCN + MoE & $79.10\pm4.5$ & $66.88\pm4.8$ & $83.20\pm1.5$ & $87.73\pm1.7$ & $99.76\pm0.2$ & $91.85\pm1.7$ \\
\midrule
GIN         & $93.88\pm0.3$ & $76.69\pm2.5$ & $80.10\pm0.8$ & $84.85\pm2.5$ & $78.86\pm3.2$ & $77.01\pm0.9$ \\
GIN + Basic Fusion & $74.95\pm3.6$ & $68.03\pm2.5$ & $76.99\pm3.0$ & $89.79\pm0.2$ & $56.92\pm0.4$ & $82.50\pm1.6$ \\
GIN + Gated Fusion & $80.00\pm3.3$ & $75.59\pm4.8$ & $80.56\pm1.4$ & $89.75\pm0.1$ & $97.87\pm1.4$ & $81.01\pm3.3$ \\
GIN + MoE & $85.10\pm4.6$ & $72.08\pm3.3$ & $80.57\pm1.1$ & $90.56\pm0.5$ & $91.15\pm0.7$ & $83.55\pm1.9$ \\
\midrule
$F_1$-score \\
\midrule
MLP         & $10.86\pm2.2$ & $2.86\pm0.2$ & $3.94\pm0.6$ & $9.02\pm0.8$ & $1.50\pm0.1$ & $2.31\pm0.3$ \\
DeepWalk    & $25.86\pm5.2$ & $7.05\pm1.0$ & $10.46\pm1.3$ & $11.46\pm0.9$ & $3.35\pm0.1$ & $5.72\pm0.6$\\
ViT         & $30.93\pm2.5$ & $6.69\pm0.5$ & $9.67\pm0.3$ & $10.00\pm0.1$ & $3.32\pm0.1$ & $5.14\pm0.5$ \\
CLIP        & $19.89\pm2.3$ & $6.83\pm0.1$ & $8.26\pm0.4$ & $9.45\pm0.4$ & $2.49\pm0.1$ & $4.87\pm0.9$ \\
GraphSAGE   & $31.57\pm2.2$ & $8.17\pm0.7$ & $8.17\pm0.7$ & $9.74\pm1.0$ & $4.34\pm1.3$ & $5.89\pm0.6$ \\
SupCon      & $26.49\pm9.6$ & $8.58\pm0.2$ & $8.90\pm1.4$ & $9.13\pm1.2$ & $4.28\pm1.0$ & $4.56\pm0.8$ \\
\midrule
GCN         & $40.50\pm1.0$ & $7.43\pm0.5$ & $10.43\pm0.3$ & $11.04\pm0.6$ & $3.33\pm0.4$ & $5.21\pm0.5$ \\
GCN + Basic Fusion & $45.42\pm4.4$ & $9.58\pm0.2$ & $10.46\pm1.3$ & $11.41\pm0.8$ & $4.96\pm1.1$ & $5.33\pm0.5$ \\
GCN + Gated Fusion & $50.27\pm5.2$ & $9.92\pm0.0$ & $11.07\pm0.8$ & $11.28\pm0.1$ & $5.03\pm0.2$ & $6.12\pm0.4$ \\
GCN + MoE & $56.78\pm5.7$ & $10.51\pm0.1$ & $11.99\pm0.7$ & $12.38\pm1.4$ & $5.21\pm0.3$ & $6.68\pm0.6$ \\
\midrule
GIN         & $64.01\pm4.3$ & $9.77\pm2.1$ & $9.53\pm0.5$ & $14.78\pm1.4$ & $2.72\pm0.0$ & $5.04\pm0.5$ \\
GIN + Basic Fusion & $70.04\pm7.5$ & $11.11\pm1.8$ & $11.27\pm0.1$ & $15.15\pm0.8$ & $4.39\pm0.7$ & $5.17\pm0.3$ \\
GIN + Gated Fusion & $73.78\pm4.7$ & $10.23\pm0.3$ & $10.87\pm0.3$ & $15.96\pm0.4$ & $4.10\pm0.9$ & $6.00\pm0.6$ \\
GIN + MoE & $76.27\pm3.8$ & $10.73\pm0.1$ & $11.67\pm1.6$ & $15.91\pm1.7$ & $4.74\pm0.4$ & $6.14\pm0.3$ \\
\bottomrule
\end{tabular}}
\end{table*}

\renewcommand{\arraystretch}{1.00}

\begin{figure*}[h]
    \centering
    \subcaptionbox{Delaware}[1.0\textwidth]{
        \centering
        \begin{minipage}[b]{0.16\textwidth}
            \includegraphics[width=0.95\textwidth]{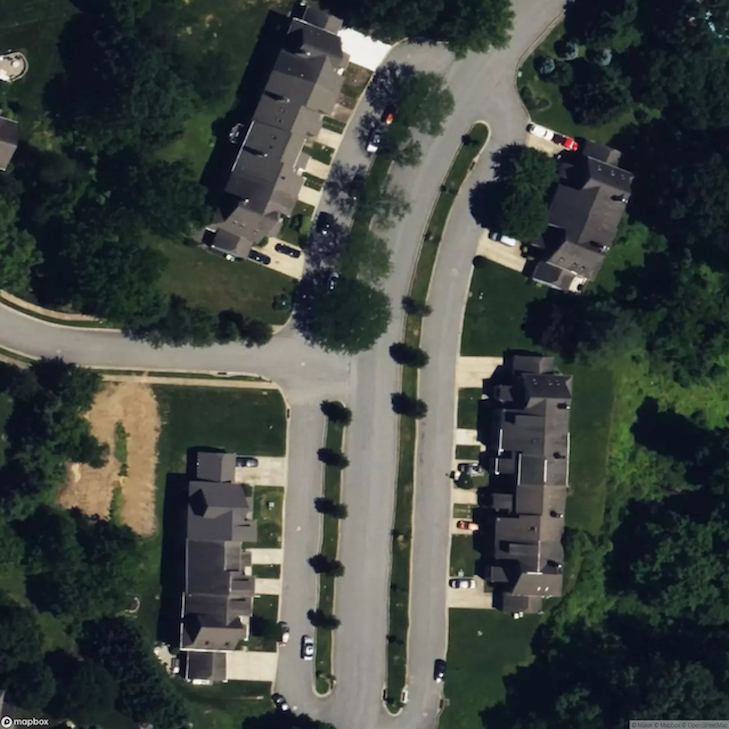}
        \end{minipage}
        \begin{minipage}[b]{0.16\textwidth}
            \includegraphics[width=0.95\textwidth]{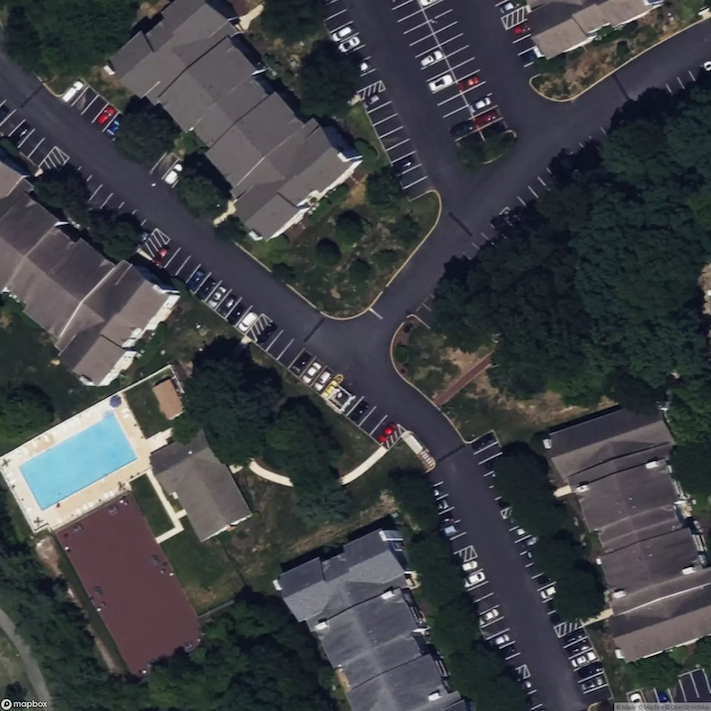}
        \end{minipage}
        \begin{minipage}[b]{0.16\textwidth}
            \includegraphics[width=0.95\textwidth]{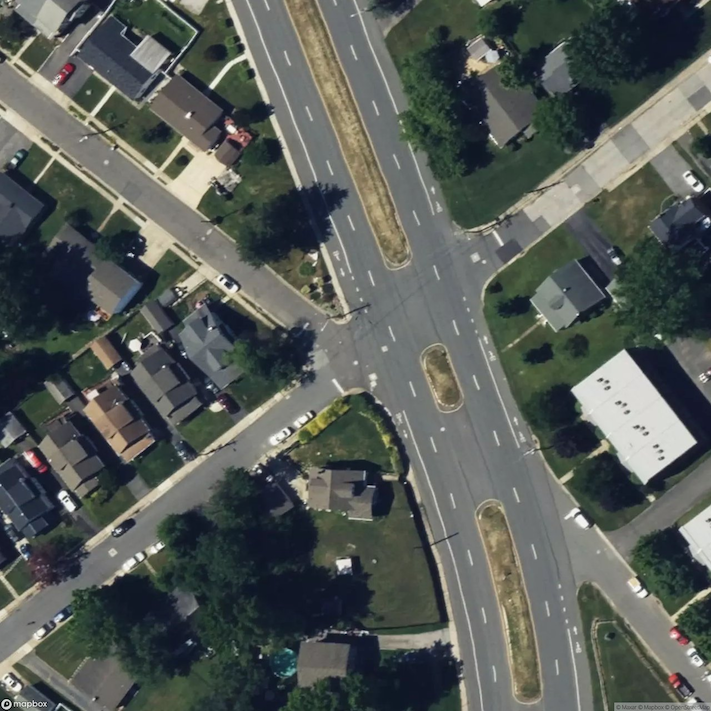}
        \end{minipage}
        \begin{minipage}[b]{0.16\textwidth}
            \includegraphics[width=0.95\textwidth]{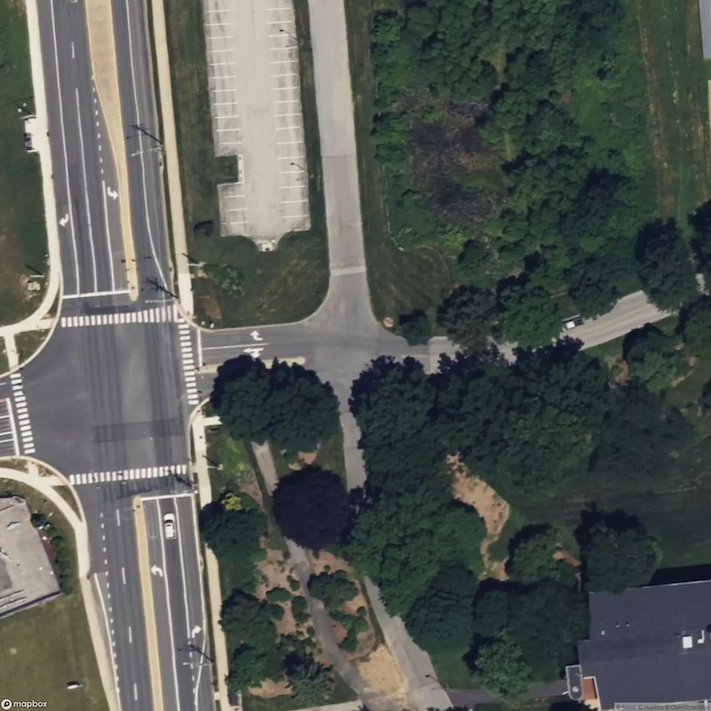}
        \end{minipage}
        \begin{minipage}[b]{0.16\textwidth}
            \includegraphics[width=0.95\textwidth]{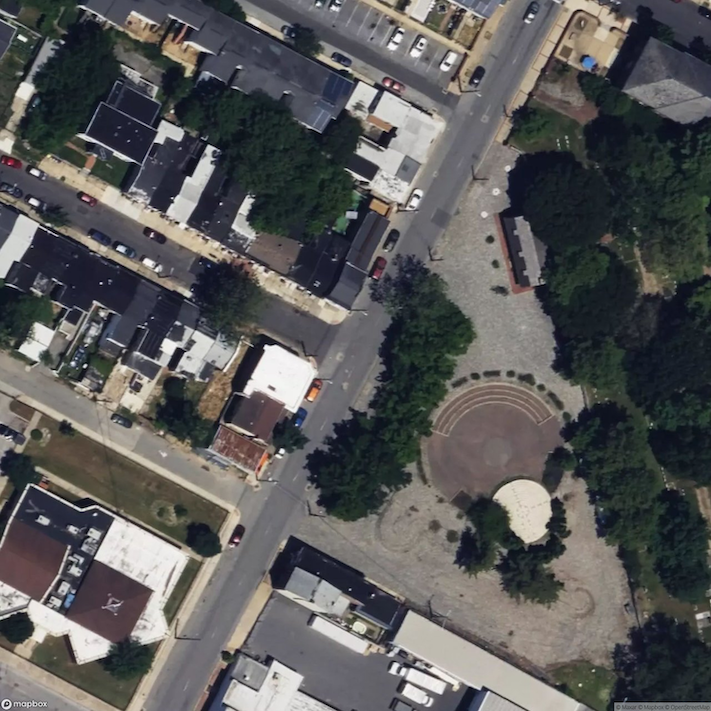}
        \end{minipage}
        \begin{minipage}[b]{0.16\textwidth}
            \includegraphics[width=0.95\textwidth]{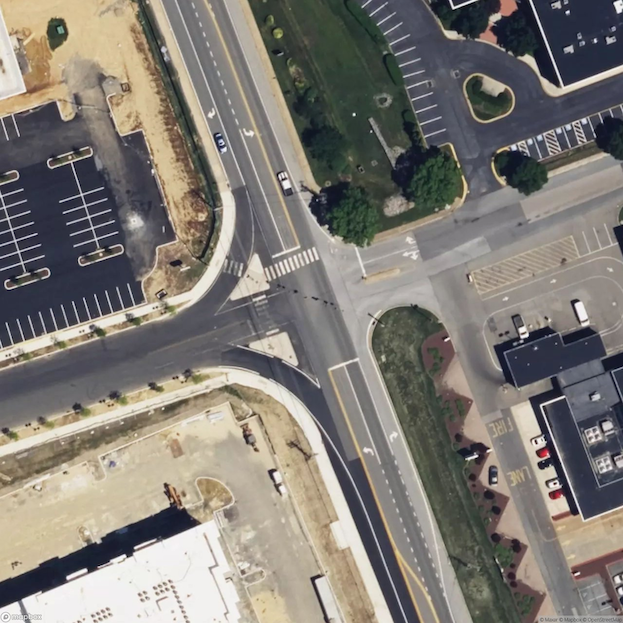}
        \end{minipage}
    }
    \subcaptionbox{Massachusetts}[1.0\textwidth]{
        \centering
        \begin{minipage}[b]{0.16\textwidth}
            \includegraphics[width=0.95\textwidth]{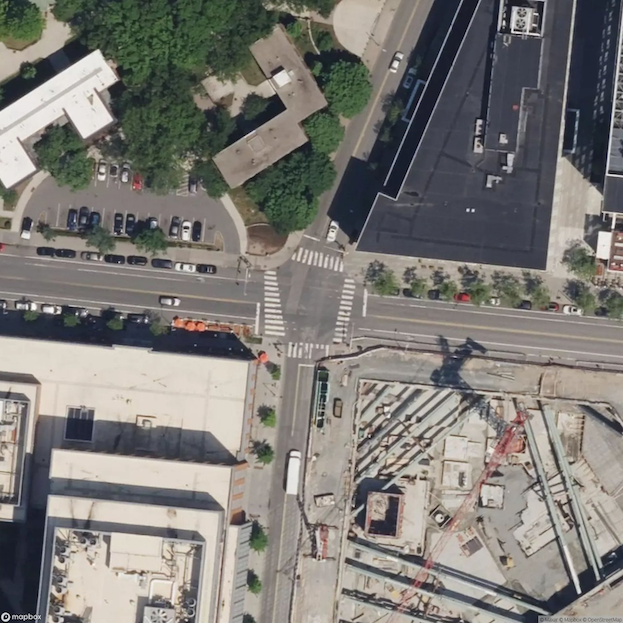}
        \end{minipage}
        \begin{minipage}[b]{0.16\textwidth}
            \includegraphics[width=0.95\textwidth]{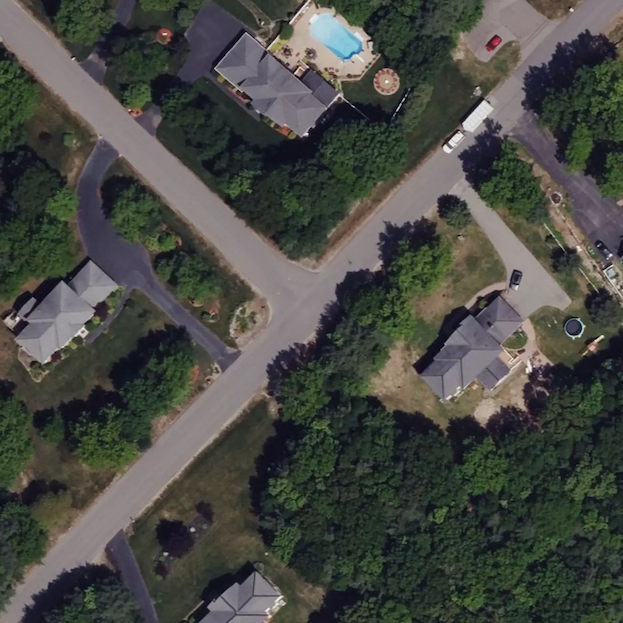}
        \end{minipage}
        \begin{minipage}[b]{0.16\textwidth}
            \includegraphics[width=0.95\textwidth]{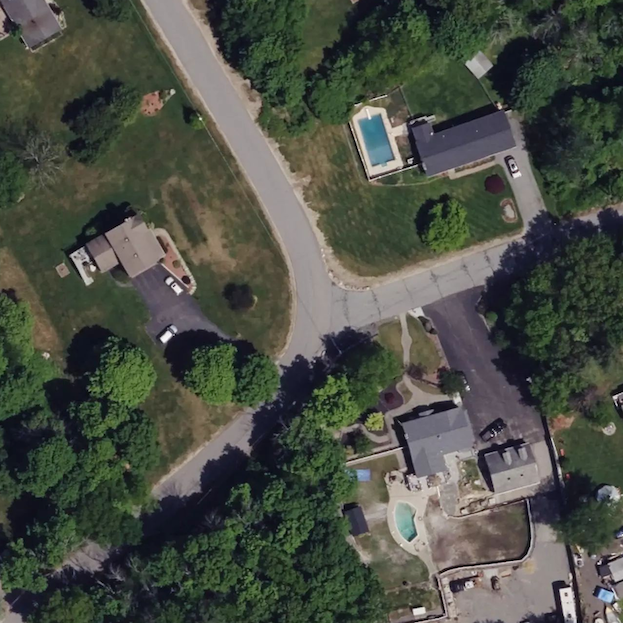}
        \end{minipage}
        \begin{minipage}[b]{0.16\textwidth}
            \includegraphics[width=0.95\textwidth]{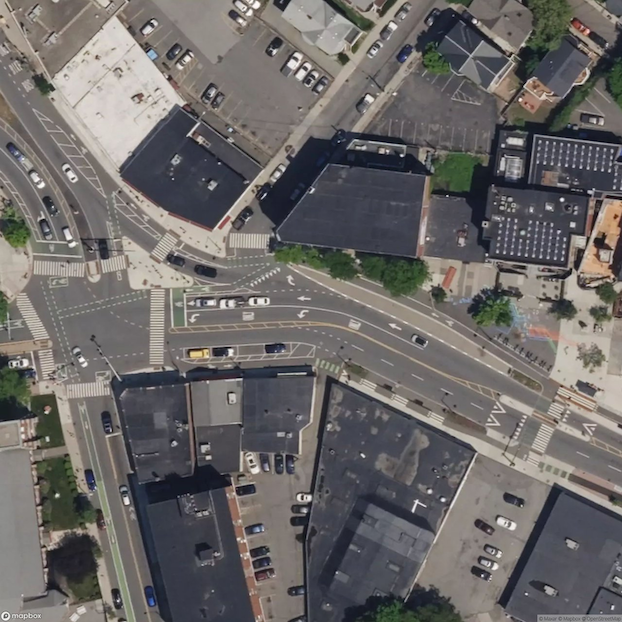}
        \end{minipage}
        \begin{minipage}[b]{0.16\textwidth}
            \includegraphics[width=0.95\textwidth]{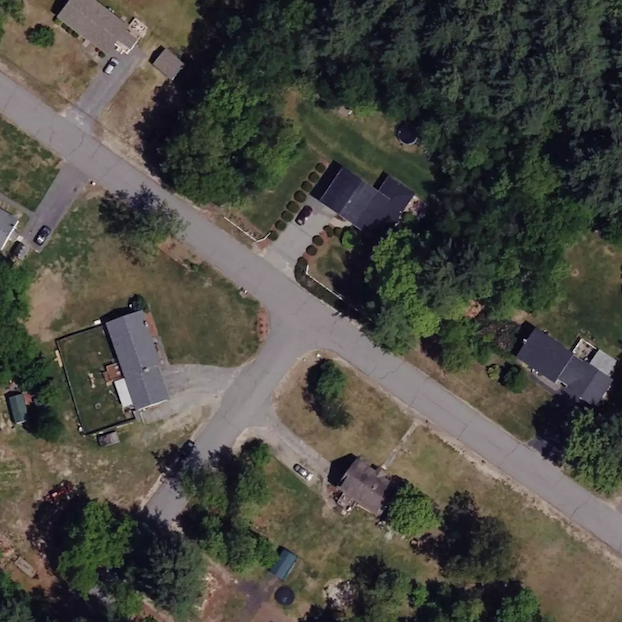}
        \end{minipage}
        \begin{minipage}[b]{0.16\textwidth}
            \includegraphics[width=0.95\textwidth]{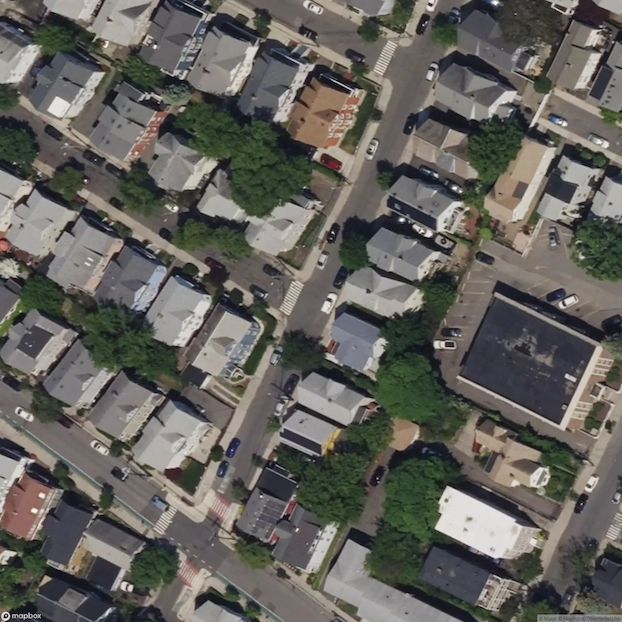}
        \end{minipage}
    }
    \subcaptionbox{Maryland}[1.0\textwidth]{
        \centering
        \begin{minipage}[b]{0.16\textwidth}
            \includegraphics[width=0.95\textwidth]{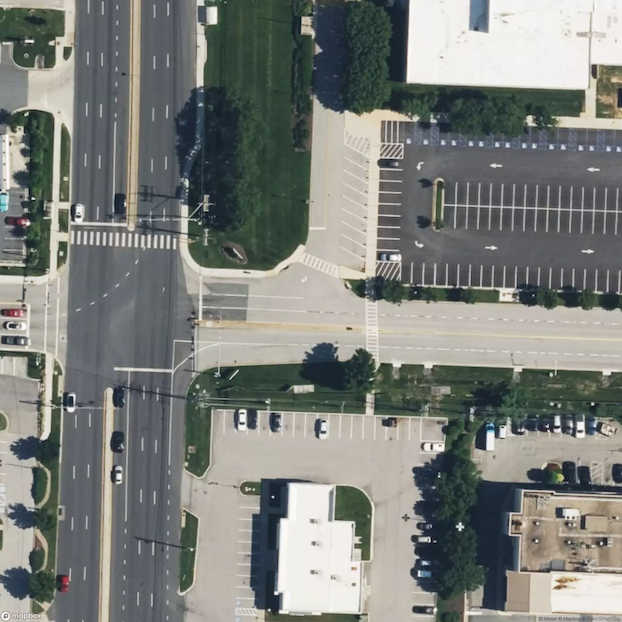}
        \end{minipage}
        \begin{minipage}[b]{0.16\textwidth}
            \includegraphics[width=0.95\textwidth]{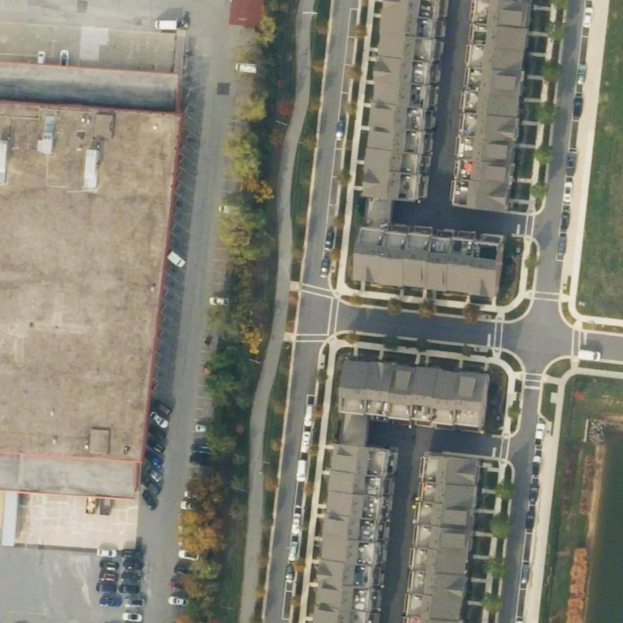}
        \end{minipage}
        \begin{minipage}[b]{0.16\textwidth}
            \includegraphics[width=0.95\textwidth]{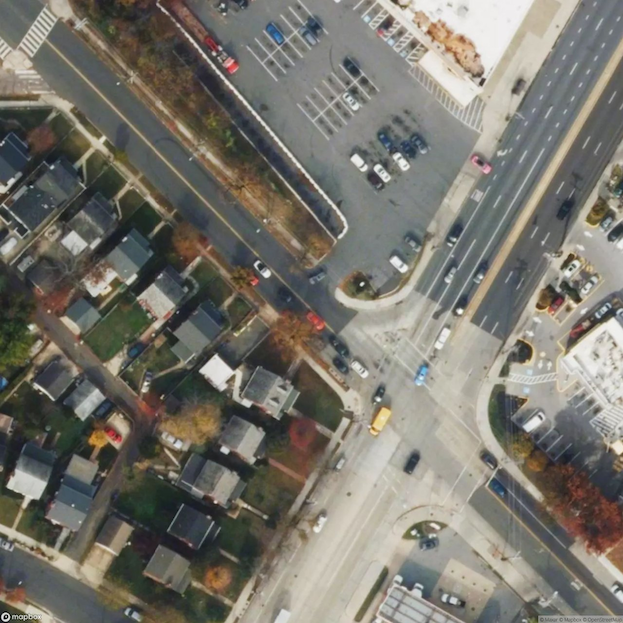}
        \end{minipage}
        \begin{minipage}[b]{0.16\textwidth}
            \includegraphics[width=0.95\textwidth]{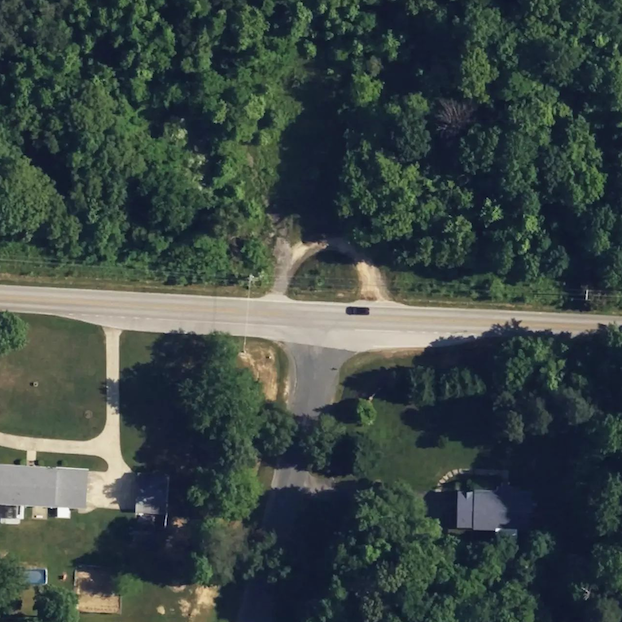}
        \end{minipage}
        \begin{minipage}[b]{0.16\textwidth}
            \includegraphics[width=0.95\textwidth]{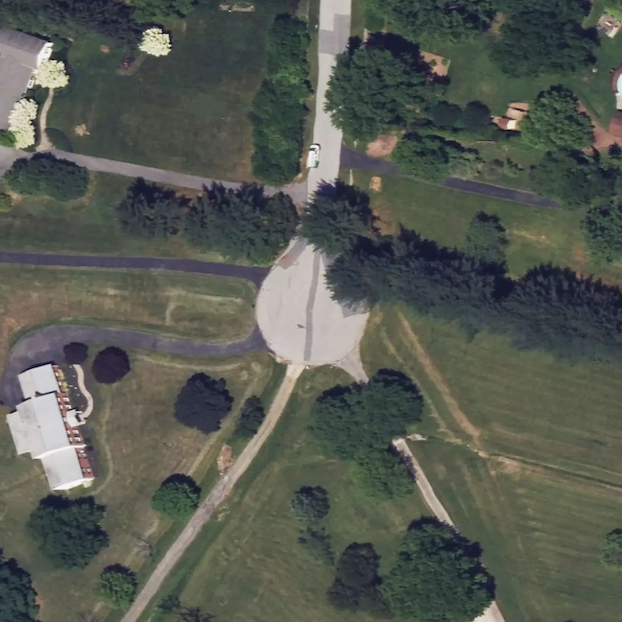}
        \end{minipage}
        \begin{minipage}[b]{0.16\textwidth}
            \includegraphics[width=0.95\textwidth]{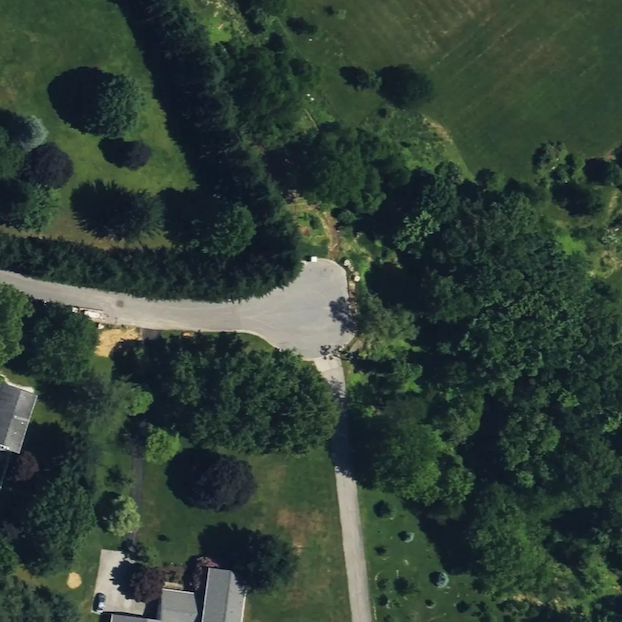}
        \end{minipage}
    }
    \subcaptionbox{Nevada}[1.0\textwidth]{
        \centering
        \begin{minipage}[b]{0.16\textwidth}
            \includegraphics[width=0.95\textwidth]{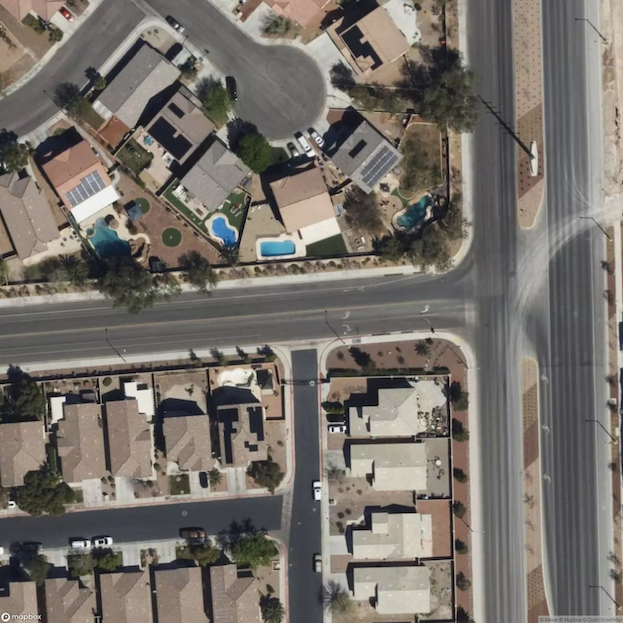}
        \end{minipage}
        \begin{minipage}[b]{0.16\textwidth}
            \includegraphics[width=0.95\textwidth]{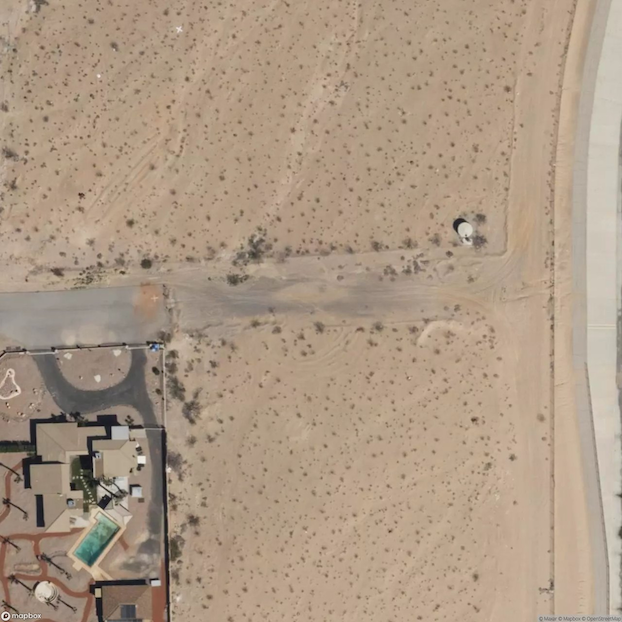}
        \end{minipage}
        \begin{minipage}[b]{0.16\textwidth}
            \includegraphics[width=0.95\textwidth]{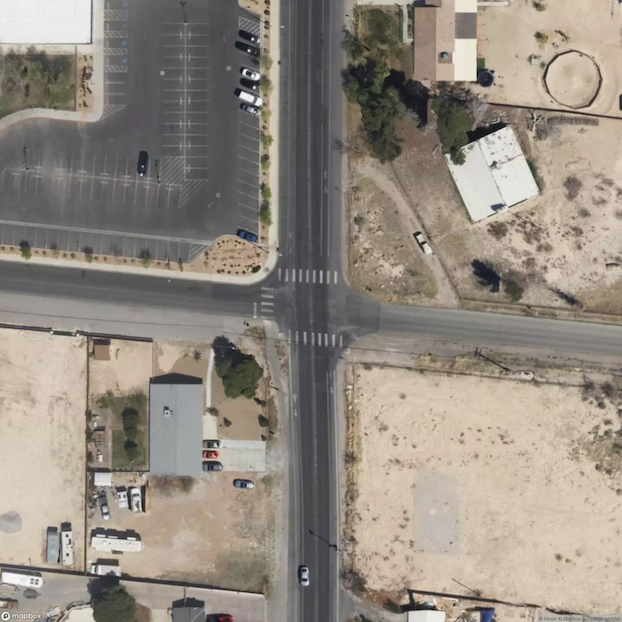}
        \end{minipage}
        \begin{minipage}[b]{0.16\textwidth}
            \includegraphics[width=0.95\textwidth]{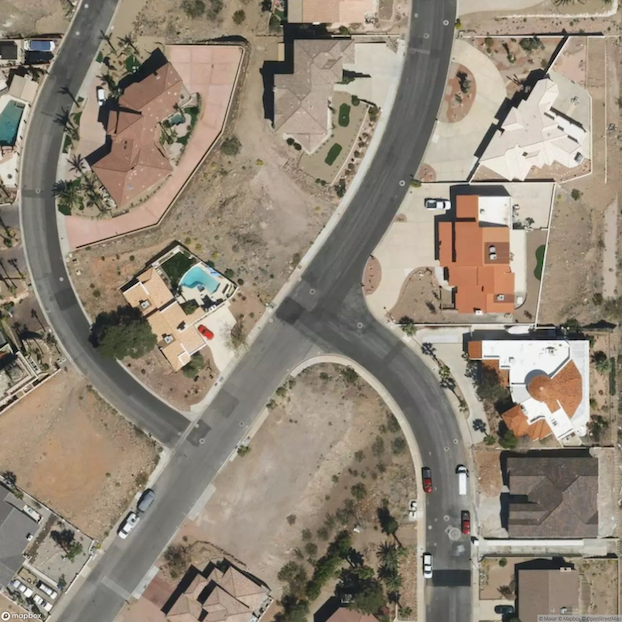}
        \end{minipage}
        \begin{minipage}[b]{0.16\textwidth}
            \includegraphics[width=0.95\textwidth]{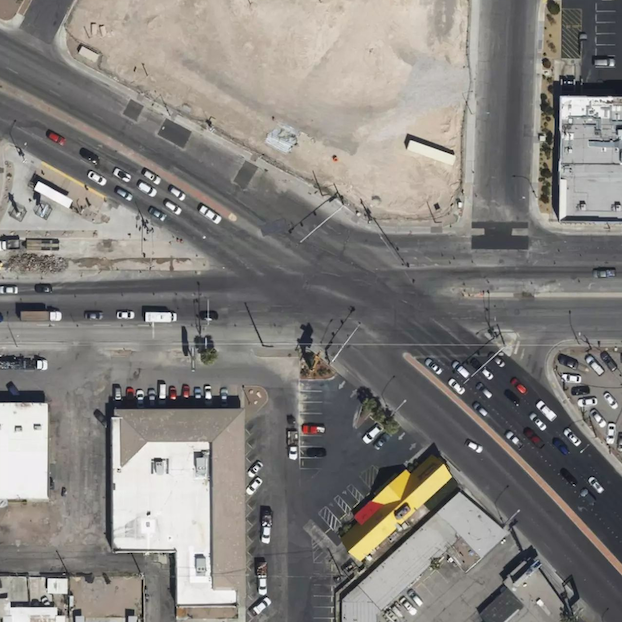}
        \end{minipage}
        \begin{minipage}[b]{0.16\textwidth}
            \includegraphics[width=0.95\textwidth]{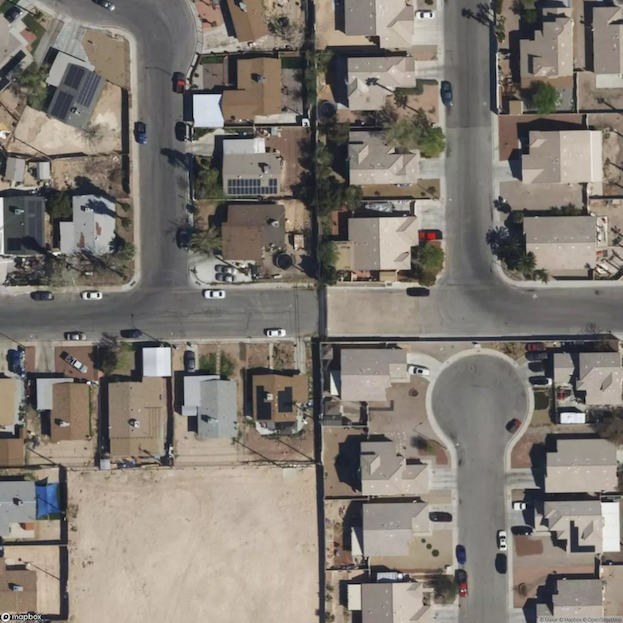}
        \end{minipage}
    }
    \subcaptionbox{Montana}[1.0\textwidth]{
        \centering
        \begin{minipage}[b]{0.16\textwidth}
            \includegraphics[width=0.95\textwidth]{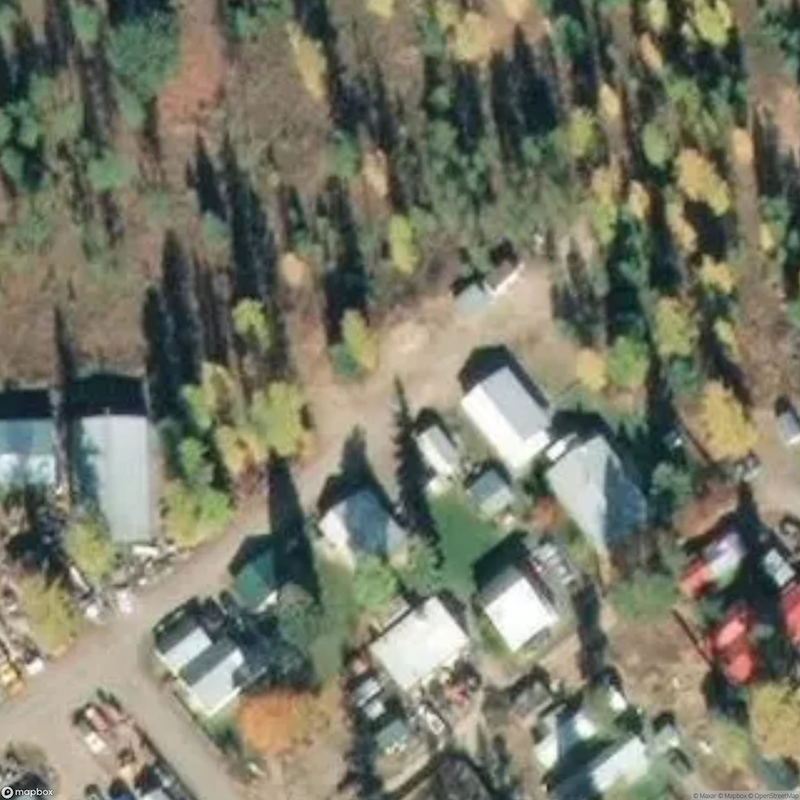}
        \end{minipage}
        \begin{minipage}[b]{0.16\textwidth}
            \includegraphics[width=0.95\textwidth]{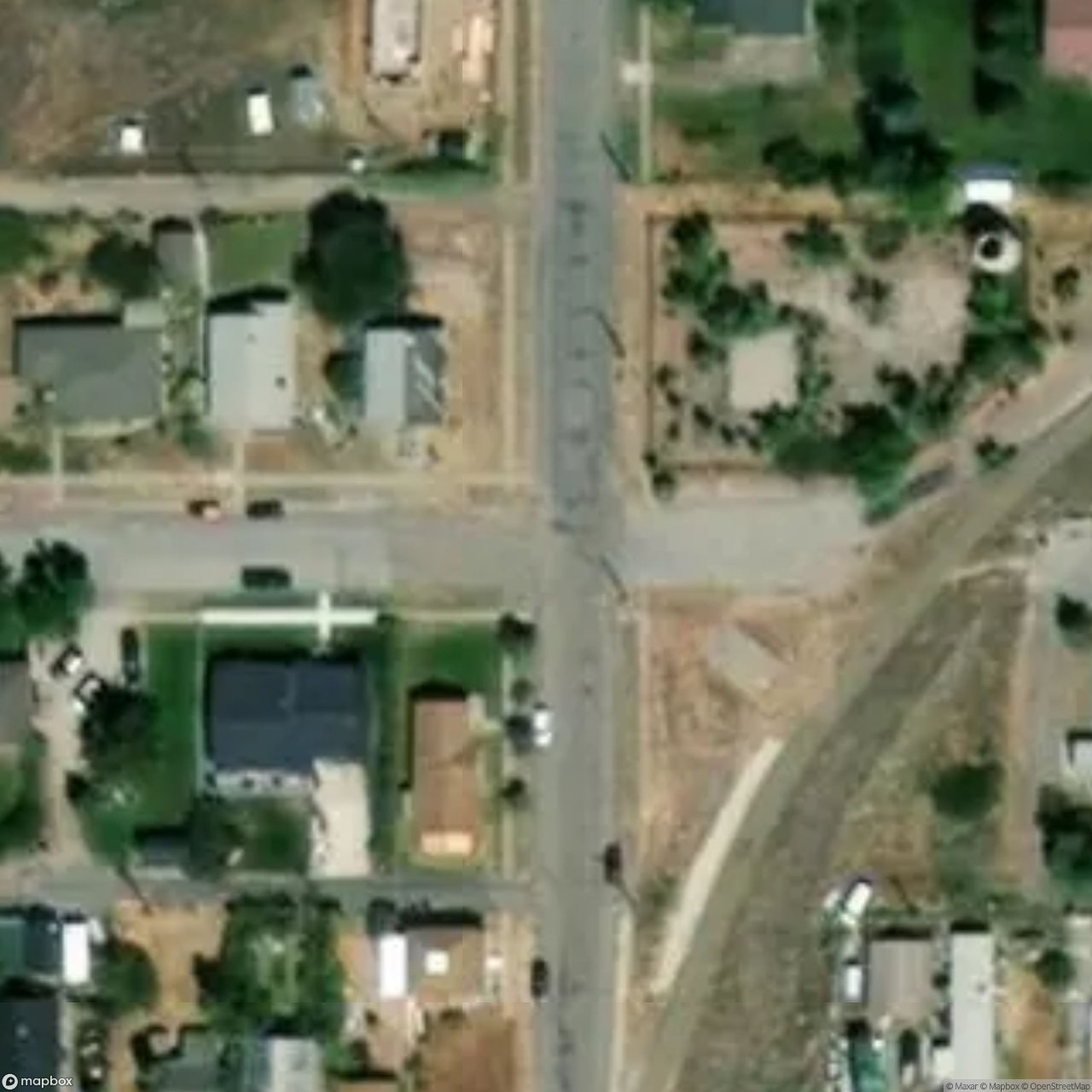}
        \end{minipage}
        \begin{minipage}[b]{0.16\textwidth}
            \includegraphics[width=0.95\textwidth]{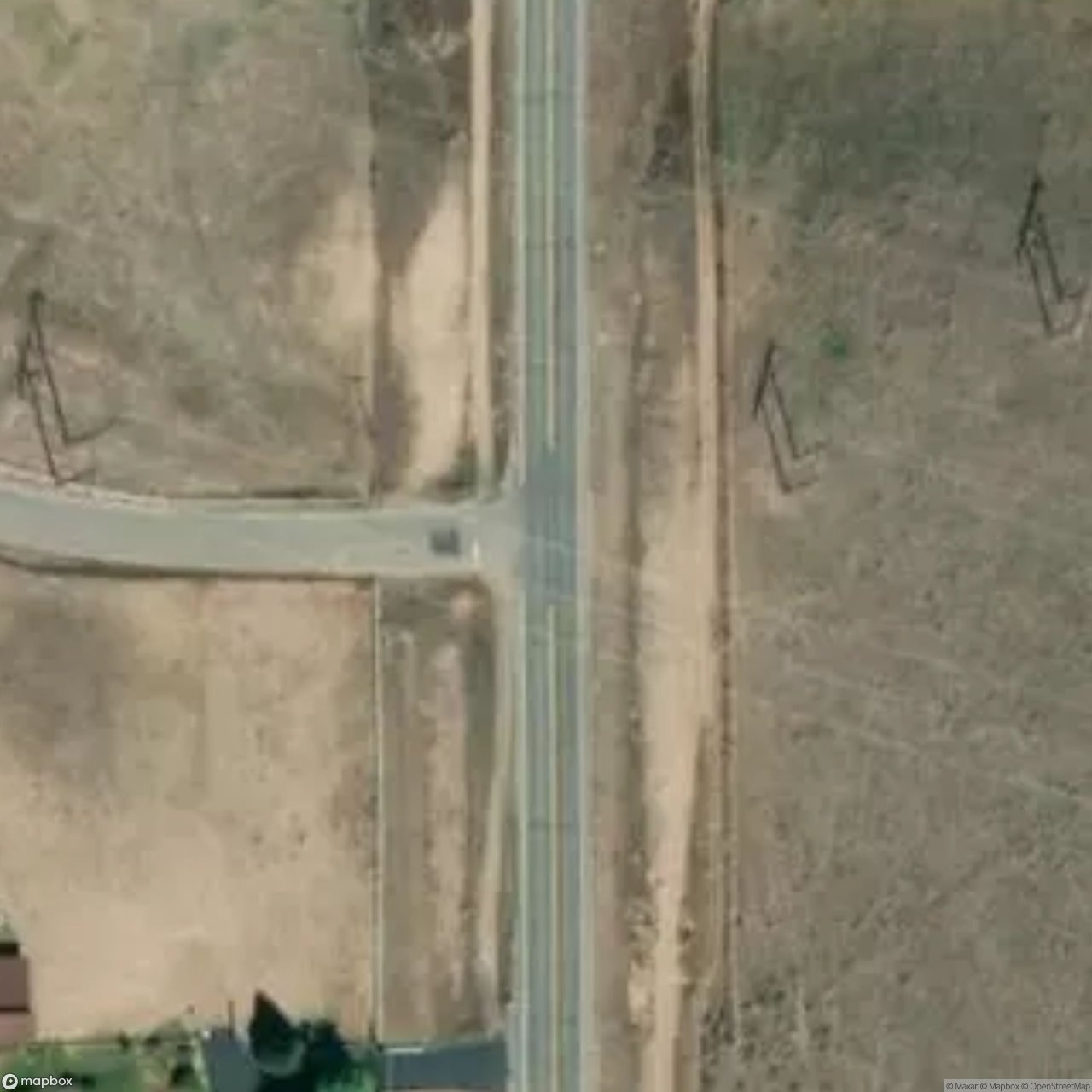}
        \end{minipage}
        \begin{minipage}[b]{0.16\textwidth}
            \includegraphics[width=0.95\textwidth]{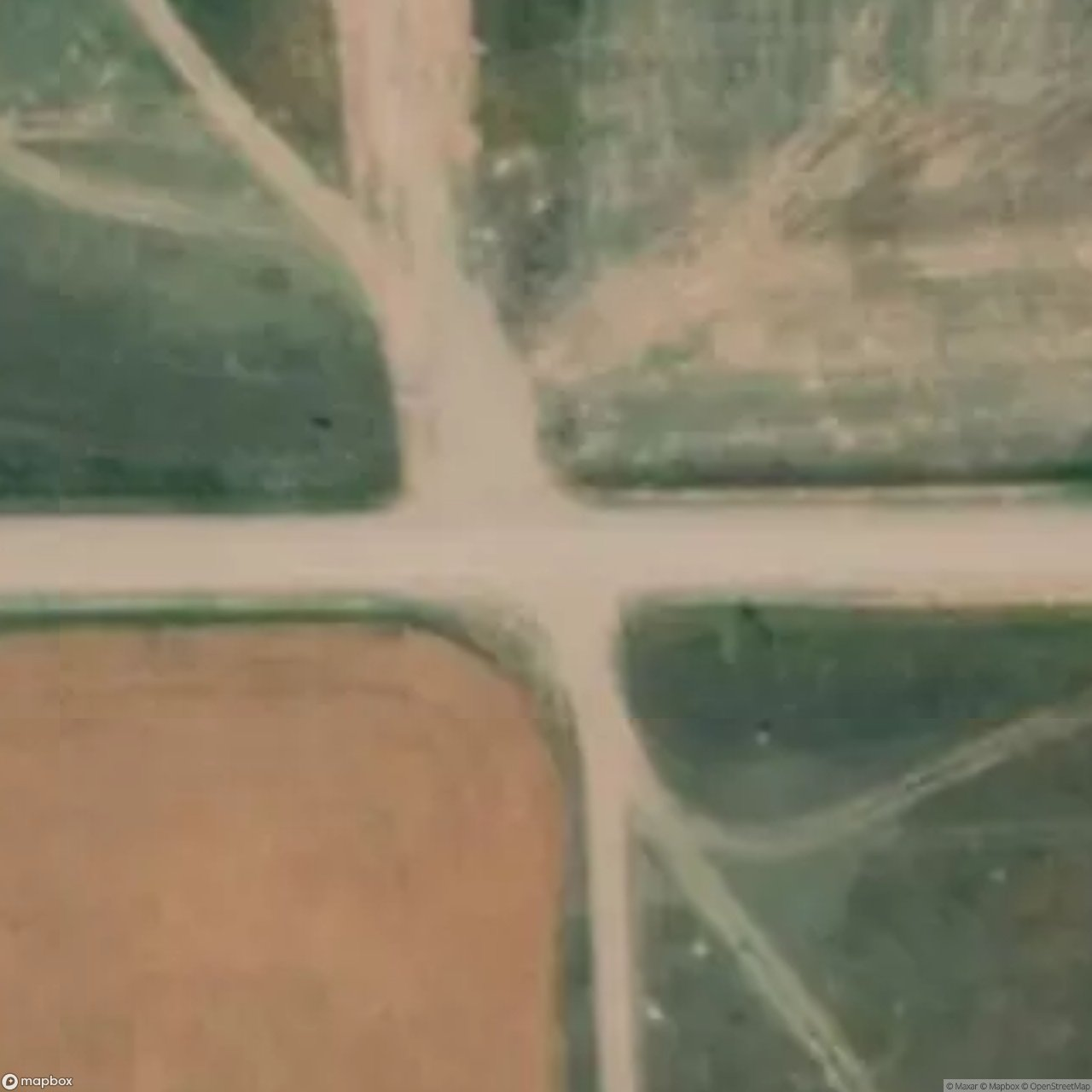}
        \end{minipage}
        \begin{minipage}[b]{0.16\textwidth}
            \includegraphics[width=0.95\textwidth]{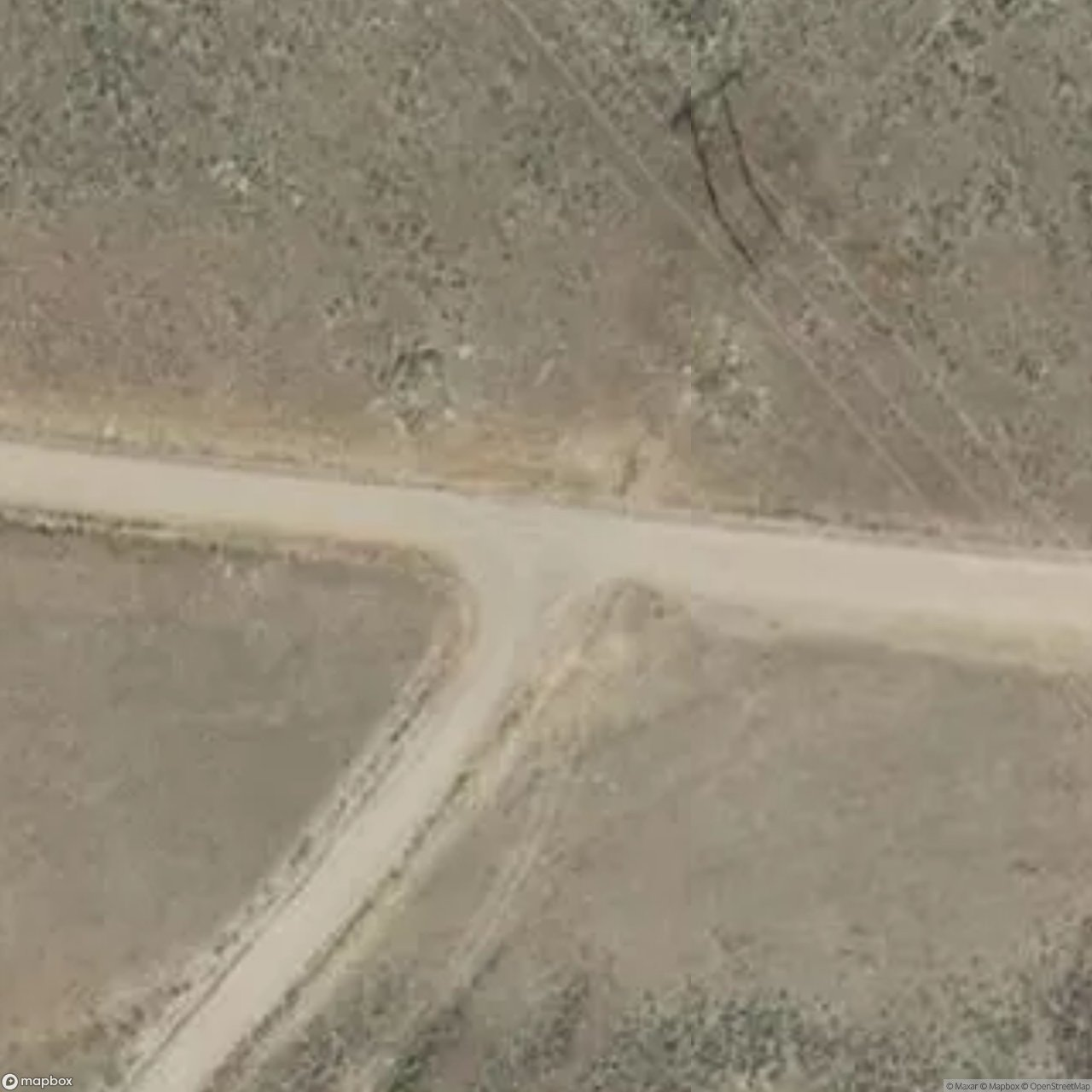}
        \end{minipage}
        \begin{minipage}[b]{0.16\textwidth}
            \includegraphics[width=0.95\textwidth]{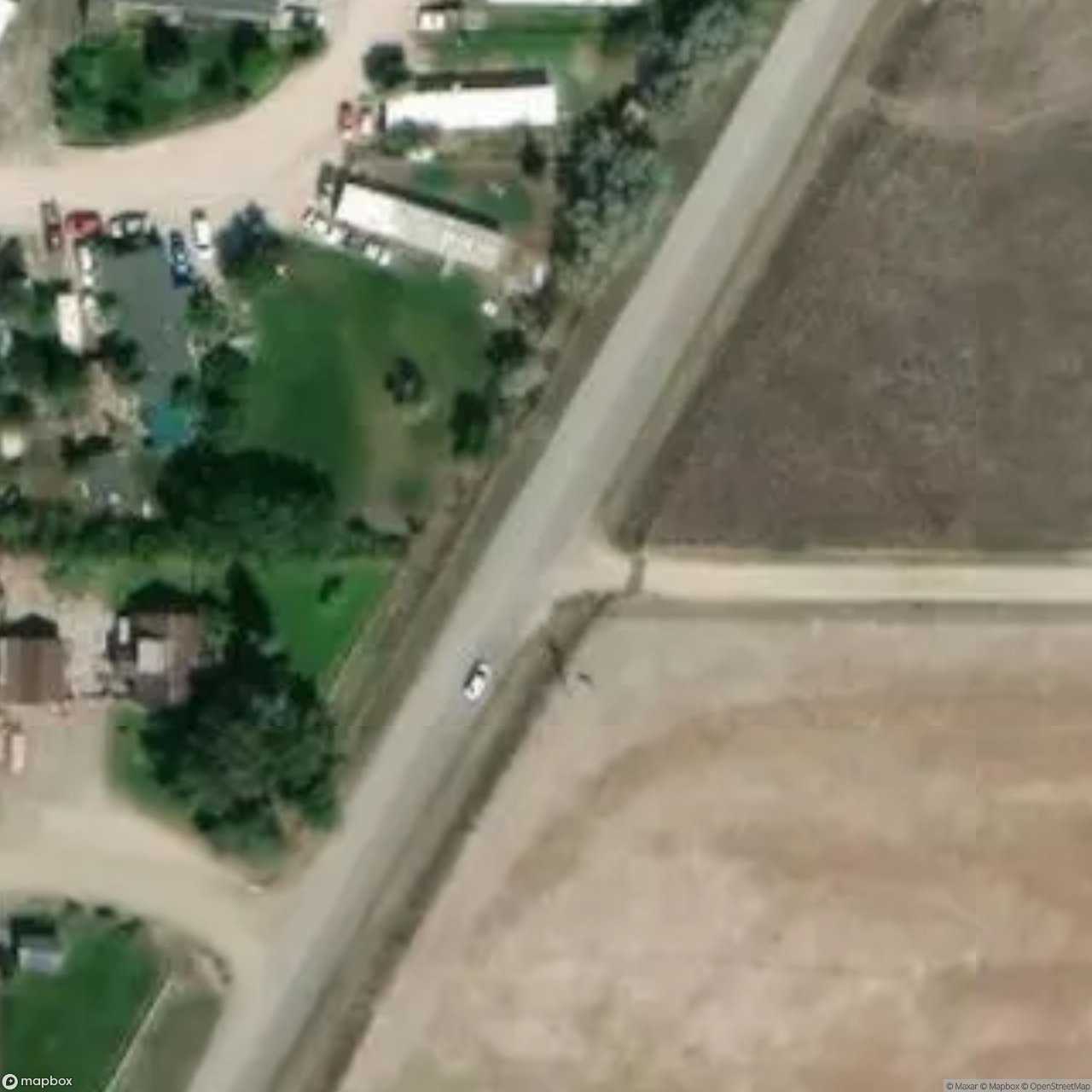}
        \end{minipage}
    }
    \subcaptionbox{Iowa}[1.0\textwidth]{
        \centering
        \begin{minipage}[b]{0.16\textwidth}
            \includegraphics[width=0.95\textwidth]{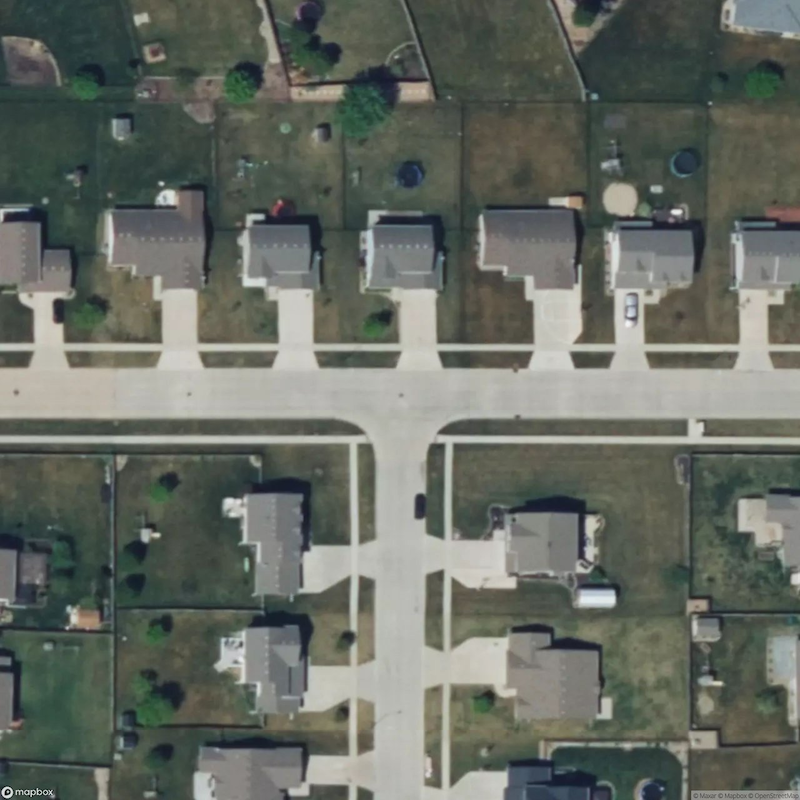}
        \end{minipage}
        \begin{minipage}[b]{0.16\textwidth}
            \includegraphics[width=0.95\textwidth]{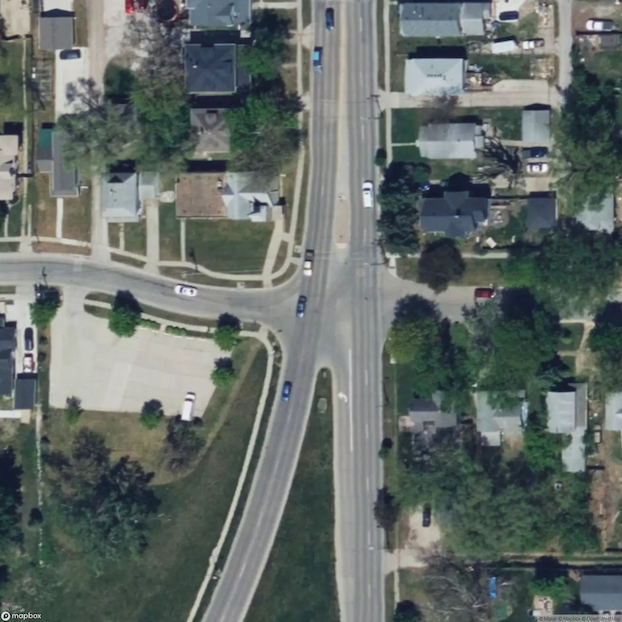}
        \end{minipage}
        \begin{minipage}[b]{0.16\textwidth}
            \includegraphics[width=0.95\textwidth]{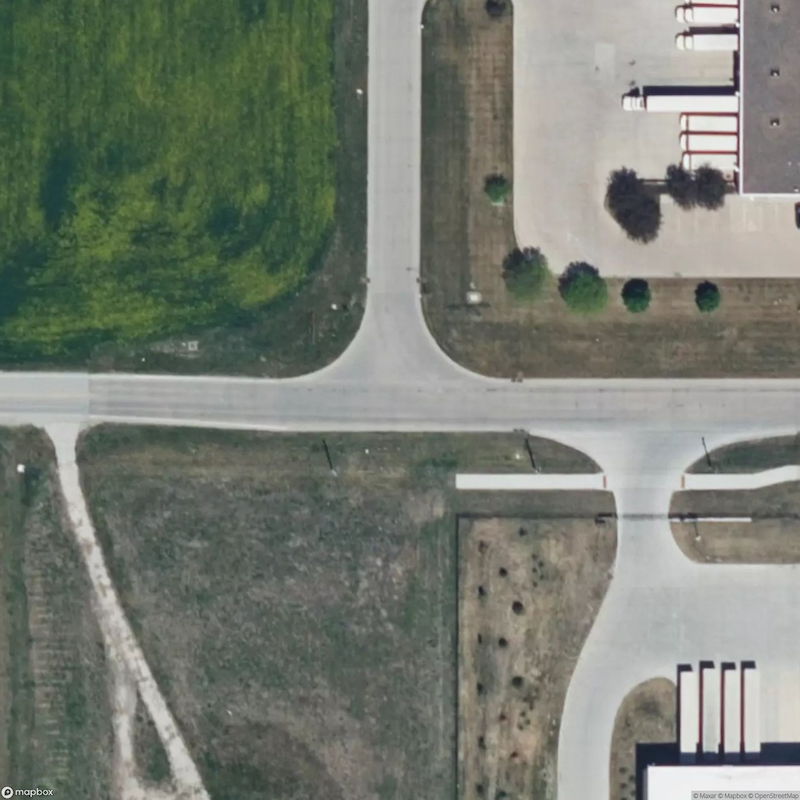}
        \end{minipage}
        \begin{minipage}[b]{0.16\textwidth}
            \includegraphics[width=0.95\textwidth]{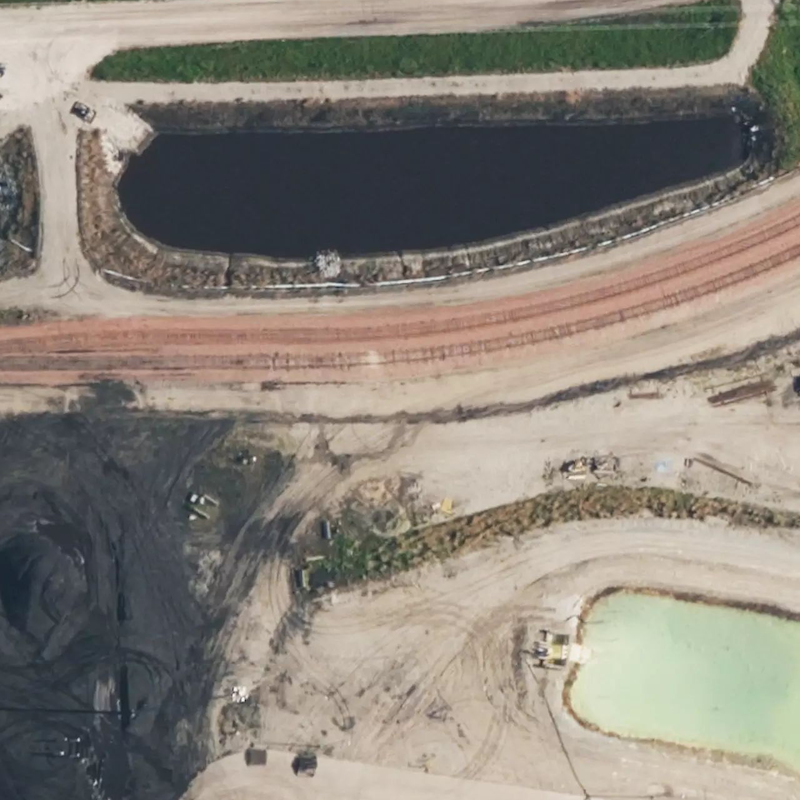}
        \end{minipage}
        \begin{minipage}[b]{0.16\textwidth}
            \includegraphics[width=0.95\textwidth]{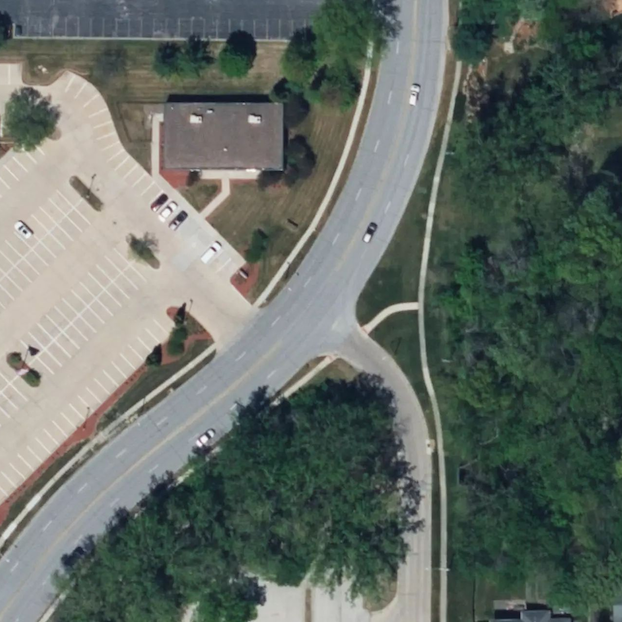}
        \end{minipage}
        \begin{minipage}[b]{0.16\textwidth}
            \includegraphics[width=0.95\textwidth]{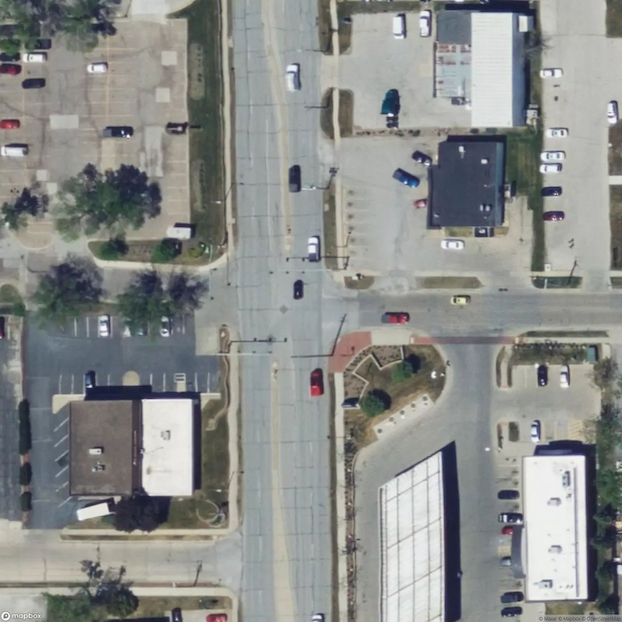}
        \end{minipage}
    }
    \caption{We showcase representative satellite images sampled from each of the six states included in our dataset. Each state is associated with six figures. These images exhibit diverse geographical and urban characteristics, ranging from dense urban intersections and suburban roadways to rural highways and mountainous terrains. Such variation reflects the heterogeneity of real-world driving environments across regions and provides rich visual cues that are critical for learning robust, transferable vision-based road features. These satellite views serve as the primary modality in our framework for capturing road layout, surrounding context, and surface-level conditions.}
    \Description{Sample satellite images of each state.}
    \label{fig_omitted_satellite_sample}
\end{figure*}